\documentclass[10pt,twocolumn,letterpaper]{article}

\usepackage[pagenumbers]{cvpr}      %

\usepackage{array}
\usepackage{longtable}
\usepackage{soul}
\usepackage{multirow}
\usepackage{dsfont}
\usepackage{anyfontsize}  %
\usepackage{xurl}  %

\usepackage{listings}
\definecolor{promptbg}{rgb}{0.97,0.97,0.97}
\lstdefinestyle{promptstyle}{
  literate={'}{\textquotesingle}1 {`}{\textasciigrave}1,
  basicstyle=\ttfamily\scriptsize,
  breaklines=true,
  breakatwhitespace=false,
  columns=fullflexible,
  keepspaces=true,
  frame=single,
  framesep=4pt,
  backgroundcolor=\color{promptbg},
  showstringspaces=false,
  xleftmargin=4pt,
  xrightmargin=4pt,
}

\definecolor{rmgreen}{HTML}{1B7A3D}

\newcommand{\ci}[1]{{\fontsize{5.5}{6.5}\selectfont #1}}
\newcommand{\cis}[1]{{\fontsize{6.5}{7.5}\selectfont #1}}
\newcommand{\val}[3]{#1\,\ci{[#2,#3]}}

\definecolor{hlincA}{HTML}{F4B6AE}
\definecolor{hlincB}{HTML}{F6C0B8}
\definecolor{hlcorA}{HTML}{B6E6C9}
\definecolor{hlcorB}{HTML}{B9E2DC}
\definecolor{hlspuC}{HTML}{DEC6EF}
\definecolor{hlparA}{HTML}{F6E29B}
\definecolor{hlparB}{HTML}{EFD27A}
\definecolor{hlparC}{HTML}{FAEDB9}

\definecolor{cCOR}{HTML}{1B7A3D}
\definecolor{cPAR}{HTML}{B8860B}
\definecolor{cINC}{HTML}{C0392B}
\definecolor{cMIS}{HTML}{D35400}
\definecolor{cSPU}{HTML}{8E44AD}

\newcommand{\jolia}{Jolia}

\providecommand{\prevver}{\textsuperscript{\textcolor{red}{$\circ$}}}

\AtBeginDocument{\crefname{appendix}{App.}{Apps.}\Crefname{appendix}{Appendix}{Appendices}%
  \crefname{subappendix}{App.}{Apps.}\Crefname{subappendix}{Appendix}{Appendices}}

\newlength{\valw}\newlength{\ciw}
\newcommand{\valwidths}[2]{\settowidth{\valw}{#1}\settowidth{\ciw}{\ci{#2}}}
\makeatletter
\renewcommand{\val}[3]{\ifdim\valw>0pt \def\val@v{#1}\def\val@c{[#2,#3]}\expandafter\val@peek\else #1\,\ci{[#2,#3]}\fi}
\newcommand{\val@peek}{\@ifnextchar${\val@box}{\val@box$$}}
\def\val@box$#1${\makebox[\valw][r]{\val@v}\,\makebox[\ciw][l]{\ci{\val@c}\if\relax\detokenize{#1}\relax\else$#1$\fi}}
\makeatother

\usepackage{placeins}
\usepackage{xurl}
\usepackage{fontawesome5}
\providecommand{\linenumbers}{}\providecommand{\nolinenumbers}{}

\definecolor{cvprblue}{rgb}{0.21,0.49,0.74}
\usepackage[pagebackref,breaklinks,colorlinks,allcolors=cvprblue]{hyperref}

\def\confName{CVPR}
\def\confYear{2027}

\title{RadMatch: Auditable Radiology Report Evaluation via Finding-Level Matching}

\author{Charles Corbière$^{1}$ \quad Léo Machado$^{1,2}$ \quad Baptiste Callard$^{1}$ \quad Aubin Charley$^{1}$\\
Korentin Le Floch$^{1,3}$ \quad Martin J. Willemink$^{4}$ \quad Pierre Manceron$^{1}$ \quad Corentin Dancette$^{1}$\\[0.6em]
{\small $^{1}$Raidium, Paris, France \qquad $^{2}$Department of Radiology, Fondation Ophtalmologique Adolphe de Rothschild, Paris, France}\\
{\small $^{3}$Service de Radiologie Interventionnelle, DACI, Gustave Roussy, Villejuif, France}\\
{\small $^{4}$Segmed, Inc., Palo Alto, CA, USA}\\[0.25em]
{\small \faGithub\ \href{https://github.com/raidium-med/RadMatch}{github.com/raidium-med/RadMatch}}
}

\begin{document}
\maketitle
{\renewcommand{\thefootnote}{}\footnotetext{Corresponding author: \textit{charles.corbiere@raidium.eu}}}
\begin{abstract}
As AI systems are increasingly used to draft radiology reports, reliably evaluating their clinical quality remains a critical challenge. We introduce \emph{RadMatch}, a multi-stage LLM-based framework that turns report evaluation into a structured, auditable finding-level record. It extracts findings from a reference and a candidate report, matches them by clinical equivalence, grades each match over ten clinical attributes, and gives every match and unmatched finding a verdict and a clinical-significance tier. To validate metrics on CT, we build and release \emph{MetricEval-BodyCT}, the first public metric-evaluation benchmark of chest and abdomen CT reports annotated by radiologists error by error. On it and three chest X-ray benchmarks, RadMatch with an open 31B judge on consumer GPUs agrees with radiologists more closely than previous published metrics, and even reaches inter-expert agreement on ReXVal and on the significant errors of MetricEval-BodyCT. Radiologists auditing its decisions on 60 studies confirm 98.5\% of its matches and 93.6\% of the errors it counts. When used to evaluate state-of-the-art radiology report generation models, RadMatch shows that most CT models, unlike several chest X-ray models, actually leave more significant errors than an image-blind normal template, as their added unsupported findings outnumber the reference findings they correctly report. We release RadMatch as open-source code and the MetricEval-BodyCT dataset.
\end{abstract}

\section{Introduction}
\label{sec:intro}

\begin{figure}[t]
  \centering
  \includegraphics[width=\linewidth]{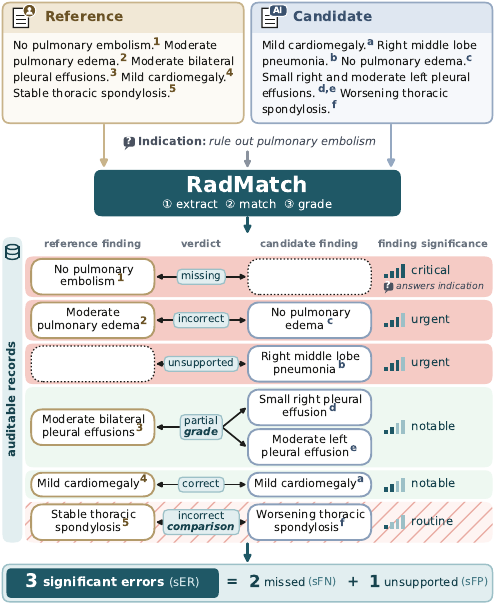}
  \caption{\textbf{RadMatch turns report evaluation into finding-level auditable records.} Every match and unmatched finding,
  normal ones included, gets a verdict, attribute errors (bold) and a significance tier. Only significant errors (red) count in
  sER, \eg the omitted ``no~\mbox{pulmonary} embolism'', not the \textsc{routine} one (hatched).}
  \label{fig:teaser}
\end{figure}

Automatically drafting trustworthy radiology reports could ease the growing burden of imaging studies radiologists must read~\cite{yildirim,flamingocxr}. Equally critical is how to \emph{evaluate} such drafts: does a generated report capture every clinically relevant finding, describe each at the right level of detail, and stay concise enough to be useful? The same finding can be phrased in many ways, yet flipping ``left'' to ``right'' or missing a pneumothorax can redirect care, whereas rewording a benign calcification does not. As vision--language models (VLMs) and agentic frameworks are increasingly used for radiology report generation~\cite{maira2,flamingocxr,structuredrrg}, reliable evaluation has become an active research direction~\cite{green,crimson,radeval,fineradscore,headctone}. Natural language generation metrics~\cite{bleu,rouge,bertscore} and clinical-concept matching~\cite{chexbert,radgraph} capture only fragments of this: they weigh every discrepancy equally, ignore clinical severity, and correlate poorly with radiologists~\cite{rexval,radeval}.

Large language model (LLM)-based metrics~\cite{green,crimson,fineradscore} have since emerged as the most clinically aligned approach. Prompted with the candidate report and its reference, they return a score or an error count that agrees with radiologists better than earlier metrics~\cite{green,crimson,radeval}, sometimes with free-text error descriptions~\cite{green} or a list of matched abnormal findings~\cite{crimson}. However, this record comes from one unchecked pass over both reports, so nothing guarantees a penalty reflects a missed pneumothorax rather than a reworded benign note; and a list of abnormal findings leaves out normal statements, such as the pertinent negative that answers the clinical question. Without a checked finding-level structure, these metrics are not auditable.

We present \emph{RadMatch}, an LLM-based framework that builds this structure by decomposing report evaluation into (1)~extraction of abnormal and normal atomic findings from each report, (2)~matching of candidate to reference findings, and (3)~significance-aware grading of the resulting pairs (\cref{fig:teaser}). Each match group or unmatched finding receives a verdict and an error profile over ten clinical attributes (status, location, laterality, extent, grade, appearance, diagnosis, certainty, comparison, measurement). The verdict, one of five message-understanding categories~\cite{muc}, separates a dangerous status inversion from an omission or an unsupported finding. Every decision is stored and shown in a dashboard, so each penalty traces to a finding pair. A few example pairs adapt RadMatch to a new region or modality, and it runs on-premises with an open judge.

We validate RadMatch at two levels: its \emph{scores}, against radiologists' error counts on established metric-evaluation benchmarks, and its \emph{decisions}, which radiologists audit one by one. As existing benchmarks are all chest X-ray, we also build and release \emph{MetricEval-BodyCT} for chest and abdomen CT, whose reports are longer and rich in measured findings. Finally, we use RadMatch to evaluate report generation on CT and chest X-ray, where its records show what a single score hides. In summary, our contributions are:
\begin{enumerate}
  \item \textbf{RadMatch}, an auditable, finding-level evaluation framework (\cref{sec:method}). With Gemma 4 31B~\cite{gemma4}, an open judge on consumer GPUs, it agrees with radiologists better than previous published metrics on four benchmarks and reaches inter-expert agreement on ReXVal and on MetricEval-BodyCT's significant errors (\cref{sec:experiments}). Radiologists auditing its decisions confirm $98.5\%$ of its matches and $93.6\%$ of the errors it counts.
  \item \textbf{MetricEval-BodyCT}, a public metric-evaluation benchmark of chest and abdomen CT reports annotated error by error: 300 candidates, each read by board-certified radiologists, with 1{,}582 errors, their category, significance and report line, and inter-expert agreement (\cref{sec:bodyct-benchmark}).
  \item \textbf{A thorough audit of radiology report generation} across 13 CT and 10 chest X-ray models on seven public datasets (\cref{sec:rrg}). We observe that several chest X-ray models beat an image-blind normal template, but most CT models leave more significant errors than it, as their added unsupported findings outnumber the reference findings they correctly report.
\end{enumerate}
We release RadMatch as open-source code and the MetricEval-BodyCT dataset.

\section{Related Work}
\label{sec:related}

\begin{figure*}[t]
  \centering
  \includegraphics[width=\linewidth]{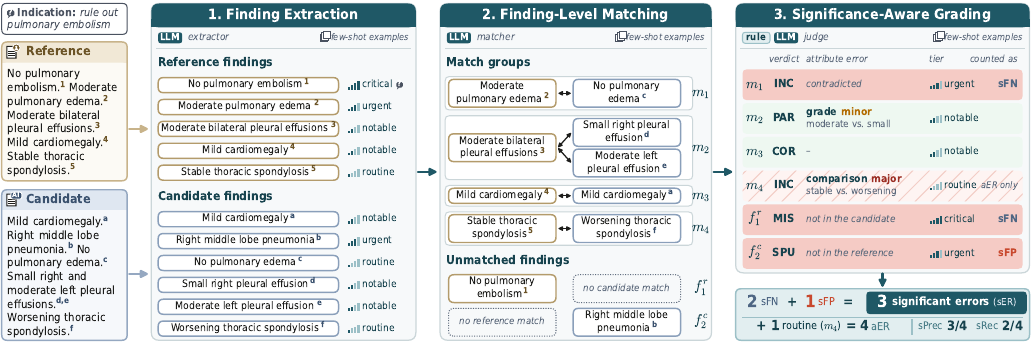}
  \caption{\textbf{RadMatch's three stages on one report pair.} \emph{Finding extraction}
  turns each report into findings with a significance tier, splitting the candidate's two
  effusions into $d$ and $e$. \emph{Finding-level matching} joins both to the reference's bilateral effusion ($m_2$) and leaves unmatched the
  reference's ``no pulmonary embolism'' ($f^r_1$), \textsc{critical} under the indication, and the candidate's pneumonia ($f^c_2$).
  \emph{Significance-aware grading} applies rules, then the LLM judge: each unit gets a verdict (\textsc{COR}rect/\textsc{PAR}tial/\textsc{INC}orrect/\textsc{MIS}sing/\textsc{SPU}rious) and graded attribute
  errors. The bottom panel adds up the last column; sPrec and sRec are on significant findings.}
  \label{fig:pipeline}
\end{figure*}

\subsection{Radiology Report Generation}
Most recent progress has targeted chest X-ray, whose 2D images are easier to model and have large public corpora, \eg MIMIC-CXR~\cite{mimiccxr}. Early models aligned a domain image encoder to a fine-tuned LLM~\cite{xraygpt,maira1}. MAIRA-2~\cite{maira2} and related models ground findings in image regions or segmentations~\cite{rgrg,mairaseg}; others add retrieval, re-alignment or self-critique~\cite{factmmrag,dart,selfcritiquing}, counter the language prior by topic-wise generation or contrastive decoding~\cite{topicguided,cwcd}, or write structured reports~\cite{structuredrrg}. Medical VLMs such as CheXagent~\cite{chexagent}, CheXOne~\cite{chexone}, MedGemma~\cite{medgemma,medgemma15} and Lingshu~\cite{lingshu} also write reports, and the latest agentic frameworks add detection, grounding and retrieval tools~\cite{medrax,mmagentrrg,radfabric,eviagent}. CT is harder: a volume yields thousands of visual tokens. 3D models pair a volume encoder, often a CT foundation model~\cite{curia,curia2,curiamae,merlin}, with an LLM~\cite{radfm,m3d,ctrate,reg2rg,med3dvlm}. Recent work turns the volume into fewer, more informative tokens~\cite{btb3d,resolutionreduction,mimickingrad}, and Jolia~\cite{jolia} learns one query per organ that pools its image features, aligned with that organ's report section. NV-Reason-CT~\cite{nvreasonct} instead reasons over the native 3D volume before reporting. These models can collapse onto template-like reports, near-identical across patients~\cite{templatecollapse}, yet are mostly scored with lexical or label-level metrics. RadMatch shows that most CT models leave more significant errors than an image-blind normal template, which a radiologist would otherwise edit (\cref{sec:rrg}).

\subsection{Evaluating Generated Radiology Reports}

\noindent\textbf{Lexical and clinical-concept metrics.}
Early metrics borrowed from natural language generation, such as BLEU~\cite{bleu}, ROUGE~\cite{rouge}, METEOR~\cite{meteor} and BERTScore~\cite{bertscore}, measure surface or contextual overlap, penalizing paraphrase yet missing factual errors~\cite{ctestmetric}. Clinical-concept metrics instead score extracted content: CheXbert~\cite{chexbert} over fixed pathology labels, RadGraph-F1~\cite{radgraph,rexval} over entities and relations, the ontology-normalized HeadCT-ONE~\cite{headctone} and the keyword-based FORTE~\cite{braingpt} beyond chest X-ray, and the learned RadCliQ composite~\cite{rexval}. For CT, CRG~\cite{crgscore} weights label-level hits and misses against false alarms by the abnormality rate, by default over CT-RATE's 18 chest CT labels. These capture concept presence but ignore clinical significance and are brittle to negation and uncertainty. RadMatch instead records each finding's status, certainty and significance tier, and a few example pairs, not a label set, adapt it to a new region or modality.

\noindent\textbf{LLM-based metrics.}
Prompting an LLM to judge factual correctness now correlates best with radiologists~\cite{fineradscore,chexprompt,gemascore,mrscore,llmradjudge,radreason}. GREEN~\cite{green} counts clinically significant errors in six categories with a free-text description of each. CRIMSON~\cite{crimson} adds patient context and significance weights. RadFact~\cite{maira2}, with a CT variant~\cite{colipri}, like general-domain atomic-fact metrics~\cite{factscore,aloha,dcscore}, splits both reports into single-finding phrases and checks each one's entailment in both directions. None gives a typed record of every finding: GREEN's errors are free text, tied to no sentence; RadFact's labels are binary, with no error type or significance; and CRIMSON extracts, matches and grades abnormal findings only, in one LLM call over both reports with no consistency check, so a dropped pertinent negative goes unseen. These metrics can also be gamed as training rewards: GREEN, as judge in CheXalign~\cite{chexalign}, rewarded reports up to $2.5\times$ longer, mostly by repetition. RadMatch builds such a record in separate, checked stages, normal findings included, with a verdict, typed attribute errors and a significance tier for every finding, so each counted error traces to a finding pair. As it counts only errors, padding does not improve its score (\cref{app:padding}).

\noindent\textbf{Metric-evaluation benchmarks.}
A metric is validated by how closely its scores rank candidate reports as radiologists judge them against the reference. On chest X-ray, radiologists count each candidate's errors, split into significant and insignificant, in ReXVal~\cite{rexval} and RadEvalExpert; the latter comes with RadEval~\cite{radeval}, which runs lexical, clinical-concept and LLM-based metrics on the same reports. RadPref~\cite{crimson} instead rates pairs of candidates and lists their errors, and CLEAR-Bench~\cite{clear} labels conditions and their attributes. CT benchmarks, CTest-Metric~\cite{ctestmetric} and CT-FineBench~\cite{ctfinebench}, only rate whole reports, so neither can check an error count, although body CT reports are longer and dominated by incidental, measured and longitudinal findings. MetricEval-BodyCT is, to our knowledge, the first public CT benchmark annotated error by error.

\section{RadMatch Evaluation Framework}
\label{sec:method}

RadMatch is an LLM-based evaluation framework that compares a candidate radiology report $c$ with a reference $r$ by building an auditable, finding-level record of how they differ, from which it computes error metrics, foremost the number of \emph{clinically significant} errors. Three stages each write part of the record (\cref{fig:pipeline}): (1)~an LLM extracts a set of structured findings from each report, (2)~an LLM matches candidate to reference findings by clinical equivalence, and (3)~rules and an LLM judge grade every difference, so that each matched and unmatched finding receives a verdict and a significance tier. Every LLM step reads few-shot examples reviewed by board-certified radiologists and, when available, the study indication, which an optional preliminary call extracts (\cref{app:prompt-indication}). While initially designed for five anatomy--modality pairs, RadMatch extends to a new one with new few-shot examples.

\subsection{Finding Extraction}
\label{sec:findings}
A \emph{finding} is a single, atomic clinical observation, meaning the smallest unit of a report to which a clinical consequence can attach (\eg ``small left pleural effusion'' or ``no pneumothorax''). The LLM extractor reads each report on its own and yields the finding sets $\mathcal{F}_r=\{f_i^r\}_i$ and $\mathcal{F}_c=\{f_j^c\}_j$. A sentence that states several findings is split, as exemplified in \cref{fig:pipeline}. Each finding $f$ is a structured record:
\begin{itemize}
  \item a \emph{status}: normal or abnormal;
  \item its free-text \emph{statement}, which carries the location, extent, grade, appearance, diagnosis and certainty descriptors graded in \cref{sec:scoring};
  \item a \emph{laterality}: left, right, bilateral or none;
  \item a longitudinal \emph{comparison}: stable, improving, worsening, new, resolved or none;
  \item an optional set of \emph{measurements}, each a (value, unit, category) triple;
  \item a \emph{significance} tier $\sigma(f)$, ordered from most to least urgent: \textsc{critical}${}>{}$\textsc{urgent}${}>{}$\textsc{notable}${}>{}$\textsc{routine}.
\end{itemize}

We adapt the tiers from the American College of Radiology (ACR) actionable-reporting framework and communication practice parameter~\cite{larson2014,acrcommunication}, which stratify findings by the urgency of the action they warrant, from immediately life-threatening (\eg tension pneumothorax) to incidental. A negative finding is tiered by the question it answers. For instance, on a CT pulmonary angiogram for suspected embolism, omitting or flipping ``no pulmonary embolism'' is \textsc{critical} (\cref{fig:teaser}), whereas a rule-out that answers no clinical question is \textsc{routine}. Our \textsc{notable} tier is broader than ACR's third category: it covers findings that need follow-up or work-up, benign ones included (\eg a 7\,mm gallbladder polyp), while \textsc{routine} holds completeness statements, low-impact normals, expected post-treatment states and a list of findings below follow-up size limits (\cref{app:prompt-extraction}). Attaching significance to the finding lets one scheme both decide which errors count and define the finding subsets the metrics report on.

\subsection{Finding-Level Matching}
\label{sec:matching}
The matching stage groups reference and candidate findings that describe the same clinical entity, \ie the same organ or structure and the same abnormality, whatever the wording or the diagnosis given to it, into \emph{match groups} $\mathcal{M}=\{m_k\}$, where each $m_k=(R_k,C_k)$ pairs a reference side $R_k\subseteq\mathcal{F}_r$ with a candidate side $C_k\subseteq\mathcal{F}_c$. Matching is \emph{many-to-many}, so it absorbs differences in how reports split findings: for instance, in \cref{fig:pipeline}, $m_2$ joins ``moderate bilateral pleural effusions'' to the candidate's right and left effusions. Findings in no group form the omissions $\mathcal{F}_r^-\subseteq\mathcal{F}_r$ and the unsupported findings $\mathcal{F}_c^-\subseteq\mathcal{F}_c$. Three choices keep each error where it belongs. A status conflict does not block a match: ``moderate pulmonary edema'' and ``no pulmonary edema'' concern one entity ($m_1$), so their disagreement is one status error, not an omission plus an unsupported finding. Likewise, a laterality flip or a different part of the same organ is matched and graded, whereas a finding in another organ is a different entity. And a negative denies only what it names (``no bowel obstruction'' does not deny colitis). The matching is a single LLM call that sees each finding's text, status and laterality; its output is checked so that every finding lies in at least one group or else in its unmatched set, and a call that fails the check is retried.

\subsection{Significance-Aware Grading}
\label{sec:scoring}

\begin{table}[t]
  \centering
  \caption{\textbf{Attributes graded within a match.} What each covers and who grades it.}
  \label{tab:attributes}
  \footnotesize
  \setlength{\tabcolsep}{3pt}
  \begin{tabular}{@{}lll@{}}
    \toprule
    Attribute & Covers & Graded by \\
    \midrule
    \emph{status} & abnormal or normal & rule \\
    \emph{location} & where in the organ: lobe, segment, station & LLM \\
    \emph{laterality} & left, right, bilateral, none & rule \\
    \emph{extent} & how many, how far: number, spread, invasion & LLM \\
    \emph{grade} & severity or category & LLM \\
    \emph{appearance} & shape, margins, density, enhancement & LLM \\
    \emph{diagnosis} & what it is: entity, subtype, age, cause & LLM \\
    \emph{certainty} & hedging: definite, probable, possible & LLM \\
    \emph{comparison} & change since a prior: new, stable, worse & rule, LLM \\
    \emph{measurement} & sizes, counts and densities & rule, LLM \\
    \bottomrule
  \end{tabular}
\end{table}

Within a match, every difference is assigned to one of the ten clinical attributes and graded \emph{major}, when it would change management, or \emph{minor}, when the candidate is partly right. Rules compare status, laterality and comparison, and counts, ratios and densities against per-category thresholds; the LLM judge grades the size and comparison differences they flag and the six free-text attributes (\cref{tab:attributes}). Detail only the candidate gives is an error only if it adds a clinical claim (\eg a complication), and two normal statements differ at most by a hedge. Between two \textsc{routine} findings, only status, a benign-to-active change, counts, ratios and densities can be major ($m_4$ in \cref{fig:pipeline}).

Each match group, omission and unsupported finding then receives a \emph{verdict}, one of the message-understanding categories~\cite{muc}: a match is \textsc{COR}rect (no difference), \textsc{PAR}tial (only minor differences) or \textsc{INC}orrect (a status inversion, a major difference, or any hedge on a \textsc{critical} reference finding); an omission is \textsc{MIS}sing; and an unmatched candidate finding is \textsc{SPU}rious, \ie unsupported. RadMatch reads the reference's silence as absence: an unsupported finding is an error only when it asserts an abnormality, and an omission is an error unless it is a \textsc{routine} normal statement, which reports include or leave out by house style. Errors are counted per finding and form $\mathcal{E}_{\mathrm{err}}$: each reference finding that no match conveys (an omission that is an error, or one whose matches are all \textsc{INC}), each unsupported abnormal finding, and each candidate finding contradicting a reference finding that another candidate finding conveys. An error $e$ on a reference finding $f^r$ with matched candidate findings $C(f^r)$ takes the highest tier among them,
\begin{equation}
\sigma(e)=\max_{f\in\{f^r\}\cup C(f^r)}\sigma(f),
\end{equation}
and an error on a candidate finding its own tier, so that under- and over-calling a significant entity are both judged at its stakes: in \cref{fig:pipeline}, $m_1$ is \textsc{urgent}, the tier of the reference's edema, although the candidate's ``no pulmonary edema'' is \textsc{routine}. The findings, groups, graded attributes, verdicts and tiers form the stored record (\cref{app:dashboard}).

\subsection{Metrics}
\label{sec:metrics}
RadMatch's main metric is the \emph{significant-error count} per report, sER:
\begin{equation}
\mathrm{sER}(r,c)=\sum_{e\in\mathcal{E}_{\mathrm{err}}}\mathds{1}\!\left[\sigma(e)\in\mathcal{S}\right],
\end{equation}
where $\mathcal{S}=\{\textsc{critical},\textsc{urgent},\textsc{notable}\}$ is the set of \emph{significant} tiers, \ie every tier but \textsc{routine}, and $\mathds{1}[\cdot]$ is the indicator function; sER is thus the number of clinically significant errors the candidate makes relative to the reference. The \emph{all-error count}, $\mathrm{aER}=|\mathcal{E}_{\mathrm{err}}|$, also counts the errors on \textsc{routine} findings. We split sER into sFN, the errors on the reference side (\textsc{MIS}, and \textsc{INC} on a significant reference finding), and sFP, those on the candidate side (abnormal \textsc{SPU}, contradicting candidate findings, and \textsc{INC} on a candidate finding whose reference counterpart is not significant). Beyond counts, we report recall and precision on significant findings (sRec, sPrec) and, at the stricter \emph{triage} threshold, on \textsc{critical} and \textsc{urgent} ones. Each finding counts at its own tier: recall is the fraction of the reference's findings at the threshold that the candidate conveys without a major error (\textsc{COR} or \textsc{PAR}), and precision the fraction of the candidate's findings at the threshold that are matched without one, unreferenced normal statements excluded. Attribute-level breakdowns and per-subset views (normal, abnormal, longitudinal, measured findings) localize \emph{where} a system errs; all follow from the stored records, so any metric can be inspected post hoc.

\section{MetricEval-BodyCT}
\label{sec:bodyct-benchmark}
Existing metric-evaluation benchmarks with radiologist-annotated errors are mostly chest X-ray. Yet body CT reports are longer, organized by organ and rich in measured findings. We thus build and release MetricEval-BodyCT, to our knowledge the first public benchmark of chest and abdomen CT reports annotated error by error: board-certified radiologists read each candidate and log each error's category and significance.

\paragraph{Studies and candidates.} We select 100 de-identified CT studies, 50 chest and 50 abdomen--pelvis, from multiple US centers.
\begin{figure}[t]
  \centering
  \includegraphics[width=\linewidth]{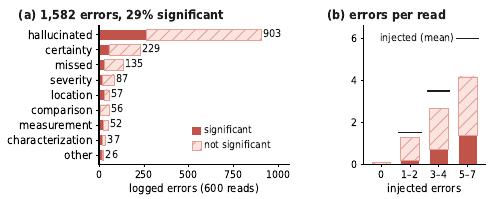}
  \caption{\textbf{MetricEval-BodyCT annotations.} (a)~Errors logged over the 600 reads, by category. (b)~Errors per read, by
  number of errors injected (line: their mean). Solid: clinically significant.}
  \label{fig:bodyct}
\end{figure}
\begin{figure*}[t]
  \centering
  \includegraphics[width=\linewidth]{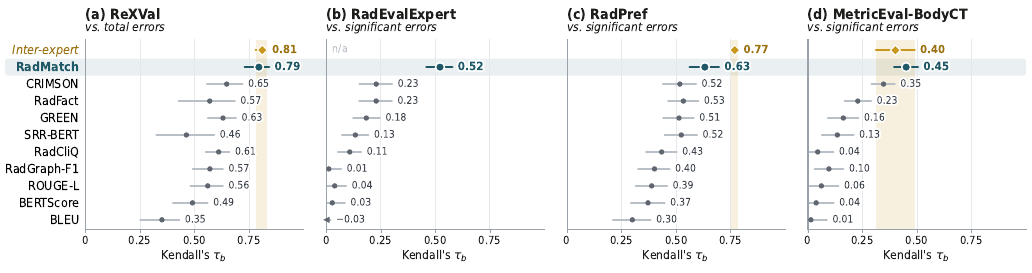}
  \caption{\textbf{Agreement with radiologists.} Kendall's $\tau_b$ (95\% CI; higher is better) of RadMatch (Gemma 4 31B)
  and prior metrics on (a)~ReXVal ($n{=}200$), (b)~RadEvalExpert ($n{=}624$), (c)~RadPref ($n{=}200$) and (d)~MetricEval-BodyCT
  ($n{=}300$). Inter-expert row and gold band: value and 95\% CI from GREEN on ReXVal and ours on MetricEval-BodyCT, mean and range of
  expert pairs on RadPref.}
  \label{fig:agreement}
\end{figure*}
Each reference report is paired with three candidates: a frontier proprietary LLM\footnote{Claude Opus 4.8~\cite{claudeopus48card}.} rewrites its findings and impression in one reporting style per study and injects an increasing number of clinically significant errors (0 to 7 per candidate), stratified over nine categories and checked against the study's findings (\cref{app:bodyct}). Controlled injection fixes style and length within a study, so its candidates differ only in content, and sets the number of errors; RadMatch is also audited on real model outputs (\cref{sec:clinical-validation}).

\paragraph{Annotation.} Two of four board-certified radiologists independently read each of the 300 candidates against its reference, blinded to the injected errors, logging every error with its category, significance (whether it would change management or follow-up) and report line, and ranking the three candidates of a study by fidelity to the reference and by how well the impression answers the indication. The injected errors only seed the candidates; the ground truth is the annotation. The radiologists logged 1{,}582 errors (2.6 per read), mostly hallucinated findings and certainty errors, and rated only $29\%$ of them significant, although every injected error was planned as such (\cref{fig:bodyct}a); per read, the logged errors rise with the number injected, on average at $70$ to $84\%$ of it (\cref{fig:bodyct}b).

\paragraph{Protocol.} A metric scores the 300 candidates; agreement is Kendall's $\tau_b$ with each radiologist's judgment~\cite{rexval,radeval,crimson}, pooled over candidates. The primary endpoint is the significant-error count; the others are the total-error count, the rankings and the per-category counts (\cref{app:bodyct}). Inter-expert agreement, a reference level for a metric, is lowest on the primary endpoint: radiologists disagree more on which errors are significant than on which exist. We release the reports, annotations and evaluation script for academic and research use under a data use agreement\footnote{\href{https://huggingface.co/datasets/raidium/MetricEval-BodyCT}{huggingface.co/datasets/raidium/MetricEval-BodyCT}}.

\section{Experiments}
\label{sec:experiments}

\subsection{Experimental Setup}
\label{sec:setup}
We validate RadMatch by how closely its scores of candidate reports agree with radiologists' judgments of the same candidates against their reference, on three chest X-ray benchmarks and on MetricEval-BodyCT. In \emph{ReXVal}~\cite{rexval}, each of 50 MIMIC-CXR references is paired with four candidates retrieved from the training set, and six radiologists count their errors. \emph{RadEvalExpert}~\cite{radeval} has 624 candidates for 208 references (148 findings and 60 impression sections), written by three report generation models, with each error rated significant or insignificant. In \emph{RadPref}~\cite{crimson}, each of 100 references has two candidates, which three radiologists rate from 1 to 5 and annotate for errors.

RadMatch runs with the open Gemma 4 31B~\cite{gemma4} as its judge, served locally on two 32\,GB consumer GPUs, with the few-shot examples of the benchmark's modality. \cref{sec:robustness} varies the judge. We compare against lexical metrics (BLEU~\cite{bleu}, ROUGE-L~\cite{rouge}, BERTScore~\cite{bertscore}), clinical-concept metrics (RadGraph-F1~\cite{radgraph,rexval}, RadCliQ~\cite{rexval}, SRR-BERT~\cite{srrbert}) and LLM-based metrics (GREEN~\cite{green}, CRIMSON~\cite{crimson}, RadFact~\cite{maira2,colipri}), using reported numbers where available. Following each benchmark's protocol, agreement is Kendall's $\tau_b$ with the radiologists' total or significant error counts (study-clustered bootstrap 95\% CIs, oriented so that higher is better); RadMatch's sER is compared with significant errors, and its aER with total errors, ratings and rankings.

\subsection{Agreement with Radiologists}
\label{sec:agreement}

\cref{fig:agreement} compares RadMatch with prior metrics. On chest X-ray, its error counts agree with radiologists more closely than previous published metrics. On ReXVal it reaches $\tau_b=0.79$, close to the inter-expert agreement reported by GREEN ($0.81$), and on the harder RadEvalExpert $0.52$, more than twice the best prior metric. On RadPref, RadMatch reaches $0.63$ on the radiologists' significant errors, against $0.53$ for the best prior metric, RadFact. For a fair comparison with prior LLM-based metrics, we run RadMatch with
the model each uses, or its closest version. On ReXVal, RadMatch with GPT-4.1~\cite{gpt41} reaches $0.79$ against $0.63$ reported for GREEN, and RadMatch with GPT-5.2~\cite{gpt52} reaches $0.79$ against $0.65$ for CRIMSON. Further endpoints and breakdowns are in \cref{app:agreement}.

On MetricEval-BodyCT, RadMatch reaches $0.45$ on significant errors, on par with the inter-expert agreement ($0.40$), against $0.35$ for CRIMSON, the second best. On total errors, it also beats CRIMSON ($0.53$ vs.\ $0.44$), mostly on hallucinated findings, the most frequent category, but stays below the inter-expert agreement ($0.69$; \cref{app:bodyct}).

\begin{figure*}[t]
  \centering
  \includegraphics[width=\linewidth]{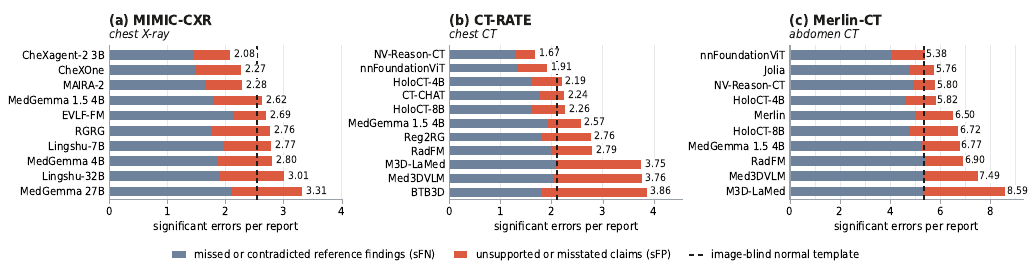}
  \caption{\textbf{Radiology report generation scored by RadMatch.} Significant errors per report (Gemma 4 31B judge) on one dataset
  per region, split into sFN and sFP, against the image-blind normal template (dashed). All seven datasets in \cref{app:rrg}.}
  \label{fig:rrg}
\end{figure*}

\subsection{Clinical Validation of RadMatch}
\label{sec:clinical-validation}

Radiologists review RadMatch's decisions one by one on 60 studies, each with one reference and one candidate report: 24 chest CT, 24 abdomen--pelvis CT and 12 chest X-ray from six datasets, with candidates from 13 report generation models and six normal templates. Each of three radiologists reviews 20 studies in a web application showing the two reports and RadMatch's decisions (1{,}091 in all; Gemma 4 31B judge), and confirms or corrects, with a reason, each match, error call (significant, insignificant or none) and attribute grade. Radiologists confirm $98.5\%$ of its matches and $99.4\%$ of the no-error calls, and $93.6\%$ of the errors RadMatch counts (\cref{app:clinical-validation}). Disagreement concentrates on significance: they keep $75\%$ of its $100$ significant-error calls, lower $25$ ($21$ to insignificant) and raise only $10$, so RadMatch over-calls significance rather than under-calls it; attribute grades follow the same pattern ($12$ of $44$ minor or major grades changed). Per study, RadMatch's significant-error count and the corrected one agree at $\tau_b=0.71$ [$0.58$, $0.82$] and lie within one error in $88\%$ of the studies, although radiologists themselves draw the significance line differently (inter-expert $\tau_b$ of $0.40$ on MetricEval-BodyCT's significant errors vs.\ $0.69$ on its total errors). RadMatch thus errs on the high side on significance, but its ranking of studies and its other decisions hold.

\subsection{Ablation}
\label{sec:ablation}
\label{sec:robustness}
\paragraph{Robustness to the judge.} We run RadMatch with eight judges, from a 4B open medical model to the latest proprietary ones, served locally or through an API (\cref{tab:agreement}).
On chest X-ray, six of them, open and proprietary alike (Qwen3.8-27B~\cite{qwen38}, Gemma 4 31B, Claude Opus 5.5~\cite{claudeopus55}, GPT-4.1, GPT-5.2 and GPT-6 Astra~\cite{gpt6astra}), fall within $\tau_b\in[0.79,0.82]$ on ReXVal, $[0.51,0.53]$ on RadEvalExpert and $[0.61,0.64]$ on RadPref; the lighter Qwen3.5-35B-A3B~\cite{qwen35} joins them on the latter two but trails on ReXVal ($0.75$). On the significant errors of MetricEval-BodyCT, the same seven judges fall within $[0.40, 0.48]$, at or above the inter-expert agreement ($0.40$). The newest proprietary model brings no clear gain: GPT-6 Astra ties the open Qwen3.8-27B on ReXVal and leads on no other benchmark. Only the small MedGemma 1.5 4B~\cite{medgemma15} degrades sharply, to $0.38$ on RadEvalExpert and $0.17$ on MetricEval-BodyCT, as its matching output often fails validation. Apart from this small model, agreement is thus stable across judges, whether open or proprietary, local or served by API. We therefore use Gemma 4 31B for all main results: it is in the top band on every benchmark; unlike an API model, its open weights can be pinned, so scores stay comparable over time; and it runs on-premises on two consumer GPUs, so reports never leave the hospital.

\providecommand{\prevver}{\textsuperscript{\textcolor{red}{$\circ$}}}
\begin{table}[!h]
  \centering
  \caption{\textbf{RadMatch with different judges.} $\tau_b$ (95\% CI) on the endpoints of \cref{fig:agreement}; bold: best. $\checkmark$: served locally, else via an API. $^\dagger$LLM of a baseline metric (GREEN with GPT-4, CRIMSON).}
  \label{tab:agreement}
  \scriptsize
  \setlength{\tabcolsep}{2pt}
  \begin{tabular}{lccccc}
    \toprule
    Judge & Local & ReXVal & \begin{tabular}[b]{@{}c@{}}RadEval\\Expert\end{tabular} & RadPref & \begin{tabular}[b]{@{}c@{}}MetricEval-\\BodyCT\end{tabular} \\
    & & \textit{total} & \textit{significant} & \textit{significant} & \textit{significant} \\
    \midrule
    qwen3.8-27b & $\checkmark$ & \val{\textbf{0.82}}{.78}{.86} & \val{\textbf{0.53}}{.47}{.59} & \val{\textbf{0.64}}{.56}{.70} & \val{0.43}{.37}{.49} \\
    gpt-6-astra & & \val{\textbf{0.82}}{.76}{.86} & \val{0.52}{.46}{.58} & \val{0.63}{.56}{.69} & \val{0.45}{.39}{.51} \\
    opus-5.5 & & \val{0.80}{.73}{.85} & \val{\textbf{0.53}}{.47}{.58} & \val{\textbf{0.64}}{.57}{.70} & \val{\textbf{0.48}}{.42}{.53} \\
    gemma-4-31b & $\checkmark$ & \val{0.79}{.73}{.84} & \val{0.52}{.46}{.58} & \val{0.63}{.56}{.70} & \val{0.45}{.40}{.50} \\
    gpt-4.1$^\dagger$ & & \val{0.79}{.73}{.84} & \val{0.51}{.44}{.57} & \val{\textbf{0.64}}{.58}{.70} & \val{0.42}{.36}{.48} \\
    gpt-5.2$^\dagger$ & & \val{0.79}{.72}{.84} & \val{\textbf{0.53}}{.47}{.58} & \val{0.61}{.53}{.68} & \val{0.40}{.33}{.46} \\
    qwen3.5-35b-a3b & $\checkmark$ & \val{0.75}{.66}{.81} & \val{\textbf{0.53}}{.47}{.58} & \val{\textbf{0.64}}{.57}{.70} & \val{0.41}{.35}{.48} \\
    medgemma-1.5-4b & $\checkmark$ & \val{0.70}{.60}{.76} & \val{0.38}{.31}{.45} & \val{0.58}{.50}{.66} & \val{0.17}{.10}{.24} \\
    \bottomrule
  \end{tabular}
\end{table}

\paragraph{Contribution of each component.} We rescore RadMatch's stored records, changing one component at a time, and compare with a single call asking the same judge for the number of significant errors (\cref{app:ablations}). \emph{Missed and unsupported findings} are both needed: on ReXVal, counting only the missed gives $0.73$, only the unsupported $0.50$, and both $0.79$. \emph{Significance tiers} let the count match what radiologists counted: on MetricEval-BodyCT, sER agrees better with their significant errors ($0.45$ vs.\ $0.38$ for aER) and aER with their total errors ($0.53$ vs.\ $0.40$ for sER). The \emph{decomposition} into findings does not by itself raise agreement: the single call scores about as well ($0.81$ vs.\ $0.79$ on ReXVal, $0.56$ vs.\ $0.52$ on RadEvalExpert, $0.76$ vs.\ $0.78$ on RadPref's total errors, $0.42$ vs.\ $0.45$ on MetricEval-BodyCT's significant errors) and better on its total errors ($0.66$ vs.\ $0.53$). What the decomposition adds is a count that can be checked and explained: splitting misses from unsupported claims shows why CT models trail the normal template (\cref{sec:rrg}), and on MetricEval-BodyCT sER, unlike the single call, sets significant errors apart (\cref{app:single-call}).

\section{Benchmarking Report Generation}
\label{sec:rrg}

We use RadMatch to audit current state-of-the-art report generation models. We score 10 chest X-ray models on the ReXrank~\cite{rexrank} splits of MIMIC-CXR~\cite{mimiccxr}, IU-Xray~\cite{iuxray} and CheXpert Plus~\cite{chexpertplus}, and 13 CT models, from general medical VLMs such as MedGemma~\cite{medgemma15} and RadFM~\cite{radfm} to CT-specialized models such as NV-Reason-CT~\cite{nvreasonct} and HoloCT~\cite{holoct}, on four CT datasets: Merlin-CT~\cite{merlin}, CT-RATE~\cite{ctrate}, CTRG-Chest~\cite{ctrg} and AMOS-MM~\cite{amos,amosmm}. Each model reads its native input, and the findings section serves as reference (\cref{fig:rrg}).
We anchor the count to the report a radiologist would otherwise edit: a fixed, image-blind \emph{normal template}, built from each training split (\cref{app:rrg}) and used for every study. It asserts only normal findings, so a model beats it only by stating more correct significant findings than unsupported ones.

\paragraph{Chest X-ray models leave few errors.} The best chest X-ray models leave few significant errors per report: $0.37$ on IU-Xray, mostly normal studies with short findings ($0.65$ significant reference findings per report), and $2.08$ on MIMIC-CXR, whose emergency and inpatient studies hold more significant findings ($2.9$ per report). The best models recover about half or more of the significant reference findings (sRec $0.49$ on MIMIC-CXR and $0.60$ on IU-Xray, vs.\ at most $0.38$ on CT), and chest X-ray-specific models lead. Several models beat the template, \ie leave statistically fewer significant errors per report: on MIMIC-CXR, three do, from CheXagent-2 3B~\cite{chexagent} ($2.08$ vs.\ $2.53$ for the template) to MAIRA-2~\cite{maira2} ($2.28$); over the three datasets, 6 of the 30 model--dataset runs do.

\paragraph{Most CT models do not beat the template.} On chest CT, only NV-Reason-CT and nnFoundationViT~\cite{nnfoundation} beat the template, both on CT-RATE ($1.67$ and $1.91$ vs.\ $2.11$ significant errors per report); on CTRG-Chest, the best models stay above it. On abdomen CT, no model beats it (nnFoundationViT is level with it on Merlin-CT), and 38 of the 41 CT runs are statistically worse than it, by up to $3.2$ significant errors per report. To beat the template, a model must recover more of the template's misses than it adds unsupported claims; CT models recover few (sRec at most $0.38$, and $0.25$ outside CT-RATE) while adding $0.4$ to $3.2$ unsupported significant claims per report. Some come from priors, not the image: CT-CHAT~\cite{ctrate} asserts atelectasis in $49\%$ of the CT-RATE studies whose reference omits it, nearly as often as when it does ($57\%$; NV-Reason-CT: $11$ vs.\ $47\%$), and Merlin~\cite{merlin} states a hysterectomy in $84\%$ of Merlin-CT studies and $86\%$ of those whose reference describes a prostate, hence a male patient (\cref{app:rrg}).

\begin{figure}[!h]
  \centering
  \includegraphics[width=\linewidth]{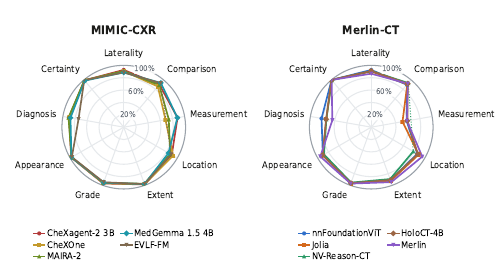}
  \caption{\textbf{Attribute accuracy on MIMIC-CXR and \mbox{Merlin-CT}.} Matched significant abnormal
  findings without a major error (other datasets in \cref{app:rrg}).}
  \label{fig:rrg-radar}
\end{figure}

\paragraph{Which attributes models get wrong.} The attribute grades show how well models describe the findings they report (\cref{fig:rrg-radar}) and differ between modalities: the five best chest X-ray models get the diagnosis right on $74$ to $91\%$ of described significant findings on MIMIC-CXR, the five best CT models on $64$ to $82\%$ on Merlin-CT, where sizes are mostly missing or wrong (NV-Reason-CT states 13 of the 1{,}852 reference sizes it matches; nnFoundationViT states most but gets $58\%$ right). Chest X-ray findings carry fewer attributes, partly explaining their near-ceiling extent and appearance scores. A model builder thus learns what to fix first, and can inspect the finding pairs behind each count in the released dashboard (\cref{app:dashboard}).

\section{Discussion and Limitations}
\label{sec:discussion}

As VLMs and agentic systems write ever more fluent reports, fluency no longer signals correctness: a report can read like a radiologist's yet miss, invent or misstate what matters. Evaluation must then show which findings are wrong and how serious each error is, as RadMatch's finding-level record does. This record also makes RadMatch easy to improve, as a radiologist's correction points to the rule or few-shot example to change (\eg the over-calls of \cref{sec:clinical-validation}), and to extend to a new region or modality with new few-shot examples (\cref{app:impl}). We therefore release RadMatch as a versioned project, whose version and judge should accompany every score.

RadMatch has two practical limits. First, its several LLM calls make it slower than a single-call metric: on two 32\,GB consumer GPUs, a report pair takes on average $7$ to $13$\,s on chest X-ray and $50$\,s on body CT, but $0.6$\,s and $4.0$\,s with 32 parallel workers (\cref{app:cost_latency}). Second, its LLM calls are stochastic, yet over five runs its agreement with radiologists varies by at most $0.017$ (\cref{app:variance}).

We suggest reporting the template's count per benchmark to gauge its difficulty: $0.38$ significant errors per report on IU-Xray vs.\ $5.34$ on Merlin-CT (\cref{app:rrg}). Like every RadMatch count, the template comparison trusts the reference, so an unsupported claim may be a true finding the radiologist omitted; yet on CTRG-Chest and AMOS-MM the most favorable run would need $43$ to $69\%$ of its unsupported claims not to count as errors to tie it (\cref{app:rrg}). Still, leaving fewer errors than a normal template does not yet mean saving time: the next step is a reader study measuring how much reading and editing time a draft saves, and whether RadMatch's record speeds up auditing a model.

\section{Conclusion}
\label{sec:conclusion}
We introduced RadMatch, an LLM-based framework that evaluates radiology reports through finding extraction, finding-level matching and significance-aware grading, turning a score into an auditable record. On four expert benchmarks spanning chest X-ray and body CT, including our MetricEval-BodyCT, its error counts agree with radiologists more closely than previous published metrics, with an open judge run on-premises, and radiologists confirm $93.6\%$ of the errors it counts. When used to evaluate report generation, it shows that several chest X-ray models beat a normal template, but most CT models do not: they add more unsupported significant findings than they correctly report. We release RadMatch and MetricEval-BodyCT so that metrics and report generation models can be checked finding by finding and error by error.

\iftoggle{cvprfinal}{%
\paragraph{Acknowledgements.} We thank Dr.~Denis Habip Gatenyo, Dr.~Alexandre Rolland and Dr.~Julien Franchin for annotating MetricEval-BodyCT, and Segmed for providing its studies. We thank Jonathan Suprijadi, Constantin Ulrich Harsy and the DKFZ team for training and sharing the nnFoundationViT report generators, and our colleagues at Raidium for insightful discussions and feedback throughout this work.
}{}

\section*{LLM Usage Statement}
We used Claude Code as an AI assistant in this research and in writing the paper: it helped draft code, analyses, figures, tables, the clinical-validation web application and parts of the text, and gave feedback on the manuscript. The authors reviewed, edited and approved all of it. The research ideas and conclusions are the authors' own. Claude Code also helped refine RadMatch's prompts and few-shot examples, which radiologists reviewed, and Claude Opus 4.8 rewrote the MetricEval-BodyCT candidates with errors that follow the authors' error plan. The benchmark's ground truth is the radiologists' blinded annotation of these candidates. The LLMs that RadMatch and the baseline metrics call are part of the method. The authors are fully responsible for the entire submission, including its text, code and data.

{
    \small
    \bibliographystyle{ieeenat_fullname}
    \bibliography{main}

\begin{thebibliography}{89}
\providecommand{\natexlab}[1]{#1}
\providecommand{\url}[1]{\texttt{#1}}
\expandafter\ifx\csname urlstyle\endcsname\relax
  \providecommand{\doi}[1]{doi: #1}\else
  \providecommand{\doi}{doi: \begingroup \urlstyle{rm}\Url}\fi

\bibitem[Acosta et~al.(2025)Acosta, Zhang, Dogra, Zhou, Payabvash, Falcone,
  Oermann, and Rajpurkar]{headctone}
Juli{\'a}n~N. Acosta, Xiaoman Zhang, Siddhant Dogra, Hong-Yu Zhou, Seyedmehdi
  Payabvash, Guido~J. Falcone, Eric~K. Oermann, and Pranav Rajpurkar.
\newblock {HeadCT-ONE}: Enabling granular and controllable automated evaluation
  of head {CT} radiology report generation.
\newblock In \emph{Conference on Health, Inference, and Learning (CHIL)}, pages
  649--671, 2025.

\bibitem[{American College of Radiology}(2025)]{acrcommunication}
{American College of Radiology}.
\newblock {ACR} practice parameter for communication of diagnostic imaging
  findings.
\newblock ACR Practice Parameter, Revised 2025 (Resolution 9).
  \url{https://gravitas.acr.org/PPTS/GetDocumentView?docId=74}, 2025.

\bibitem[{Anthropic}(2026{\natexlab{a}})]{claudeopus48card}
{Anthropic}.
\newblock {Claude Opus 4.8} system card.
\newblock
  \url{https://www-cdn.anthropic.com/0b4915911bb0d19eca5b5ee635c80fef830a37ea/Claude\%20Opus\%204.8\%20System\%20Card.pdf},
  2026{\natexlab{a}}.

\bibitem[{Anthropic}(2026{\natexlab{b}})]{claudeopus55}
{Anthropic}.
\newblock Introducing {Claude Opus 5.5}.
\newblock \url{https://www.anthropic.com/claude-opus-5-5}, 2026{\natexlab{b}}.

\bibitem[Baharoon et~al.(2026)Baharoon, Heintz, Raissi, Alabbad, Alhammad,
  AlOmaish, Kim, Banerjee, and Rajpurkar]{crimson}
Mohammed Baharoon, Thibault Heintz, Siavash Raissi, Mahmoud Alabbad, Mona
  Alhammad, Hassan AlOmaish, Sung~Eun Kim, Oishi Banerjee, and Pranav
  Rajpurkar.
\newblock {CRIMSON}: A clinically-grounded {LLM}-based metric for generative
  radiology report evaluation.
\newblock In \emph{Medical Image Computing and Computer Assisted Intervention
  (MICCAI)}, pages 98--107, 2026.

\bibitem[Bai et~al.(2024)Bai, Du, Huang, Meng, and Zhao]{m3d}
Fan Bai, Yuxin Du, Tiejun Huang, Max Q.-H. Meng, and Bo Zhao.
\newblock {M3D}: Advancing {3D} medical image analysis with multi-modal large
  language models.
\newblock \emph{arXiv preprint arXiv:2404.00578}, 2024.

\bibitem[Bai et~al.(2026)Bai, Cheng, Zhou, Zhou, Thirunavukarasu, Ke,
  et~al.]{evlffm}
Yang Bai, Haoran Cheng, Yang Zhou, Jun Zhou, Arun Thirunavukarasu, Yuhe Ke,
  et~al.
\newblock A clinical multimodal vision language foundation model with
  fine-grain explainability.
\newblock \emph{Nature Medicine}, 32\penalty0 (11):\penalty0 4208--4220, 2026.

\bibitem[Banerjee and Lavie(2005)]{meteor}
Satanjeev Banerjee and Alon Lavie.
\newblock {METEOR}: An automatic metric for {MT} evaluation with improved
  correlation with human judgments.
\newblock In \emph{Proceedings of the ACL Workshop on Intrinsic and Extrinsic
  Evaluation Measures for Machine Translation and/or Summarization}, pages
  65--72, 2005.

\bibitem[Bannur et~al.(2024)Bannur, Bouzid, Castro, Schwaighofer, Thieme,
  Bond-Taylor, et~al.]{maira2}
Shruthi Bannur, Kenza Bouzid, Daniel~C. Castro, Anton Schwaighofer, Anja
  Thieme, Sam Bond-Taylor, et~al.
\newblock {MAIRA-2}: Grounded radiology report generation.
\newblock \emph{arXiv preprint arXiv:2406.04449}, 2024.

\bibitem[Blankemeier et~al.(2026)Blankemeier, Kumar, Cohen, Liu, Liu, Van~Veen,
  et~al.]{merlin}
Louis Blankemeier, Ashwin Kumar, Joseph~Paul Cohen, Jiaming Liu, Longchao Liu,
  Dave Van~Veen, et~al.
\newblock {Merlin}: A computed tomography vision-language foundation model and
  dataset.
\newblock \emph{Nature}, 652\penalty0 (8112):\penalty0 1318--1328, 2026.

\bibitem[Chambon et~al.(2024)Chambon, Delbrouck, Sounack, Huang, Chen, Varma,
  Truong, Chuong, and Langlotz]{chexpertplus}
Pierre Chambon, Jean-Benoit Delbrouck, Thomas Sounack, Shih-Cheng Huang,
  Zhihong Chen, Maya Varma, Steven Q.~H. Truong, Chu~The Chuong, and Curtis~P.
  Langlotz.
\newblock {CheXpert Plus}: Augmenting a large chest {X}-ray dataset with text
  radiology reports, patient demographics and additional image formats.
\newblock \emph{arXiv preprint arXiv:2405.19538}, 2024.

\bibitem[Chen et~al.(2026)Chen, Dong, Ding, Shi, Zeng, Zhou, et~al.]{radfabric}
Wenting Chen, Yi Dong, Zhaojun Ding, Yucheng Shi, Fang Zeng, Yifan Zhou, et~al.
\newblock Interpretable agentic {AI} system with localized reasoning for
  radiology.
\newblock \emph{npj Digital Medicine}, 9:\penalty0 862, 2026.

\bibitem[Chen et~al.(2020)Chen, Song, Chang, and Wan]{r2gen}
Zhihong Chen, Yan Song, Tsung-Hui Chang, and Xiang Wan.
\newblock Generating radiology reports via memory-driven transformer.
\newblock In \emph{Proceedings of the 2020 Conference on Empirical Methods in
  Natural Language Processing (EMNLP)}, pages 1439--1449, 2020.

\bibitem[Chen et~al.(2024)Chen, Varma, Xu, Paschali, Van~Veen, Johnston,
  et~al.]{chexagent}
Zhihong Chen, Maya Varma, Justin Xu, Magdalini Paschali, Dave Van~Veen, Andrew
  Johnston, et~al.
\newblock A vision-language foundation model to enhance efficiency of chest
  {X}-ray interpretation.
\newblock \emph{arXiv preprint arXiv:2401.12208}, 2024.

\bibitem[Chen et~al.(2025{\natexlab{a}})Chen, Bie, Jin, and Chen]{reg2rg}
Zhixuan Chen, Yequan Bie, Haibo Jin, and Hao Chen.
\newblock Large language model with region-guided referring and grounding for
  {CT} report generation.
\newblock \emph{IEEE Transactions on Medical Imaging}, 44\penalty0
  (8):\penalty0 3139--3150, 2025{\natexlab{a}}.

\bibitem[Chen et~al.(2025{\natexlab{b}})Chen, Luo, Bie, and Chen]{diallama}
Zhixuan Chen, Luyang Luo, Yequan Bie, and Hao Chen.
\newblock {Dia-LLaMA}: Towards large language model-driven {CT} report
  generation.
\newblock In \emph{Medical Image Computing and Computer Assisted Intervention
  (MICCAI)}, pages 141--151, 2025{\natexlab{b}}.

\bibitem[Cheng and Subramanian(2026)]{topicguided}
Sheng Cheng and Devika Subramanian.
\newblock Rethinking radiology report generation: From narrative flow to
  topic-guided findings.
\newblock In \emph{International Conference on Learning Representations
  (ICLR)}, 2026.

\bibitem[Chinchor and Sundheim(1993)]{muc}
Nancy Chinchor and Beth Sundheim.
\newblock {MUC-5} evaluation metrics.
\newblock In \emph{Fifth Message Understanding Conference (MUC-5)}, 1993.

\bibitem[Dancette et~al.(2025)Dancette, Khlaut, Saporta, Philippe, Ferreres,
  Callard, Danielou, Alberge, et~al.]{curia}
Corentin Dancette, Julien Khlaut, Antoine Saporta, Helene Philippe, Elodie
  Ferreres, Baptiste Callard, Th{\'e}o Danielou, L{\'e}o Alberge, et~al.
\newblock {Curia}: A multi-modal foundation model for radiology.
\newblock \emph{arXiv preprint arXiv:2509.06830}, 2025.

\bibitem[Danielou et~al.(2026)Danielou, Saporta, Alberge, and
  Dancette]{curiamae}
Th{\'e}o Danielou, Antoine Saporta, L{\'e}o Alberge, and Corentin Dancette.
\newblock {Curia-MAE}: Multi-modal multi-anatomy {MAE} pre-training for {3D}
  medical image segmentation.
\newblock In \emph{European Conference on Computer Vision (ECCV) Workshops},
  2026.

\bibitem[Delbrouck et~al.(2025)Delbrouck, Xu, Moll, Thomas, Chen, Ostmeier,
  Azhar, Li, Johnston, Bluethgen, Reis, Muneer, Varma, and Langlotz]{srrbert}
Jean-Benoit Delbrouck, Justin Xu, Johannes Moll, Alois Thomas, Zhihong Chen,
  Sophie Ostmeier, Asfandyar Azhar, Kelvin~Zhenghao Li, Andrew Johnston,
  Christian Bluethgen, Eduardo~Pontes Reis, Mohamed~S. Muneer, Maya Varma, and
  Curtis Langlotz.
\newblock Automated structured radiology report generation.
\newblock In \emph{Proceedings of the 63rd Annual Meeting of the Association
  for Computational Linguistics (ACL)}, pages 26813--26829, 2025.

\bibitem[Demner-Fushman et~al.(2016)Demner-Fushman, Kohli, Rosenman, Shooshan,
  Rodriguez, Antani, Thoma, and McDonald]{iuxray}
Dina Demner-Fushman, Marc~D. Kohli, Marc~B. Rosenman, Sonya~E. Shooshan,
  Laritza Rodriguez, Sameer Antani, George~R. Thoma, and Clement~J. McDonald.
\newblock Preparing a collection of radiology examinations for distribution and
  retrieval.
\newblock \emph{Journal of the American Medical Informatics Association},
  23\penalty0 (2):\penalty0 304--310, 2016.

\bibitem[Fallahpour et~al.(2025)Fallahpour, Ma, Munim, Lyu, and Wang]{medrax}
Adibvafa Fallahpour, Jun Ma, Alif Munim, Hongwei Lyu, and Bo Wang.
\newblock {MedRAX}: Medical reasoning agent for chest {X}-ray.
\newblock In \emph{Proceedings of the 42nd International Conference on Machine
  Learning (ICML)}, pages 15661--15676, 2025.

\bibitem[{Gemma Team}(2026)]{gemma4}
{Gemma Team}.
\newblock {Gemma 4} technical report.
\newblock \emph{arXiv preprint arXiv:2607.02770}, 2026.

\bibitem[Hamamci et~al.(2025{\natexlab{a}})Hamamci, Er, Shit, Reynaud, Kainz,
  and Menze]{crgscore}
Ibrahim~Ethem Hamamci, Sezgin Er, Suprosanna Shit, Hadrien Reynaud, Bernhard
  Kainz, and Bjoern Menze.
\newblock {CRG Score}: A distribution-aware clinical metric for radiology
  report generation.
\newblock In \emph{Medical Imaging with Deep Learning (MIDL), Short Papers},
  2025{\natexlab{a}}.

\bibitem[Hamamci et~al.(2025{\natexlab{b}})Hamamci, Er, Shit, Reynaud, Yang,
  Guo, Edgar, Xu, Kainz, and Menze]{btb3d}
Ibrahim~Ethem Hamamci, Sezgin Er, Suprosanna Shit, Hadrien Reynaud, Dong Yang,
  Pengfei Guo, Marc Edgar, Daguang Xu, Bernhard Kainz, and Bjoern Menze.
\newblock Better tokens for better {3D}: Advancing vision-language modeling in
  {3D} medical imaging.
\newblock In \emph{Advances in Neural Information Processing Systems
  (NeurIPS)}, pages 149770--149798, 2025{\natexlab{b}}.

\bibitem[Hamamci et~al.(2026)Hamamci, Er, Wang, Almas, Simsek, Esirgun,
  et~al.]{ctrate}
Ibrahim~Ethem Hamamci, Sezgin Er, Chenyu Wang, Furkan Almas, Ayse~Gulnihan
  Simsek, Sevval~Nil Esirgun, et~al.
\newblock Generalist foundation models from a multimodal dataset for {3D}
  computed tomography.
\newblock \emph{Nature Biomedical Engineering}, 10\penalty0 (8):\penalty0
  1610--1628, 2026.

\bibitem[Hein et~al.(2025)Hein, Chen, Ostmeier, Xu, Varma, Reis, Michalson,
  Bluethgen, Shin, Langlotz, and Chaudhari]{chexalign}
Dennis Hein, Zhihong Chen, Sophie Ostmeier, Justin Xu, Maya Varma,
  Eduardo~Pontes Reis, Arne~Edward Michalson, Christian Bluethgen, Hyun~Joo
  Shin, Curtis Langlotz, and Akshay~S. Chaudhari.
\newblock {CheXalign}: Preference fine-tuning in chest {X}-ray interpretation
  models without human feedback.
\newblock In \emph{Proceedings of the 63rd Annual Meeting of the Association
  for Computational Linguistics (ACL)}, pages 27679--27702, 2025.

\bibitem[Huang et~al.(2024)Huang, Banerjee, Wu, Reis, and
  Rajpurkar]{fineradscore}
Alyssa Huang, Oishi Banerjee, Kay Wu, Eduardo~Pontes Reis, and Pranav
  Rajpurkar.
\newblock {FineRadScore}: A radiology report line-by-line evaluation technique
  generating corrections with severity scores.
\newblock In \emph{Proceedings of the 9th Machine Learning for Healthcare
  Conference (MLHC)}, 2024.

\bibitem[Hyland et~al.(2023)Hyland, Bannur, Bouzid, Castro, Ranjit,
  Schwaighofer, et~al.]{maira1}
Stephanie~L. Hyland, Shruthi Bannur, Kenza Bouzid, Daniel~C. Castro, Mercy
  Ranjit, Anton Schwaighofer, et~al.
\newblock {MAIRA-1}: A specialised large multimodal model for radiology report
  generation.
\newblock \emph{arXiv preprint arXiv:2311.13668}, 2023.

\bibitem[Jain et~al.(2021)Jain, Agrawal, Saporta, Truong, Duong, Bui, Chambon,
  Zhang, Lungren, Ng, Langlotz, and Rajpurkar]{radgraph}
Saahil Jain, Ashwin Agrawal, Adriel Saporta, Steven Q.~H. Truong, Du~Nguyen
  Duong, Tan Bui, Pierre Chambon, Yuhao Zhang, Matthew~P. Lungren, Andrew~Y.
  Ng, Curtis~P. Langlotz, and Pranav Rajpurkar.
\newblock {R}ad{G}raph: Extracting clinical entities and relations from
  radiology reports.
\newblock In \emph{Advances in Neural Information Processing Systems (NeurIPS),
  Datasets and Benchmarks Track}, 2021.

\bibitem[Ji et~al.(2022)Ji, Bai, Ge, Yang, Zhu, Zhang, Li, Zhang, Ma, Wan, and
  Luo]{amos}
Yuanfeng Ji, Haotian Bai, Chongjian Ge, Jie Yang, Ye Zhu, Ruimao Zhang, Zhen
  Li, Lingyan Zhang, Wanling Ma, Xiang Wan, and Ping Luo.
\newblock {AMOS}: A large-scale abdominal multi-organ benchmark for versatile
  medical image segmentation.
\newblock In \emph{Advances in Neural Information Processing Systems (NeurIPS),
  Datasets and Benchmarks Track}, pages 36722--36732, 2022.

\bibitem[Ji et~al.(2024)Ji, Ge, Li, and Luo]{amosmm}
Yuanfeng Ji, Chongjian Ge, Ruijiang Li, and Ping Luo.
\newblock {AMOS-MM}: Abdominal multimodal analysis challenge: Structured
  description of the challenge design.
\newblock MICCAI 2024 challenge, Zenodo,
  \url{https://doi.org/10.5281/zenodo.10992155}, 2024.

\bibitem[Jiang et~al.(2025)Jiang, Chen, Wang, Li, Tang, Mervak, Chelala,
  Straus, Chahine, Armato~III, and Tan]{clear}
Yuyang Jiang, Chacha Chen, Shengyuan Wang, Feng Li, Zecong Tang, Benjamin~M.
  Mervak, Lydia Chelala, Christopher~M. Straus, Reve Chahine, Samuel~G.
  Armato~III, and Chenhao Tan.
\newblock {CLEAR}: A clinically grounded tabular framework for radiology report
  evaluation.
\newblock In \emph{Findings of the Association for Computational Linguistics:
  EMNLP 2025}, pages 15914--15933, 2025.

\bibitem[Johnson et~al.(2019)Johnson, Pollard, Berkowitz, Greenbaum, Lungren,
  Deng, Mark, and Horng]{mimiccxr}
Alistair E.~W. Johnson, Tom~J. Pollard, Seth~J. Berkowitz, Nathaniel~R.
  Greenbaum, Matthew~P. Lungren, Chih-ying Deng, Roger~G. Mark, and Steven
  Horng.
\newblock {MIMIC-CXR}, a de-identified publicly available database of chest
  radiographs with free-text reports.
\newblock \emph{Scientific Data}, 6\penalty0 (1):\penalty0 317, 2019.

\bibitem[Kang et~al.(2025)Kang, Lee, Jung, Kim, Kim, and Joo]{structuredrrg}
Seongjae Kang, Dong~Bok Lee, Juho Jung, Dongseop Kim, Won~Hwa Kim, and Sunghoon
  Joo.
\newblock Automated structured radiology report generation with rich clinical
  context.
\newblock \emph{arXiv preprint arXiv:2510.00428}, 2025.

\bibitem[Khlaut et~al.(2026)Khlaut, Corbi{\`e}re, Callard, Prat, Butsanets,
  Saporta, Danielou, Machado, et~al.]{jolia}
Julien Khlaut, Charles Corbi{\`e}re, Baptiste Callard, Amaury Prat, Leo
  Butsanets, Antoine Saporta, Th{\'e}o Danielou, Leo Machado, et~al.
\newblock {Jolia}: Concept-level vision-language alignment for {3D} {CT}
  contrastive learning.
\newblock \emph{arXiv preprint arXiv:2606.24570}, 2026.

\bibitem[Larson et~al.(2014)Larson, Berland, Griffith, Kahn, and
  Liebscher]{larson2014}
Paul~A. Larson, Lincoln~L. Berland, Brent Griffith, Charles~E. Kahn, Jr., and
  Lawrence~A. Liebscher.
\newblock Actionable findings and the role of {IT} support: Report of the {ACR
  Actionable Reporting Work Group}.
\newblock \emph{Journal of the American College of Radiology}, 11\penalty0
  (6):\penalty0 552--558, 2014.

\bibitem[Li et~al.(2025{\natexlab{a}})Li, Chang, Yang, Wu, Chen, Bansal, Chen,
  Yang, Chen, Chen, Chen, Lirng, Chang, and Chiou]{braingpt}
Cheng-Yi Li, Kao-Jung Chang, Cheng-Fu Yang, Hsin-Yu Wu, Wenting Chen, Hritik
  Bansal, Ling Chen, Yi-Ping Yang, Yu-Chun Chen, Shih-Pin Chen, Shih-Jen Chen,
  Jiing-Feng Lirng, Kai-Wei Chang, and Shih-Hwa Chiou.
\newblock Towards a holistic framework for multimodal {LLM} in {3D} brain {CT}
  radiology report generation.
\newblock \emph{Nature Communications}, 16:\penalty0 2258, 2025{\natexlab{a}}.

\bibitem[Li et~al.(2025{\natexlab{b}})Li, Liu, Liu, Wang, and Zhou]{radreason}
Yingshu Li, Yunyi Liu, Lingqiao Liu, Lei Wang, and Luping Zhou.
\newblock {RadReason}: Radiology report evaluation metric with reasons and
  sub-scores.
\newblock \emph{arXiv preprint arXiv:2508.15464}, 2025{\natexlab{b}}.

\bibitem[Lin(2004)]{rouge}
Chin-Yew Lin.
\newblock {ROUGE}: A package for automatic evaluation of summaries.
\newblock In \emph{Text Summarization Branches Out (ACL Workshop)}, pages
  74--81, 2004.

\bibitem[Liu et~al.(2026)Liu, Wei, Zheng, Wu, and Wang]{mimickingrad}
Hong Liu, Dong Wei, Yefeng Zheng, Xian Wu, and Liansheng Wang.
\newblock Mimicking radiologists: A coarse-to-fine framework with structural
  sparse tokens for dual-{LLM} computed tomography report generation.
\newblock In \emph{European Conference on Computer Vision (ECCV)}, pages
  376--393, 2026.

\bibitem[Liu et~al.(2024)Liu, Wang, Li, Liang, Liu, Wang, and Zhou]{mrscore}
Yunyi Liu, Zhanyu Wang, Yingshu Li, Xinyu Liang, Lingqiao Liu, Lei Wang, and
  Luping Zhou.
\newblock {MRScore}: Evaluating medical report with {LLM}-based reward system.
\newblock In \emph{Medical Image Computing and Computer Assisted Intervention
  (MICCAI)}, pages 283--292, 2024.

\bibitem[MacMahon et~al.(2017)MacMahon, Naidich, Goo, Lee, Leung, Mayo,
  et~al.]{fleischner}
Heber MacMahon, David~P. Naidich, Jin~Mo Goo, Kyung~Soo Lee, Ann N.~C. Leung,
  John~R. Mayo, et~al.
\newblock Guidelines for management of incidental pulmonary nodules detected on
  {CT} images: From the {Fleischner Society} 2017.
\newblock \emph{Radiology}, 284\penalty0 (1):\penalty0 228--243, 2017.

\bibitem[Maye-Lasserre et~al.(2026)Maye-Lasserre, Li, Jian, Ghahremani,
  Wiestler, and Wachinger]{templatecollapse}
Tom Maye-Lasserre, Yitong Li, Bailiang Jian, Morteza Ghahremani, Benedikt
  Wiestler, and Christian Wachinger.
\newblock Generating reports or repeating templates? {M}easuring and mitigating
  template collapse in {3D} {CT} report generation.
\newblock In \emph{Proceedings of the 2026 Conference on Empirical Methods in
  Natural Language Processing (EMNLP)}, pages 48270--48286, 2026.

\bibitem[Min et~al.(2023)Min, Krishna, Lyu, Lewis, Yih, Koh, Iyyer,
  Zettlemoyer, and Hajishirzi]{factscore}
Sewon Min, Kalpesh Krishna, Xinxi Lyu, Mike Lewis, Wen-tau Yih, Pang~Wei Koh,
  Mohit Iyyer, Luke Zettlemoyer, and Hannaneh Hajishirzi.
\newblock {FActScore}: Fine-grained atomic evaluation of factual precision in
  long form text generation.
\newblock In \emph{Proceedings of the 2023 Conference on Empirical Methods in
  Natural Language Processing (EMNLP)}, pages 12076--12100, 2023.

\bibitem[Myronenko et~al.(2026)Myronenko, Yang, Tang, Turkbey, Simon, Harmon,
  et~al.]{nvreasonct}
Andriy Myronenko, Dong Yang, Yucheng Tang, Baris Turkbey, Benjamin Simon,
  Stephanie Harmon, et~al.
\newblock {NV-Reason-CT}: {3D} visual language model for {CT} analysis.
\newblock \emph{arXiv preprint arXiv:2609.27511}, 2026.

\bibitem[{OpenAI}(2025{\natexlab{a}})]{gpt41}
{OpenAI}.
\newblock Introducing {GPT-4.1} in the {API}.
\newblock \url{https://openai.com/index/gpt-4-1/}, 2025{\natexlab{a}}.

\bibitem[{OpenAI}(2025{\natexlab{b}})]{gpt52}
{OpenAI}.
\newblock Introducing {GPT-5.2}.
\newblock \url{https://openai.com/index/introducing-gpt-5-2/},
  2025{\natexlab{b}}.

\bibitem[{OpenAI}(2026)]{gpt6astra}
{OpenAI}.
\newblock {GPT-6} {Astra}: A new generation of intelligence.
\newblock \url{https://openai.com/index/gpt-6-astra/}, 2026.

\bibitem[Ostmeier et~al.(2024)Ostmeier, Xu, Chen, Varma, Blankemeier,
  Bluethgen, Michalson, Moseley, Langlotz, Chaudhari, and Delbrouck]{green}
Sophie Ostmeier, Justin Xu, Zhihong Chen, Maya Varma, Louis Blankemeier,
  Christian Bluethgen, Arne~Edward Michalson, Michael Moseley, Curtis Langlotz,
  Akshay~S. Chaudhari, and Jean-Benoit Delbrouck.
\newblock {GREEN}: Generative radiology report evaluation and error notation.
\newblock In \emph{Findings of the Association for Computational Linguistics:
  EMNLP 2024}, pages 374--390, 2024.

\bibitem[Papineni et~al.(2002)Papineni, Roukos, Ward, and Zhu]{bleu}
Kishore Papineni, Salim Roukos, Todd Ward, and Wei-Jing Zhu.
\newblock {BLEU}: A method for automatic evaluation of machine translation.
\newblock In \emph{Proceedings of the 40th Annual Meeting of the Association
  for Computational Linguistics (ACL)}, pages 311--318, 2002.

\bibitem[Park et~al.(2025)Park, Heo, Shin, Son, Oh, and Kam]{dart}
Sang-Jun Park, Keun-Soo Heo, Dong-Hee Shin, Young-Han Son, Ji-Hye Oh, and
  Tae-Eui Kam.
\newblock {DART}: Disease-aware image-text alignment and self-correcting
  re-alignment for trustworthy radiology report generation.
\newblock In \emph{IEEE/CVF Conference on Computer Vision and Pattern
  Recognition (CVPR)}, pages 15580--15589, 2025.

\bibitem[Petryk et~al.(2024)Petryk, Chan, Kachinthaya, Zou, Canny, Gonzalez,
  and Darrell]{aloha}
Suzanne Petryk, David~M. Chan, Anish Kachinthaya, Haodi Zou, John Canny,
  Joseph~E. Gonzalez, and Trevor Darrell.
\newblock {ALOHa}: A new measure for hallucination in captioning models.
\newblock In \emph{Proceedings of the 2024 Conference of the North American
  Chapter of the Association for Computational Linguistics: Human Language
  Technologies (NAACL), Volume 2: Short Papers}, pages 342--357, 2024.

\bibitem[Qi et~al.(2026)Qi, Bu, Xiang, and Dai]{eviagent}
Tuoshi Qi, Shenshen Bu, Yingfei Xiang, and Zhiming Dai.
\newblock {EviAgent}: Evidence-driven agent for radiology report generation.
\newblock \emph{arXiv preprint arXiv:2603.13956}, 2026.

\bibitem[{Qwen Team}(2026{\natexlab{a}})]{qwen35}
{Qwen Team}.
\newblock {Qwen3.5}: Towards native multimodal agents.
\newblock \url{https://qwen.ai/blog?id=qwen3.5}, 2026{\natexlab{a}}.

\bibitem[{Qwen Team}(2026{\natexlab{b}})]{qwen38}
{Qwen Team}.
\newblock {Qwen3.8-27B}.
\newblock \url{https://huggingface.co/Qwen/Qwen3.8-27B}, 2026{\natexlab{b}}.

\bibitem[Saporta et~al.(2026)Saporta, Callard, Dancette, Khlaut, Corbi{\`e}re,
  Butsanets, Prat, and Manceron]{curia2}
Antoine Saporta, Baptiste Callard, Corentin Dancette, Julien Khlaut, Charles
  Corbi{\`e}re, Leo Butsanets, Amaury Prat, and Pierre Manceron.
\newblock {Curia-2}: Scaling self-supervised learning for radiology foundation
  models.
\newblock In \emph{Medical Image Computing and Computer Assisted Intervention
  (MICCAI)}, pages 144--154, 2026.

\bibitem[Sellergren et~al.(2026{\natexlab{a}})Sellergren, Gao, Mahvar,
  Kohlberger, et~al.]{medgemma15}
Andrew Sellergren, Chufan Gao, Fereshteh Mahvar, Timo Kohlberger, et~al.
\newblock {MedGemma 1.5} technical report.
\newblock \emph{arXiv preprint arXiv:2604.05081}, 2026{\natexlab{a}}.

\bibitem[Sellergren et~al.(2026{\natexlab{b}})Sellergren, Kazemzadeh, Mahvar,
  Kiraly, Traverse, Kohlberger, Gao, Jaroensri, et~al.]{medgemma}
Andrew Sellergren, Sahar Kazemzadeh, Fereshteh Mahvar, Atilla Kiraly, Madeleine
  Traverse, Timo Kohlberger, Chufan Gao, Ronnachai Jaroensri, et~al.
\newblock An open vision-language model for diverse medical applications.
\newblock \emph{Nature Medicine}, 32\penalty0 (10):\penalty0 3695--3704,
  2026{\natexlab{b}}.

\bibitem[Sharma et~al.(2025)Sharma, Salvatelli, Srivastav, Bouzid, Bannur,
  Castro, et~al.]{mairaseg}
Harshita Sharma, Valentina Salvatelli, Shaury Srivastav, Kenza Bouzid, Shruthi
  Bannur, Daniel~C. Castro, et~al.
\newblock {MAIRA-Seg}: Enhancing radiology report generation with
  segmentation-aware multimodal large language models.
\newblock In \emph{Proceedings of the 4th Machine Learning for Health Symposium
  (ML4H)}, pages 941--960, 2025.

\bibitem[Sharma et~al.(2026)Sharma, Bejar, Durak, and Bagci]{ctestmetric}
Vanshali Sharma, Andrea~Mia Bejar, Gorkem Durak, and Ulas Bagci.
\newblock {CTest-Metric}: A unified framework to assess clinical validity of
  metrics for {CT} report generation.
\newblock In \emph{2026 IEEE 23rd International Symposium on Biomedical Imaging
  (ISBI)}, pages 1--5, 2026.

\bibitem[Smit et~al.(2020)Smit, Jain, Rajpurkar, Pareek, Ng, and
  Lungren]{chexbert}
Akshay Smit, Saahil Jain, Pranav Rajpurkar, Anuj Pareek, Andrew~Y. Ng, and
  Matthew~P. Lungren.
\newblock Combining automatic labelers and expert annotations for accurate
  radiology report labeling using {BERT}.
\newblock In \emph{Proceedings of the 2020 Conference on Empirical Methods in
  Natural Language Processing (EMNLP)}, pages 1500--1519, 2020.

\bibitem[Srivastava et~al.(2026)Srivastava, Bhosale, Doermann, and Gao]{cwcd}
Shantam Srivastava, Mahesh Bhosale, David Doermann, and Mingchen Gao.
\newblock {CWCD}: Category-wise contrastive decoding for structured medical
  report generation.
\newblock In \emph{Proceedings of The 9th International Conference on Medical
  Imaging with Deep Learning (MIDL)}, pages 868--893, 2026.

\bibitem[Sun et~al.(2025)Sun, Zhao, Han, and Xiong]{factmmrag}
Liwen Sun, James~Jialun Zhao, Wenjing Han, and Chenyan Xiong.
\newblock Fact-aware multimodal retrieval augmentation for accurate medical
  radiology report generation.
\newblock In \emph{Proceedings of the 2025 Conference of the Nations of the
  Americas Chapter of the Association for Computational Linguistics: Human
  Language Technologies (NAACL)}, pages 643--655, 2025.

\bibitem[Suprijadi et~al.(2026)Suprijadi, Stock, Langenberg, Zimmerer, Kahl,
  Denner, et~al.]{resolutionreduction}
Jonathan Suprijadi, Raphael Stock, Moritz Langenberg, David Zimmerer,
  Kim-Celine Kahl, Stefan Denner, et~al.
\newblock Resolution meets reduction: Efficient visual context for {3D}
  radiology report generation.
\newblock \emph{arXiv preprint arXiv:2608.08713}, 2026.

\bibitem[Tang et~al.(2024)Tang, Yang, Zhang, and Yuan]{ctrg}
Yuhao Tang, Haichen Yang, Liyan Zhang, and Ye Yuan.
\newblock Work like a doctor: Unifying scan localizer and dynamic generator for
  automated computed tomography report generation.
\newblock \emph{Expert Systems with Applications}, 237:\penalty0 121442, 2024.

\bibitem[Tanida et~al.(2023)Tanida, M{\"u}ller, Kaissis, and Rueckert]{rgrg}
Tim Tanida, Philip M{\"u}ller, Georgios Kaissis, and Daniel Rueckert.
\newblock Interactive and explainable region-guided radiology report
  generation.
\newblock In \emph{IEEE/CVF Conference on Computer Vision and Pattern
  Recognition (CVPR)}, pages 7433--7442, 2023.

\bibitem[Tanno et~al.(2025)Tanno, Barrett, Sellergren, Ghaisas, Dathathri, See,
  et~al.]{flamingocxr}
Ryutaro Tanno, David G.~T. Barrett, Andrew Sellergren, Sumedh Ghaisas, Sumanth
  Dathathri, Abigail See, et~al.
\newblock Collaboration between clinicians and vision--language models in
  radiology report generation.
\newblock \emph{Nature Medicine}, 31\penalty0 (2):\penalty0 599--608, 2025.

\bibitem[Thawakar et~al.(2024)Thawakar, Shaker, Mullappilly, Cholakkal, Anwer,
  Khan, Laaksonen, and Khan]{xraygpt}
Omkar Thawakar, Abdelrahman Shaker, Sahal~Shaji Mullappilly, Hisham Cholakkal,
  Rao~Muhammad Anwer, Salman Khan, Jorma Laaksonen, and Fahad~Shahbaz Khan.
\newblock {XrayGPT}: Chest radiographs summarization using large medical
  vision-language models.
\newblock In \emph{Proceedings of the 23rd Workshop on Biomedical Natural
  Language Processing (BioNLP @ ACL)}, pages 440--448, 2024.

\bibitem[Ulrich~Harsy et~al.(2026)Ulrich~Harsy, Wald, Gotkowski, Kirchhoff,
  Knopp, Rokuss, et~al.]{nnfoundation}
Constantin Ulrich~Harsy, Tassilo Wald, Karol Gotkowski, Yannick Kirchhoff,
  Marcel Knopp, Maximilian Rokuss, et~al.
\newblock {nnFoundation}: {3D} foundation models for radiology.
\newblock \emph{arXiv preprint arXiv:2609.26924}, 2026.

\bibitem[Wald et~al.(2026)Wald, Hamamci, Gao, Bond-Taylor, Sharma, Ilse, Lo,
  Melnichenko, Schwaighofer, Codella, Wetscherek, Maier-Hein, Korfiatis,
  Salvatelli, Alvarez-Valle, and P{\'e}rez-Garc{\'i}a]{colipri}
Tassilo Wald, Ibrahim~Ethem Hamamci, Yuan Gao, Sam Bond-Taylor, Harshita
  Sharma, Maximilian Ilse, Cynthia Lo, Olesya Melnichenko, Anton Schwaighofer,
  Noel C.~F. Codella, Maria~Teodora Wetscherek, Klaus~H. Maier-Hein, Panagiotis
  Korfiatis, Valentina Salvatelli, Javier Alvarez-Valle, and Fernando
  P{\'e}rez-Garc{\'i}a.
\newblock Comprehensive language-image pre-training for {3D} medical image
  understanding.
\newblock In \emph{European Conference on Computer Vision (ECCV)}, pages
  388--406, 2026.

\bibitem[Wang et~al.(2024)Wang, Luo, Jiang, Li, and Qiu]{llmradjudge}
Zilong Wang, Xufang Luo, Xinyang Jiang, Dongsheng Li, and Lili Qiu.
\newblock {LLM-RadJudge}: Achieving radiologist-level evaluation for {X}-ray
  report generation.
\newblock \emph{arXiv preprint arXiv:2404.00998}, 2024.

\bibitem[Wu et~al.(2025)Wu, Zhang, Zhang, Hui, Wang, and Xie]{radfm}
Chaoyi Wu, Xiaoman Zhang, Ya Zhang, Hui Hui, Yanfeng Wang, and Weidi Xie.
\newblock Towards generalist foundation model for radiology by leveraging
  web-scale {2D}\&{3D} medical data.
\newblock \emph{Nature Communications}, 16:\penalty0 7866, 2025.

\bibitem[Xie et~al.(2026)Xie, Li, Xu, Bai, Shi, Chen, et~al.]{holoct}
Yulu Xie, Zilong Li, Shen Xu, Yifan Bai, Liuhong Shi, Jiao Chen, et~al.
\newblock A {3D} foundation model unifying {CT} interpretation and
  segmentation.
\newblock \emph{arXiv preprint arXiv:2611.06471}, 2026.

\bibitem[Xin et~al.(2026)Xin, Ates, Gong, and Shao]{med3dvlm}
Yu Xin, Gorkem~Can Ates, Kuang Gong, and Wei Shao.
\newblock {Med3DVLM}: An efficient vision-language model for {3D} medical image
  analysis.
\newblock \emph{IEEE Journal of Biomedical and Health Informatics}, 30\penalty0
  (3):\penalty0 2524--2536, 2026.

\bibitem[Xu et~al.(2025)Xu, Zhang, Abderezaei, Bauml, Boodoo, Haghighi,
  Ganjizadeh, Brattain, Van~Veen, Meng, Eyre, and Delbrouck]{radeval}
Justin Xu, Xi Zhang, Javid Abderezaei, Julie Bauml, Roger Boodoo, Fatemeh
  Haghighi, Ali Ganjizadeh, Eric Brattain, Dave Van~Veen, Zaiqiao Meng,
  David~W. Eyre, and Jean-Benoit Delbrouck.
\newblock {RadEval}: A framework for radiology text evaluation.
\newblock In \emph{Proceedings of the 2025 Conference on Empirical Methods in
  Natural Language Processing: System Demonstrations}, pages 546--557, 2025.

\bibitem[Xu et~al.(2026)Xu, Chan, Li, Aljunied, Yuan, Wang, et~al.]{lingshu}
Weiwen Xu, Hou~Pong Chan, Long Li, Mahani Aljunied, Ruifeng Yuan, Jianyu Wang,
  et~al.
\newblock {Lingshu}: Generalist foundation model for unified multimodal medical
  understanding and reasoning.
\newblock \emph{IEEE Transactions on Pattern Analysis and Machine
  Intelligence}, 2026.
\newblock Early access.

\bibitem[Yan et~al.(2026)Yan, Wang, Yin, Cheung, Cheung, and
  See]{selfcritiquing}
Sixing Yan, Ziao Wang, Kejing Yin, William~K. Cheung, Ka~Chun Cheung, and Simon
  See.
\newblock Learning self-critiquing mechanisms for region-guided chest {X}-ray
  report generation.
\newblock In \emph{International Conference on Learning Representations
  (ICLR)}, 2026.

\bibitem[Ye et~al.(2025)Ye, Zeng, Li, Li, and Fan]{dcscore}
Qinghao Ye, Xianhan Zeng, Fu Li, Chunyuan Li, and Haoqi Fan.
\newblock Painting with words: Elevating detailed image captioning with
  benchmark and alignment learning.
\newblock In \emph{International Conference on Learning Representations
  (ICLR)}, 2025.

\bibitem[Yi et~al.(2026)Yi, Liu, Xiao, and Albert]{mmagentrrg}
Ziruo Yi, Jinyu Liu, Ting Xiao, and Mark~V. Albert.
\newblock A multimodal multi-agent framework for radiology report generation.
\newblock In \emph{Internet Computing, Internet of Things, Artificial
  Intelligence, and Applications (CSCE 2025)}, pages 147--161. Springer, 2026.

\bibitem[Yildirim et~al.(2024)Yildirim, Richardson, Wetscherek, Bajwa, Jacob,
  Pinnock, Harris, Coelho~de Castro, Bannur, Hyland, Ghosh, Ranjit, Bouzid,
  Schwaighofer, P{\'e}rez-Garc{\'i}a, Sharma, Oktay, Lungren, Alvarez-Valle,
  Nori, and Thieme]{yildirim}
Nur Yildirim, Hannah Richardson, Maria~T. Wetscherek, Junaid Bajwa, Joseph
  Jacob, Mark~A. Pinnock, Stephen Harris, Daniel Coelho~de Castro, Shruthi
  Bannur, Stephanie~L. Hyland, Pratik Ghosh, Mercy Ranjit, Kenza Bouzid, Anton
  Schwaighofer, Fernando P{\'e}rez-Garc{\'i}a, Harshita Sharma, Ozan Oktay,
  Matthew Lungren, Javier Alvarez-Valle, Aditya Nori, and Anja Thieme.
\newblock Multimodal healthcare {AI}: Identifying and designing clinically
  relevant vision-language applications for radiology.
\newblock In \emph{Proceedings of the 2024 CHI Conference on Human Factors in
  Computing Systems (CHI)}, 2024.

\bibitem[Yu et~al.(2023)Yu, Endo, Krishnan, Pan, Tsai, Reis, Fonseca, Lee,
  Abad, Ng, Langlotz, Venugopal, and Rajpurkar]{rexval}
Feiyang Yu, Mark Endo, Rayan Krishnan, Ian Pan, Andy Tsai, Eduardo~Pontes Reis,
  Eduardo Kaiser Ururahy~Nunes Fonseca, Henrique Min~Ho Lee, Zahra
  Shakeri~Hossein Abad, Andrew~Y. Ng, Curtis~P. Langlotz, Vasantha~Kumar
  Venugopal, and Pranav Rajpurkar.
\newblock Evaluating progress in automatic chest {X}-ray radiology report
  generation.
\newblock \emph{Patterns}, 4\penalty0 (9):\penalty0 100802, 2023.

\bibitem[Yuan et~al.(2026)Yuan, Chang, Cao, Shi, Wei, Zhang, and
  Zhang]{ctfinebench}
Ruifeng Yuan, Wanxing Chang, Weiwei Cao, Bowen Shi, Zhongyu Wei, Ling Zhang,
  and Jianpeng Zhang.
\newblock {CT-FineBench}: A diagnostic fidelity benchmark for fine-grained
  evaluation of {CT} report generation.
\newblock In \emph{Proceedings of the 64th Annual Meeting of the Association
  for Computational Linguistics (ACL)}, pages 44045--44058, 2026.

\bibitem[Zambrano~Chaves et~al.(2025)Zambrano~Chaves, Huang, Xu, Xu, Usuyama,
  Zhang, et~al.]{chexprompt}
Juan~Manuel Zambrano~Chaves, Shih-Cheng Huang, Yanbo Xu, Hanwen Xu, Naoto
  Usuyama, Sheng Zhang, et~al.
\newblock A clinically accessible small multimodal radiology model and
  evaluation metric for chest {X}-ray findings.
\newblock \emph{Nature Communications}, 16:\penalty0 3108, 2025.

\bibitem[Zhang et~al.(2020)Zhang, Kishore, Wu, Weinberger, and
  Artzi]{bertscore}
Tianyi Zhang, Varsha Kishore, Felix Wu, Kilian~Q. Weinberger, and Yoav Artzi.
\newblock {BERTScore}: Evaluating text generation with {BERT}.
\newblock In \emph{International Conference on Learning Representations
  (ICLR)}, 2020.

\bibitem[Zhang et~al.(2025)Zhang, Zhou, Yang, Banerjee, Acosta, Miller, Huang,
  and Rajpurkar]{rexrank}
Xiaoman Zhang, Hong-Yu Zhou, Xiaoli Yang, Oishi Banerjee, Juli{\'a}n~N. Acosta,
  Josh Miller, Ouwen Huang, and Pranav Rajpurkar.
\newblock {ReXrank}: A public leaderboard for {AI}-powered radiology report
  generation.
\newblock In \emph{Proceedings of the First AAAI Bridge Program on AI for
  Medicine and Healthcare}, pages 90--99, 2025.

\bibitem[Zhang et~al.(2026{\natexlab{a}})Zhang, Wang, Gao, Liu, Varma, Xu,
  et~al.]{chexone}
Yabin Zhang, Chong Wang, Yunhe Gao, Jiaming Liu, Maya Varma, Justin Xu, et~al.
\newblock A reasoning-enabled vision-language foundation model for chest
  {X}-ray interpretation.
\newblock \emph{arXiv preprint arXiv:2604.00493}, 2026{\natexlab{a}}.

\bibitem[Zhang et~al.(2026{\natexlab{b}})Zhang, Lee, Jing, Deng, Zhou, Jin,
  Huang, Gao, Marshall, Fang, and Yang]{gemascore}
Zhenxuan Zhang, KinHei Lee, Peiyuan Jing, Weihang Deng, Huichi Zhou, Zihao Jin,
  Jiahao Huang, Zhifan Gao, Dominic~C. Marshall, Yingying Fang, and Guang Yang.
\newblock {GEMA-Score}: Granular explainable multi-agent scoring framework for
  radiology report evaluation.
\newblock In \emph{Proceedings of the AAAI Conference on Artificial
  Intelligence (AAAI)}, pages 13025--13033, 2026{\natexlab{b}}.

\end{thebibliography}
}

\twocolumn[\centering\Large\bfseries Appendix\par\vspace{1.5em}]
\setcounter{table}{0}\setcounter{figure}{0}
\renewcommand{\thetable}{S\arabic{table}}\renewcommand{\thefigure}{S\arabic{figure}}
\raggedbottom
\widowpenalty=10000 \clubpenalty=10000  %
\renewcommand{\topfraction}{0.9}\renewcommand{\dbltopfraction}{0.9}\renewcommand{\textfraction}{0.07}
\renewcommand{\floatpagefraction}{0.8}\renewcommand{\dblfloatpagefraction}{0.8}
\appendix
\crefalias{section}{appendix}\crefalias{subsection}{subappendix}

\cref{app:impl} details the RadMatch pipeline, its structured comparators and its runtime. \cref{app:bodyct} describes how MetricEval-BodyCT was built and annotated and gives its full results. \cref{app:agreement} lists the values of \cref{fig:agreement} and breaks agreement down on ReXVal, RadEvalExpert and RadPref. \cref{app:clinical-validation} reports the radiologists' review of RadMatch's decisions, qualitative audits and the RadMatch dashboard. \cref{app:ablations} covers run-to-run variance, judge score levels, the score definition, the single-call baseline and padding. \cref{app:rrg} gives the setup and full results of the report generation benchmark, and \cref{app:prompts} the verbatim prompts. Unless a judge is named, results use Gemma 4 31B. Differences quoted in the text are computed before rounding, so they can differ by $0.01$ from those of the rounded table values.

\section{Implementation Details}
\label{app:impl}
All results in the paper and this appendix use RadMatch 0.3.9, the released version.

\subsection{Pipeline}
Findings are extracted from the reference and the candidate independently (two LLM calls), a single call matches the two
finding sets, and batched calls of up to ten matched pairs grade their free-text attributes (at most one call for almost every
chest X-ray report pair, none when the rules fix the outcome of every pair). When a study indication is available, an optional
preliminary call extracts it and passes it as clinical context to the three downstream prompts. All other work, the structured
comparators, the verdict assignment and every aggregate score, is deterministic Python. The stored record, from which every
score can be recomputed, is described in \cref{sec:scoring}. API judges run at their provider's default reasoning setting, and open judges are served locally with thinking disabled and $T{=}0$. RadMatch ships eight few-shot bundles. Five (chest X-ray, chest CT, abdomen CT, head CT and brain MRI) hold three de-identified report pairs each with their stage outputs, further reference reports for extraction, some with sentences added to illustrate the significance rules, and constructed matching and grading examples for rarer patterns; the other three (abdominal ultrasound, mammography, musculoskeletal radiographs) are written from scratch. None overlaps the metric-evaluation benchmarks.

\subsection{Consistency Checks and Fallbacks}
Each LLM output is parsed against a JSON schema, and the matching is checked for bookkeeping, not semantics: every reference and candidate finding must be accounted for, in one or more matches or as unmatched, with no invented identifier. On failure the call is repeated with the list of violations, up to two retries. If the matching still fails, RadMatch keeps the individually valid matches and treats every unaccounted finding as unmatched (a \emph{fallback}). \cref{sec:robustness} reports MedGemma 1.5 4B's frequent fallbacks; GPT-4.1 and Qwen3.5-35B-A3B fall back on about $5\%$ of the MetricEval-BodyCT pairs and Gemma 4 31B on one ReXVal pair.

\subsection{Structured Comparators}
Status, laterality and comparison are compared deterministically, and measurements partly so.
Status disagreement is a major error (a status inversion) and the pair's only error. A laterality difference between two abnormal findings (left vs.\ right, bilateral vs.\ unilateral, a laterality the reference gives and the candidate leaves out, except against a bilateral reference) is major unless the findings of each report in the match together have the same laterality; a laterality only the candidate gives is not an error, and between two \textsc{routine} findings a difference is minor. A longitudinal-comparison difference is
major when it crosses between the benign (stable, improving, resolved)
and active (worsening, new) trajectories and minor otherwise, including a comparison only one report gives; the LLM grades a new or growing change that only one report gives, major on a finding tracked by its size (\eg a nodule, an aneurysm, an effusion). Counts, ratios and attenuation are compared per category: any change in a count, a ratio
difference above $30\%$, or an attenuation difference above $20$\,HU or one
that crosses a tissue-characterization boundary ($-10$, $10$ or $20$\,HU, separating macroscopic
fat, lipid-rich adenoma and simple fluid) is major. Sizes are flagged when they differ by more than $20\%$ and at least $2$\,mm, or when the candidate leaves one out, and the LLM grades the flag. A flagged size is major when the difference crosses a decision threshold of the structure (\eg the Fleischner~\cite{fleischner} $6$ and $8$\,mm nodule thresholds, a $10$\,mm nodal short axis) or exceeds $20\%$ and $2$\,mm on a lesion tracked by its size, or when the candidate leaves out the size of a lesion tracked or staged by it without a covering size class (\eg ``subcentimeter''), and minor otherwise. A difference within measurement variability and a size only the candidate gives are not errors.

\subsection{Runtime and Energy}
\label{app:cost_latency}
\cref{tab:cost} times the full pipeline with one worker, with Gemma 4 31B served with vLLM in FP8 on two 32\,GB consumer GPUs (RTX~5090). A chest X-ray pair takes $7$--$13$\,s, $25$--$27$k prompt and $0.5$--$0.8$k completion tokens over three to four calls, and $1.6$--$2.7$\,Wh; a body CT pair, with much longer reports, takes $50$\,s, $53$k and $3.1$k tokens and $11$\,Wh. The shared system prompts and few-shot exemplars dominate the prompt and are reused by the server's prefix cache ($89$--$95\%$ of prompt tokens). As pairs are independent, concurrency amortizes this latency: with 32 workers, a pair costs $0.58$\,s of wall-clock on ReXVal and $4.0$\,s on body CT (measured on 100 and 60 pairs), so ReXVal is scored in under two minutes and the 300 pairs of MetricEval-BodyCT in about 20 minutes, while the latency of an individual pair rises under load. Gemma's thinking mode does not improve agreement on ReXVal ($+0.004$ [$-0.035$, $0.046$] against thinking off, both sampled at $T{=}1$) but multiplies completion tokens by $9$ and wall-clock by $7.8$.

\begin{table}[t]
  \centering
  \caption{\textbf{Runtime and energy of RadMatch with the open Gemma 4 31B judge.} vLLM in FP8 on $2\times$RTX~5090, one worker, full pipeline on 25 pairs per benchmark. Per pair: mean wall-clock, split into the extraction of both reports, matching and attribute grading; prompt and completion tokens, with the prefix-cached share of the prompt; GPU time and energy of the two GPUs.}
  \label{tab:cost}
  \scriptsize
  \setlength{\tabcolsep}{2.5pt}
  \begin{tabular}{lccccc}
    \toprule
    Benchmark & s/pair & \begin{tabular}[b]{@{}c@{}}extract / match /\\grade (s)\end{tabular} & \begin{tabular}[b]{@{}c@{}}tokens in / out\\(prefix-cached)\end{tabular} & GPU-s & Wh \\
    \midrule
    ReXVal        & 7.2  & 4.7 / 1.7 / 0.8   & 27.0k / 0.5k (95\%) & 14  & 1.6 \\
    RadEvalExpert & 10.6 & 7.4 / 2.2 / 1.0   & 25.0k / 0.7k (94\%) & 21  & 2.3 \\
    RadPref       & 12.6 & 8.8 / 3.0 / 0.9   & 26.2k / 0.8k (94\%) & 25  & 2.7 \\
    MetricEval-BodyCT & 49.7 & 35.1 / 11.8 / 2.8 & 52.7k / 3.1k (89\%) & 99 & 10.6 \\
    \bottomrule
  \end{tabular}
\end{table}

\section{MetricEval-BodyCT: Construction, Annotation and Full Results}
\label{app:bodyct}
\begin{table*}[t]
  \centering
  \caption{\textbf{Agreement with the radiologists on MetricEval-BodyCT, per endpoint and region.} Kendall's $\tau_b$ (95\% CI, study-clustered bootstrap; higher is better), pooled over the candidates and averaged over the two radiologists. \emph{Total} and \emph{significant errors}: the radiologists' error counts; \emph{fidelity} and \emph{indication rank}: their ranking of a study's three candidates by fidelity to the reference and by how well the impression answers the clinical indication; \emph{by region}: total errors on the 50 chest and the 50 abdomen--pelvis studies. Inter-expert: agreement between the two radiologists. RadMatch is scored with sER on significant errors, the primary endpoint, and with aER elsewhere; RadFact is the F1 of RadEval's RadFact-CT. Bold: best metric per column.}
  \label{tab:bodyct}
  \scriptsize
  \setlength{\tabcolsep}{4pt}
  \providecommand{\valwidths}[2]{}\valwidths{$-$0.00}{[$-$.00,.00]}
  \begin{tabular}{lcccccc}
    \toprule
    & \multicolumn{4}{c}{All 100 studies} & \multicolumn{2}{c}{Total errors by region} \\ \cmidrule(lr){2-5}\cmidrule(lr){6-7}
    Metric & \begin{tabular}[b]{@{}c@{}}Total\\errors\end{tabular} & \begin{tabular}[b]{@{}c@{}}Significant\\errors\end{tabular} & \begin{tabular}[b]{@{}c@{}}Fidelity\\rank\end{tabular} & \begin{tabular}[b]{@{}c@{}}Indication\\rank\end{tabular} & \begin{tabular}[b]{@{}c@{}}Chest\\CT\end{tabular} & \begin{tabular}[b]{@{}c@{}}Abdomen\\CT\end{tabular} \\
    \midrule
    \textit{Inter-expert} & \val{0.69}{.63}{.74} & \val{0.40}{.31}{.49} & \val{0.65}{.54}{.75} & \val{0.67}{.58}{.74} & \val{0.65}{.58}{.72} & \val{0.73}{.63}{.82} \\
    \midrule
    {\color{rmgreen}RadMatch} & \val{\textbf{0.53}}{.45}{.59} & \val{\textbf{0.45}}{.40}{.50} & \val{\textbf{0.51}}{.46}{.57} & \val{0.38}{.31}{.44} & \val{\textbf{0.56}}{.51}{.62} & \val{\textbf{0.50}}{.38}{.61} \\
    CRIMSON & \val{0.44}{.39}{.50} & \val{0.35}{.29}{.40} & \val{0.46}{.41}{.51} & \val{\textbf{0.39}}{.34}{.45} & \val{0.49}{.41}{.57} & \val{0.41}{.33}{.49} \\
    RadFact & \val{0.36}{.29}{.44} & \val{0.23}{.17}{.29} & \val{0.30}{.24}{.36} & \val{0.19}{.13}{.26} & \val{0.39}{.29}{.48} & \val{0.35}{.25}{.45} \\
    GREEN & \val{0.16}{.07}{.25} & \val{0.16}{.09}{.23} & \val{0.14}{.06}{.22} & \val{0.17}{.08}{.25} & \val{0.15}{.02}{.27} & \val{0.17}{.05}{.30} \\
    SRR-BERT & \val{0.14}{.07}{.21} & \val{0.13}{.06}{.21} & \val{0.19}{.13}{.24} & \val{0.13}{.06}{.20} & \val{0.23}{.15}{.32} & \val{0.10}{$-$.01}{.21} \\
    RadCliQ & \val{0.11}{.05}{.17} & \val{0.04}{$-$.03}{.12} & \val{0.07}{.02}{.12} & \val{0.02}{$-$.04}{.08} & \val{0.10}{.01}{.20} & \val{0.11}{.03}{.19} \\
    RadGraph-F1 & \val{0.08}{.02}{.14} & \val{0.10}{.03}{.16} & \val{0.07}{.03}{.12} & \val{0.03}{$-$.02}{.07} & \val{0.08}{.01}{.15} & \val{0.12}{.04}{.22} \\
    ROUGE-L & \val{0.07}{.02}{.13} & \val{0.06}{$-$.01}{.14} & \val{0.03}{$-$.01}{.08} & \val{0.03}{$-$.02}{.08} & \val{0.06}{$-$.03}{.16} & \val{0.06}{$-$.01}{.14} \\
    BERTScore & \val{0.05}{.00}{.11} & \val{0.04}{$-$.04}{.12} & \val{0.01}{$-$.04}{.05} & \val{$-$0.01}{$-$.06}{.04} & \val{0.04}{$-$.06}{.13} & \val{0.03}{$-$.05}{.11} \\
    BLEU & \val{0.02}{$-$.04}{.07} & \val{0.01}{$-$.06}{.09} & \val{$-$0.02}{$-$.06}{.02} & \val{$-$0.01}{$-$.06}{.03} & \val{0.00}{$-$.09}{.08} & \val{0.00}{$-$.08}{.08} \\
    \bottomrule
  \end{tabular}
\end{table*}

\begin{table*}[t]
  \centering
  \caption{\textbf{Agreement with the radiologists' count of each error category on MetricEval-BodyCT.} Kendall's $\tau_b$ (95\% CI) between each metric's score and the radiologists' count of errors of one category, pooled over the 300 candidates. No score here is a per-category count: each metric enters with its single report-level score (RadMatch: aER). \emph{Studies}: studies whose candidates differ in that count; \emph{inter-expert}: agreement between the two radiologists. Bold: best metric per row.}
  \label{tab:bodyct-category}
  \scriptsize
  \setlength{\tabcolsep}{3pt}
  \providecommand{\valwidths}[2]{}\valwidths{$-$0.00}{[$-$.00,.00]}
  \begin{tabular}{lcccccc}
    \toprule
    Category & Studies & Inter-expert & {\color{rmgreen}RadMatch} & CRIMSON & GREEN & RadFact \\
    \midrule
    Hallucinated & 98 & \val{0.82}{.77}{.86} & \val{\textbf{0.63}}{.58}{.68} & \val{0.44}{.39}{.50} & \val{0.16}{.07}{.26} & \val{0.22}{.15}{.29} \\
    Missed & 33 & \val{0.66}{.54}{.77} & \val{0.08}{$-$.01}{.17} & \val{0.11}{.02}{.19} & \val{$-$0.03}{$-$.15}{.08} & \val{\textbf{0.18}}{.09}{.26} \\
    Certainty & 40 & \val{0.59}{.48}{.69} & \val{0.06}{$-$.02}{.14} & \val{0.01}{$-$.07}{.10} & \val{\textbf{0.09}}{$-$.01}{.17} & \val{$-$0.05}{$-$.14}{.03} \\
    Severity & 29 & \val{0.58}{.43}{.72} & \val{0.01}{$-$.07}{.08} & \val{0.09}{.00}{.17} & \val{$-$0.05}{$-$.14}{.04} & \val{\textbf{0.22}}{.13}{.30} \\
    Location & 20 & \val{0.66}{.50}{.80} & \val{0.03}{$-$.07}{.12} & \val{0.12}{.06}{.18} & \val{0.02}{$-$.09}{.11} & \val{\textbf{0.20}}{.12}{.27} \\
    Measurement & 18 & \val{0.82}{.65}{.94} & \val{0.04}{$-$.04}{.13} & \val{0.07}{$-$.01}{.14} & \val{$-$0.02}{$-$.12}{.07} & \val{\textbf{0.17}}{.10}{.24} \\
    Characterization & 13 & \val{0.48}{.26}{.66} & \val{0.06}{$-$.03}{.15} & \val{0.09}{.02}{.15} & \val{0.06}{$-$.02}{.12} & \val{\textbf{0.13}}{.05}{.21} \\
    Comparison & 8 & \val{0.42}{$-$.02}{.69} & \val{0.00}{$-$.09}{.10} & \val{0.11}{.05}{.17} & \val{0.03}{$-$.05}{.10} & \val{\textbf{0.12}}{.02}{.22} \\
    \bottomrule
  \end{tabular}
\end{table*}

\subsection{Construction of the Candidates}
The 100 studies are de-identified chest and abdomen--pelvis CT reports from multiple US centers, one per patient. Each candidate is a copy of the reference whose findings and impression were rewritten in one of about ten reporting styles (terse, itemized, measurement-dense, \etc), one per study, and into which the generator injected a planned set of errors: 16 candidates carry none, 84 one or two, 100 three or four and 100 five to seven (1{,}072 of 1{,}075 planned errors realized). The plan is stratified over error categories and checked against each study's findings: comparison errors only when a prior is mentioned, measurement errors only when the report has measurements, and normal studies receive over-calls rather than omissions. Outside the injected errors, measurements, laterality, negations and hedges are kept unchanged. Length and style are balanced across error levels, so that within a study candidates differ only in content. A separate LLM pass checked each candidate against its plan; the plans themselves are never shown to the radiologists, whose annotation is the only ground truth.

\subsection{Radiologist Annotation}
Four board-certified radiologists annotated the benchmark, each candidate being read by two of them, blinded to the error plan and with the candidates' order shuffled. They logged 1{,}582 errors, $94.5\%$ of them anchored to a report line, and $0.1$, $1.3$, $2.7$ and $4.2$ errors per read on the candidates with $0$, $1$--$2$, $3$--$4$ and $5$--$7$ injected errors. On the 16 candidates without injected errors, they logged only 3 errors in 32 reads, none significant. Of all logged errors (\cref{fig:bodyct}a), $57\%$ are hallucinations, $14\%$ certainty errors, $9\%$ omissions, and $2$--$6\%$ each wrong severity, location, comparison, measurement and characterization; $29\%$ are marked clinically significant. Abdomen reads contain more errors (841) than chest reads (741), but fewer significant ones (198 vs.\ 262). Although every injected error was planned as clinically significant, the radiologists graded $71\%$ of the hallucinations insignificant, and less often at higher doses ($83\%$, $74\%$ and $65\%$): an added finding is often benign. As a check that the totals follow the injected dose, an oracle that knows the dose agrees with the radiologists' total error count at $\tau_b=0.70$, the level of the inter-expert agreement; which errors are significant remains the radiologists' judgment.

\subsection{Agreement by Endpoint, Region and Error Category}
\cref{tab:bodyct} reports agreement on the four endpoints and, for total errors, by region: RadMatch agrees most on both error counts, on the fidelity ranking and in both regions, and CRIMSON on the indication ranking ($0.39$ vs.\ $0.38$). \cref{tab:bodyct-category} correlates each metric's single report-level score with the radiologists' count of one error category; these endpoints suit metrics that output typed errors, and no metric is scored here with such counts. On hallucinated findings, by far the most frequent category, RadMatch agrees most ($0.63$ vs.\ $0.44$ for CRIMSON; inter-expert $0.82$). On the other categories, each a small share of the errors (at most 40 studies whose candidates differ in the count), a report-level score agrees little with the category count: RadFact has the highest point estimate on six of the seven ($0.12$ to $0.22$), and all metrics stay far below the inter-expert agreement.

\section{Agreement with Radiologists: Values and Breakdowns}
\label{app:agreement}
\begin{table*}[t]
  \centering
  \caption{\textbf{Values of \cref{fig:agreement}, with RadPref's total-error endpoint.} Kendall's $\tau_b$ (95\% CI; higher is better) against the radiologists' error counts. ReXVal: total errors, pooled, $n{=}200$; RadEvalExpert: significant errors, pooled, $n{=}624$; RadPref: significant and total errors, pooled over the $200$ candidates and averaged over the three radiologists, inter-expert value as the mean (range) of the three pairs of radiologists; MetricEval-BodyCT: significant errors, pooled over the $300$ candidates and averaged over the two radiologists, inter-expert value with its 95\% CI. RadMatch is scored with aER against total errors and with sER against significant errors. $^\dagger$Reported by the original papers~\cite{green,radeval} (GREEN: its released model); all other values are computed on our runs. RadFact: RadEval's implementation (F1, GPT-4o-mini).}
  \label{tab:agreement-values}
  \scriptsize
  \setlength{\tabcolsep}{5pt}
  \providecommand{\valwidths}[2]{}\valwidths{$-$0.00}{[$-$.00,.00]}
  \begin{tabular}{lccccc}
    \toprule
    & ReXVal & RadEvalExpert & \multicolumn{2}{c}{RadPref} & MetricEval-BodyCT \\ \cmidrule(lr){4-5}
    Metric & total & significant & significant & total & significant \\
    \midrule
    \textit{Inter-expert} & \val{0.81}{.78}{.83}$^\dagger$ & -- & \val{0.77}{.75}{.78} & \val{0.89}{.88}{.92} & \val{0.40}{.31}{.49} \\
    \midrule
    {\color{rmgreen}RadMatch} & \val{0.79}{.73}{.84} & \val{0.52}{.46}{.58} & \val{0.63}{.56}{.70} & \val{0.78}{.73}{.83} & \val{0.45}{.40}{.50} \\
    CRIMSON & \val{0.65}{.56}{.72} & \val{0.23}{.15}{.30} & \val{0.52}{.44}{.59} & \val{0.54}{.46}{.61} & \val{0.35}{.29}{.40} \\
    RadFact & \val{0.57}{.43}{.68} & \val{0.23}{.15}{.30} & \val{0.53}{.46}{.60} & \val{0.61}{.54}{.67} & \val{0.23}{.17}{.29} \\
    GREEN & \val{0.63}{.56}{.69}$^\dagger$ & \val{0.18}{.12}{.25}$^\dagger$ & \val{0.51}{.44}{.58} & \val{0.62}{.55}{.68} & \val{0.16}{.09}{.23} \\
    SRR-BERT & \val{0.46}{.32}{.59} & \val{0.13}{.07}{.19}$^\dagger$ & \val{0.52}{.45}{.60} & \val{0.54}{.46}{.61} & \val{0.13}{.06}{.21} \\
    RadCliQ & \val{0.61}{.55}{.66}$^\dagger$ & \val{0.11}{.05}{.16}$^\dagger$ & \val{0.43}{.36}{.50} & \val{0.49}{.42}{.56} & \val{0.04}{$-$.03}{.12} \\
    RadGraph-F1 & \val{0.57}{.49}{.63}$^\dagger$ & \val{0.01}{$-$.05}{.07}$^\dagger$ & \val{0.40}{.32}{.47} & \val{0.44}{.36}{.51} & \val{0.10}{.03}{.16} \\
    ROUGE-L & \val{0.56}{.48}{.63}$^\dagger$ & \val{0.04}{$-$.02}{.09}$^\dagger$ & \val{0.39}{.31}{.46} & \val{0.42}{.35}{.48} & \val{0.06}{$-$.01}{.14} \\
    BERTScore & \val{0.49}{.40}{.56}$^\dagger$ & \val{0.03}{$-$.03}{.09}$^\dagger$ & \val{0.37}{.29}{.45} & \val{0.42}{.35}{.50} & \val{0.04}{$-$.04}{.12} \\
    BLEU & \val{0.35}{.25}{.43}$^\dagger$ & \val{$-$0.03}{$-$.09}{.02}$^\dagger$ & \val{0.30}{.21}{.38} & \val{0.36}{.28}{.44} & \val{0.01}{$-$.06}{.09} \\
    \bottomrule
  \end{tabular}
\end{table*}

\begin{table}[t]
  \centering
  \caption{\textbf{Agreement on RadEvalExpert by protocol and report section.} Kendall's $\tau_b$ (higher is better) against the radiologists' significant or insignificant error counts, for RadMatch and the RadEval metrics, all computed on our runs (published values in \cref{tab:agreement-values}). Pooled: all 624 candidates; blocked: as in the benchmark's scripts, the report pairs pooled over the studies whose three candidates the metric scores differently (of 148 findings and 60 impression sections). Bold: best per column; \textcolor{black!45}{gray}: 95\% CI includes zero.}
  \label{tab:radeval-category}
  \scriptsize
  \setlength{\tabcolsep}{3pt}
  \begin{tabular}{lccccc}
    \toprule
    & \multicolumn{4}{c}{Significant errors} & Insignificant \\ \cmidrule(lr){2-5}\cmidrule(lr){6-6}
    Metric & \begin{tabular}[b]{@{}c@{}}All,\\pooled\end{tabular} & \begin{tabular}[b]{@{}c@{}}All,\\blocked\end{tabular} & \begin{tabular}[b]{@{}c@{}}Findings,\\blocked\end{tabular} & \begin{tabular}[b]{@{}c@{}}Impression,\\blocked\end{tabular} & \begin{tabular}[b]{@{}c@{}}All,\\blocked\end{tabular} \\
    \midrule
    \textcolor{rmgreen}{RadMatch} & \textbf{0.52} & \textbf{0.47} & \textbf{0.49} & \textbf{0.46} & \textcolor{black!45}{0.03} \\
    \midrule
    CRIMSON & 0.23 & 0.20 & 0.23 & 0.17 & \textcolor{black!45}{$-$0.04} \\
    GREEN & 0.19 & 0.21 & 0.28 & \textcolor{black!45}{0.11} & \textcolor{black!45}{0.02} \\
    RadFact & 0.23 & 0.31 & 0.35 & 0.26 & \textcolor{black!45}{0.02} \\
    SRR-BERT & 0.15 & 0.17 & 0.22 & \textcolor{black!45}{0.07} & \textbf{0.08} \\
    RadCliQ & 0.14 & 0.14 & 0.18 & \textcolor{black!45}{0.10} & \textcolor{black!45}{0.05} \\
    RadGraph-F1 & \textcolor{black!45}{0.01} & 0.12 & 0.14 & \textcolor{black!45}{0.08} & \textcolor{black!45}{0.03} \\
    ROUGE-L & \textcolor{black!45}{0.03} & \textcolor{black!45}{0.06} & \textcolor{black!45}{0.07} & \textcolor{black!45}{0.06} & \textcolor{black!45}{0.00} \\
    BERTScore & \textcolor{black!45}{0.04} & \textcolor{black!45}{0.06} & \textcolor{black!45}{0.09} & \textcolor{black!45}{0.05} & \textcolor{black!45}{0.01} \\
    BLEU & \textcolor{black!45}{0.02} & \textcolor{black!45}{0.12} & \textcolor{black!45}{0.10} & \textcolor{black!45}{0.19} & \textcolor{black!45}{$-$0.11} \\
    \bottomrule
  \end{tabular}
\end{table}

\subsection{Values of \texorpdfstring{\cref{fig:agreement}}{Fig. 4} and ReXVal by Candidate Subset}
\cref{tab:agreement-values} lists the values plotted in \cref{fig:agreement} and, on RadPref, agreement with the radiologists' total error count. On the four ReXVal candidate subsets of CRIMSON's breakdown ($n{=}50$ each, significant errors, baseline values as reported by CRIMSON~\cite{crimson}), RadMatch's $\tau_b$ ranges from $0.73$ to $0.77$, against at most $0.71$ for the best baseline, and its Pearson's $r$ from $0.82$ to $0.90$, against $0.71$ to $0.84$ for CRIMSON, ahead of it on three subsets.

\subsection{RadEvalExpert by Protocol, Section and Error Category}
On RadEvalExpert (\cref{tab:radeval-category}), RadMatch agrees about as much on the findings section ($\tau_b{=}0.49$) as on impressions ($0.46$). The blocked columns follow the benchmark's scripts, which keep for each metric the studies where its scores vary and pool the pairs; with one common set of studies per endpoint, RadMatch's and CRIMSON's values move by at most $0.05$, but restricting $\tau_b$ to pairs of candidates of the same study shrinks RadMatch's lead over CRIMSON on significant errors from $+0.27$ to $+0.10$ [$0.01$, $0.20$]: part of the pooled agreement reflects which studies have more errors. Against the significant errors of one category (blocked), RadMatch has the highest point estimate, with a CI above zero, on four of the six clinical categories (false prediction, $0.41$ vs.\ $0.36$ for CRIMSON, level on a common set of studies; omission; location; spurious comparison), mostly by small margins; on severity and omitted change no metric agrees significantly, and on the non-clinical \emph{inarticulate} category CRIMSON, BERTScore and RadCliQ agree more.

\subsection{RadPref}
\label{app:radpref}
RadPref~\cite{crimson} pairs each of 100 ReXGradient-160K reference reports with two candidates; in the released data, every pair contrasts a MedGemma generation with a GPT-written report. Three cardiothoracic radiologists rate each candidate from 1 to 5 and annotate its false findings, missing findings and attribute errors, each with its clinical significance; significant errors are the false and missing findings labeled urgent or actionable and the attribute errors labeled significant. All metrics score the full reports. RadMatch receives the patient's age, sex and indication as its study indication, and CRIMSON receives them as patient context, as in CRIMSON's evaluation. RadEval's CRIMSON implementation does not forward this context, so we pass it explicitly; our GPT-5.2 CRIMSON run then reproduces the published agreement ($\tau_b=0.64$, $r=0.82$). Under CRIMSON's rating protocol (the correlation over the 100 cases between a metric's score difference of the two candidates and the radiologists' rating difference), RadMatch is level with CRIMSON on $\tau_b$ against the average of the three radiologists ($0.64$ each; ahead for one radiologist, behind for the other two), CRIMSON has the higher Pearson's $r$ ($0.82$ vs.\ $0.76$), and RadMatch, which counts errors as integers, gives both candidates the same score on 26 of the 100 cases. On the radiologists' error counts (\cref{tab:agreement-values}), RadMatch exceeds CRIMSON by $0.11$ on significant errors (95\% CI [$0.03$, $0.21$], paired study-clustered bootstrap) and by $0.24$ [$0.16$, $0.32$] on total errors.

\section{Clinical Validation, Qualitative Analysis and Dashboard}
\label{app:clinical-validation}
\label{app:dashboard}  %

\subsection{Decisions Confirmed by the Radiologists}
The radiologists saw RadMatch's decisions and confirmed or corrected each one (studies and protocol in \cref{sec:clinical-validation}), so the rates of \cref{tab:clinical-validation} measure how often its decisions survive expert review, not its agreement with an unaided read. \cref{tab:clinical-validation} gives them per region, with the item counts. Of RadMatch's 100 significant-error calls, the radiologists keep 75 and lower 25 (21 to insignificant, 4 to no error); they raise 10 of its insignificant calls to significant, and 3 more errors they call significant have no counterpart in RadMatch's record (13 of their 88 significant errors in all). Per study, RadMatch's significant-error count and the corrected one lie within one error in 53 of the 60 studies ($88\%$).

\begin{table*}[t]
  \centering
  \caption{\textbf{Share of RadMatch's decisions that the radiologists confirm.} Gemma 4 31B judge; 60 studies, each reviewed by one of three radiologists; share (\%) of decisions confirmed, with study-clustered bootstrap 95\% CI. \emph{Matching}: linked findings describe the same observation, or an unlinked one has no counterpart; \emph{error call}: the radiologists keep RadMatch's call (significant, insignificant or no error); \emph{error, any significance}: an item counted as an error is one, whatever its significance; \emph{attribute grades}: the grade (none, minor, major) of each attribute of a linked pair. $^*$Significant-error calls the radiologists keep, and per-study $\tau_b$ between RadMatch's significant-error count and the radiologist-corrected one.}
  \label{tab:clinical-validation}
  \scriptsize
  \setlength{\tabcolsep}{5pt}
  \begin{tabular}{lcccc}
    \toprule
    & \multicolumn{4}{c}{RadMatch} \\ \cmidrule(lr){2-5}
    & Chest CT & Abdomen CT & Chest X-ray & All \\
    \midrule
    Studies (items) & 24 (424) & 24 (570) & 12 (97) & 60 (1{,}091) \\
    \midrule
    Matching confirmed & 97.2 & 99.6 & 97.8 & 98.5\,\ci{[97.4, 99.4]} \\
    \multicolumn{5}{@{}l}{\textit{Error call confirmed}} \\
    \quad significant error & 82.1 & 70.5 & 81.8 & 75.0\,\ci{[65.2, 85.3]} \\
    \quad insignificant error & 91.8 & 84.7 & 76.2 & 86.2\,\ci{[80.7, 91.7]} \\
    \quad no error & 99.4 & 99.3 & 100.0 & 99.4\,\ci{[98.8, 99.9]} \\
    \quad error, any significance & 97.8 & 91.8 & 90.6 & 93.6\,\ci{[90.0, 96.9]} \\
    \multicolumn{5}{@{}l}{\textit{Attribute grades confirmed}} \\
    \quad all & 99.8 & 99.6 & 99.4 & 99.7\,\ci{[99.4, 99.8]} \\
    \quad minor or major & 89.5 & 46.7 & 80.0 & 72.7\,\ci{[54.3, 91.9]}  \\
    \midrule
    Significant calls kept\textsuperscript{*} & 23 of 28 & 43 of 61 & 9 of 11 & 75 of 100 \\
    Per-study $\tau_b$\textsuperscript{*} & 0.88 & 0.64 & 0.55 & 0.71\,\ci{[0.58, 0.82]}  \\
    \bottomrule
  \end{tabular}
\end{table*}

\subsection{Qualitative Analysis}
\label{app:qualitative}
We audit the 20 ReXVal pairs where RadMatch's significant-error count (Claude Opus 5.5 judge) diverges most from the radiologists' mean, by standardized residual. For each pair, the local Gemma 4 31B, so that the reports stay on our servers, reads the two reports, the radiologists' error counts and RadMatch's stored decisions, and names the main cause from a fixed list of failure modes of the extraction, tiering, matching and grading stages. Eleven pairs are over-counts, mostly from grading (a difference scored major that the radiologists do not count as significant, five pairs) and atomization (one clinical judgment split into several scored findings, four); nine are under-counts, from a significant difference graded minor (four) or a lenient match that books a mischaracterization as a partial match rather than an error (three). Significance tiering explains two pairs in each direction.

\subsection{Diagnostic Views and the RadMatch Dashboard}
\begin{figure*}[tp]
  \centering
  \begin{minipage}[t]{0.5\textwidth}\vspace{0pt}
    \includegraphics[width=\linewidth]{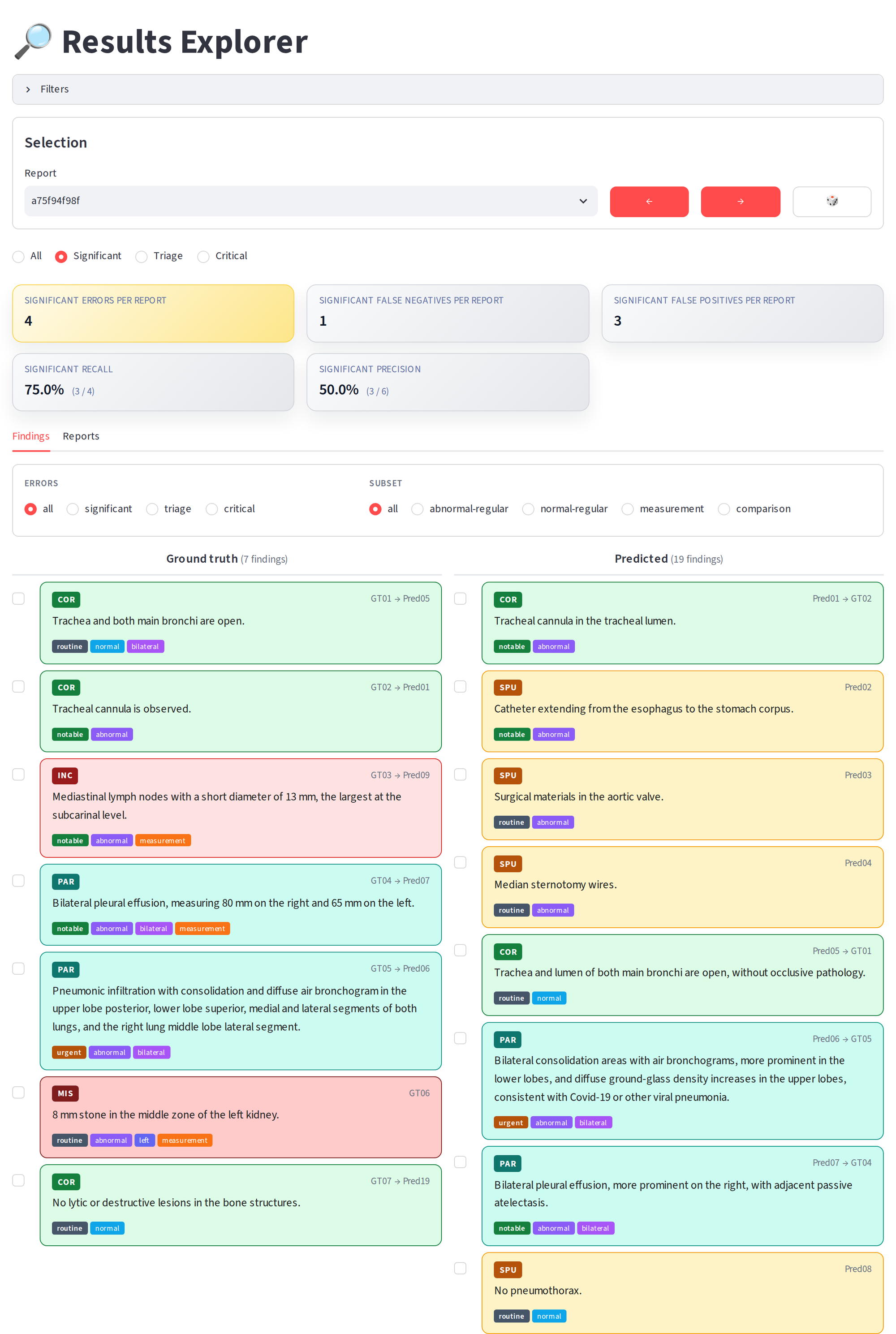}
  \end{minipage}\hfill
  \begin{minipage}[t]{0.48\textwidth}\vspace{0pt}
    \includegraphics[width=\linewidth]{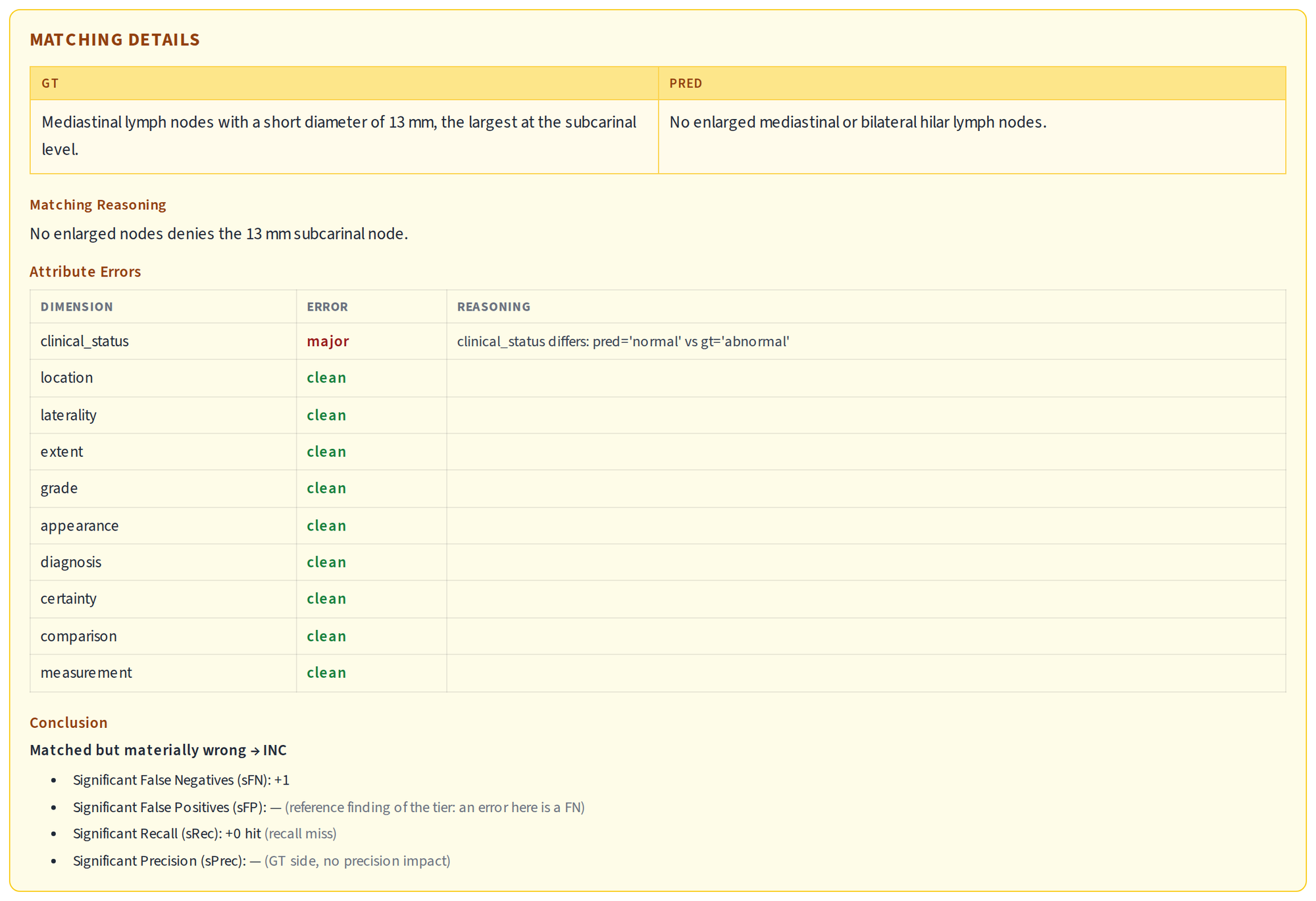}
    \vspace{4pt}
    \caption{\textbf{RadMatch dashboard, study view.} NV-Reason-CT on CT-RATE study \texttt{a75f94f98f} (no indication given); all seven reference findings and the first eight generated ones. \emph{Left:} reference and candidate findings side by side, colored by verdict and tagged with their significance tier and status. \emph{Right:} the matching rationale and the grade per attribute of a selected pair, here the reference's 13\,mm subcarinal node against the candidate's denial of enlarged mediastinal lymph nodes: an inverted clinical status, so the match is \textsc{INC}.}
    \label{fig:dashboard-explorer}
  \end{minipage}
\end{figure*}
Because every counted error traces to a finding pair, RadMatch's records give views a single score cannot: by tier, verdict and finding subset. At the triage tier (\textsc{critical} and \textsc{urgent} findings), for instance, recall is $0.42$ on ReXVal, whose candidates are other patients' reports, $0.20$ on RadEvalExpert, whose candidates are model generations, and $0.90$ on MetricEval-BodyCT, whose candidates are edited copies of the reference.

RadMatch persists every intermediate record (extracted findings, the matching with its rationale, each attribute grade with the judge's reasoning), and the released code ships an optional interactive dashboard (two commands on a results folder) that reads these records back, so that a model builder can go from a dataset-level number to the finding pairs behind it and a radiologist can check, and contest, each verdict. For NV-Reason-CT on the 1{,}549 CT-RATE validation studies, its summary view reproduces the benchmark numbers of \cref{tab:rrg-ct-detail} ($1.67$ significant errors per report: $1.29$ missed or contradicted reference findings and $0.38$ unsupported claims) and splits them by verdict, attribute and finding subset: an inverted clinical status is the most frequent major attribute error ($1{,}490$) and an imprecise location the most frequent minor one ($1{,}063$), and recall falls from $38\%$ overall to about $20\%$ on measured findings, whose size the model rarely states, and on longitudinal comparisons. Its study view (\cref{fig:dashboard-explorer}) shows why one study scores 4 significant errors although its bilateral pleural effusions and pneumonia are matched: the model asserts a catheter running from the esophagus into the stomach, an enlarged heart and a dilated main pulmonary artery that the reference does not mention, and denies enlarged mediastinal lymph nodes where the reference measures a 13\,mm subcarinal node, an inverted clinical status that counts as a significant false negative. \textsc{routine} findings and unreferenced normal statements appear as unmatched but, as in the score, are not counted. A radiologist who judges that the catheter, in this intubated patient, was present but left unreported can see exactly which finding to overrule.

\section{Robustness and Ablations}
\label{app:ablations}
This section measures run-to-run variance and judge score levels, varies the score definition, and tests a single-call count and padding.

\subsection{Run-to-Run Variance and Judge Score Levels}
\label{app:variance}
We re-run RadMatch (Gemma 4 31B, $T{=}0$) and CRIMSON (GPT-5.2) on every pair of ReXVal, RadEvalExpert and MetricEval-BodyCT (\cref{tab:intra-variance}). RadMatch returns the same count in every run for $99.5\%$, $99.5\%$ and $76\%$ of the pairs, and its benchmark-level mean moves by at most $0.03$ errors per report (standard deviation over runs), so a change in the mean above about $0.06$ exceeds run-to-run noise. Changing the serving hardware or concurrency moves more pairs ($7\%$, $13\%$ and $26\%$), yet agreement with radiologists stays within $0.002$, $0.004$ and $0.017$ $\tau_b$. CRIMSON's per-report score changes more often (unchanged for $62\%$, $50\%$ and $19\%$ of the pairs), and relative to the spread across reports RadMatch is the more repeatable (ICC $0.957$--$1.000$ vs.\ $0.919$--$0.970$). Neither is bitwise reproducible, as CRIMSON's reasoning judge takes no temperature and batched GPU inference is not deterministic even at $T{=}0$, whereas GREEN (greedy decoding) and the lexical, embedding and graph-based metrics are.

\begin{table}[t]
  \centering
  \caption{\textbf{Run-to-run variance of LLM-based scores.} The same report pairs re-scored with the same judge and settings: RadMatch four times (Gemma 4 31B, one server, fixed concurrency, $T{=}0$), CRIMSON five times (GPT-5.2, which does not accept a temperature). Score: RadMatch's errors per report (aER on ReXVal and MetricEval-BodyCT, sER on RadEvalExpert; lower is better) and the CRIMSON score (higher is better). \emph{Mean score}: corpus mean $\pm$ standard deviation over runs; \emph{per-report SD}: a report's standard deviation across runs, averaged over reports; \emph{unchanged reports}: share scored identically in all runs; \emph{largest change}: largest gap between a report's highest and lowest score; ICC(1,1): one-way intraclass correlation across runs, comparable between the two metrics. The main results come from a fifth run, on other hardware and concurrency.}
  \label{tab:intra-variance}
  \scriptsize
  \setlength{\tabcolsep}{2pt}
  \begin{tabular}{llccc}
    \toprule
    & & ReXVal & \begin{tabular}[b]{@{}c@{}}RadEval\\Expert\end{tabular} & \begin{tabular}[b]{@{}c@{}}MetricEval-\\BodyCT\end{tabular} \\
    \midrule
    \multirow{5}{*}{\begin{tabular}{@{}l@{}}RadMatch\\gemma-4-31b\end{tabular}} & Mean score & 1.461\,\cis{$\pm$0.003} & 2.213\,\cis{$\pm$0.001} & 2.210\,\cis{$\pm$0.030} \\
     & Per-report SD & 0.003 & 0.003 & 0.153 \\
     & Unchanged reports & 99.5\% & 99.5\% & 75.7\% \\
     & Largest change & 1 & 2 & 4 \\
     & ICC(1,1) & 1.000 & 0.999 & 0.957 \\
    \midrule
    \multirow{5}{*}{\begin{tabular}{@{}l@{}}CRIMSON\\gpt-5.2\end{tabular}} & Mean score & 0.479\,\cis{$\pm$0.006} & $-$0.025\,\cis{$\pm$0.003} & 0.704\,\cis{$\pm$0.004} \\
     & Per-report SD & 0.041 & 0.051 & 0.057 \\
     & Unchanged reports & 62.0\% & 49.7\% & 18.7\% \\
     & Largest change & 1.00 & 1.03 & 1.00 \\
     & ICC(1,1) & 0.970 & 0.939 & 0.919 \\
    \bottomrule
  \end{tabular}
\end{table}

\paragraph{Score level.} Judges with close agreement count at different levels (\cref{tab:judge-levels}): on ReXVal, Gemma 4 31B, Claude Opus 5.5 and GPT-4.1 reach $\tau_b{=}0.79$, $0.80$ and $0.79$ but find $1.46$, $1.41$ and $1.62$ errors per report. RadMatch scores are thus comparable only under a fixed judge, to be reported with them, whereas its agreement with radiologists, which depends on how a judge ranks reports, stays close across strong judges (\cref{tab:agreement}).

\providecommand{\prevver}{\textsuperscript{\textcolor{red}{$\circ$}}}
\begin{table}[t]
  \centering
  \caption{\textbf{RadMatch's score level depends on the judge.} Mean errors per report (aER on ReXVal and MetricEval-BodyCT, sER on RadEvalExpert) on the same report pairs, rows in the order of \cref{tab:agreement}. Among the judges within 0.05 $\tau_b$ of the best one on a benchmark (and at most 10\% matching fallbacks), the mean ranges over 1.40--1.62 on ReXVal and 2.16--2.63 on RadEvalExpert, far beyond the run-to-run variation of one judge (\cref{tab:intra-variance}). $^\dagger$LLM of a baseline metric.}
  \label{tab:judge-levels}
  \scriptsize
  \setlength{\tabcolsep}{3pt}
  \begin{tabular}{lccc}
    \toprule
    LLM & ReXVal & \begin{tabular}[b]{@{}c@{}}RadEval\\Expert\end{tabular} & \begin{tabular}[b]{@{}c@{}}MetricEval-\\BodyCT\end{tabular} \\
    \midrule
    qwen3.8-27b & 1.50 & 2.34 & 2.55 \\
    gpt-6-astra & 1.43 & 2.16 & 2.20 \\
    opus-5.5 & 1.41 & 2.18 & 2.24 \\
    gemma-4-31b & 1.46 & 2.18 & 2.18 \\
    gpt-4.1$^\dagger$ & 1.62 & 2.63 & 2.91 \\
    gpt-5.2$^\dagger$ & 1.40 & 2.20 & 2.64 \\
    qwen3.5-35b-a3b & 1.51 & 2.26 & 2.85 \\
    medgemma-1.5-4b & 1.33 & 2.36 & 10.97 \\
    \bottomrule
  \end{tabular}
\end{table}

\subsection{Score Definition}
\cref{tab:score-definition} re-derives alternative scores from the same stored records, isolating the definition of the score from the judge's verdicts; each of our two counts is best or tied-best where we use it. Counting only significant errors loses $0.07$ to $0.12$ against total errors and ratings; counting all errors loses $0.07$ against MetricEval-BodyCT's significant errors and is level on RadEvalExpert's ($+0.05$ [$-0.01$, $0.10$]). The triage or \textsc{critical} tier alone loses up to $0.68$, and the misses or the unsupported claims alone lose on almost every endpoint (the misses most on MetricEval-BodyCT, the unsupported claims on ReXVal and RadPref).

\begin{table}[t]
  \centering
  \caption{\textbf{Alternative RadMatch scores read from the same records.} Kendall's $\tau_b$ under each benchmark's protocol. Bold: the count RadMatch uses on that endpoint (aER against total errors and ratings, sER against significant errors). Triage tier: \textsc{critical} and \textsc{urgent} findings; FN or FP only: the misses or the unsupported claims alone. $^*$: the paired study-clustered bootstrap difference from that count excludes zero.}
  \label{tab:score-definition}
  \scriptsize
  \setlength{\tabcolsep}{3.2pt}
  \begin{tabular}{lcccccc}
    \toprule
    & & & \multicolumn{2}{c}{RadPref} & \multicolumn{2}{c}{MetricEval-BodyCT} \\ \cmidrule(lr){4-5}\cmidrule(lr){6-7}
    Score & \begin{tabular}[b]{@{}c@{}}ReXVal\end{tabular} & \begin{tabular}[b]{@{}c@{}}RadEval\\Expert\end{tabular} & \begin{tabular}[b]{@{}c@{}}rating\end{tabular} & \begin{tabular}[b]{@{}c@{}}total\end{tabular} & \begin{tabular}[b]{@{}c@{}}total\end{tabular} & \begin{tabular}[b]{@{}c@{}}significant\end{tabular} \\
    \midrule
    All errors (aER) & \textbf{0.79} & 0.57 & \textbf{0.64} & \textbf{0.78} & \textbf{0.53} & 0.38\rlap{$^*$} \\
    Significant errors (sER) & 0.72\rlap{$^*$} & \textbf{0.52} & 0.57 & 0.69\rlap{$^*$} & 0.40\rlap{$^*$} & \textbf{0.45} \\
    \midrule
    Triage tier only & 0.47\rlap{$^*$} & 0.19\rlap{$^*$} & 0.45\rlap{$^*$} & 0.44\rlap{$^*$} & 0.25\rlap{$^*$} & 0.30\rlap{$^*$} \\
    \textsc{critical} tier only & 0.23\rlap{$^*$} & 0.14\rlap{$^*$} & 0.18\rlap{$^*$} & 0.11\rlap{$^*$} & 0.12\rlap{$^*$} & 0.15\rlap{$^*$} \\
    aER, FN only & 0.73\rlap{$^*$} & 0.46\rlap{$^*$} & 0.64 & 0.73\rlap{$^*$} & 0.30\rlap{$^*$} & 0.25\rlap{$^*$} \\
    aER, FP only & 0.50\rlap{$^*$} & 0.31\rlap{$^*$} & 0.18\rlap{$^*$} & 0.36\rlap{$^*$} & 0.39\rlap{$^*$} & 0.25\rlap{$^*$} \\
    sER, FN only & 0.63\rlap{$^*$} & 0.38\rlap{$^*$} & 0.60 & 0.55\rlap{$^*$} & 0.18\rlap{$^*$} & 0.21\rlap{$^*$} \\
    sER, FP only & 0.48\rlap{$^*$} & 0.32\rlap{$^*$} & 0.30\rlap{$^*$} & 0.43\rlap{$^*$} & 0.40\rlap{$^*$} & 0.44 \\
    \bottomrule
  \end{tabular}
\end{table}

\subsection{Single-Call Baseline}
\label{app:single-call}
The single call asks the same Gemma 4 31B judge for the number of clinically significant errors. As reported in \cref{sec:ablation}, it is on par with RadMatch on ReXVal ($0.81$ vs.\ $0.79$), on RadEvalExpert ($0.56$ vs.\ $0.52$), on RadPref's total errors ($0.76$ vs.\ $0.78$) and on MetricEval-BodyCT's significant errors, the benchmark's primary endpoint ($0.42$ vs.\ $0.45$), and it agrees more than RadMatch with MetricEval-BodyCT's total errors ($+0.14$ [$0.08$, $0.21$]; \cref{tab:agreement-sources}, which also lists CRIMSON's and GREEN's official scores and the candidate's word count). The two count different things (\cref{tab:singlecall-significance}). Run at the judge's default temperature ($T{=}1$), the single call returns a count closer to the radiologists' total than to their significant errors on ReXVal, RadPref and MetricEval-BodyCT, and on the last two it agrees with their insignificant errors as much as with their significant ones. RadMatch's sER separates the two better: it matches the radiologists' significant count on ReXVal and RadPref, overcounts on MetricEval-BodyCT ($1.15$ vs.\ $0.77$) and undercounts on RadEvalExpert ($2.18$ vs.\ $3.94$; single call $3.43$).

\begin{table}[t]
  \centering
  \caption{\textbf{Level of the significant-error counts and what they track.} Gemma 4 31B judge. \emph{Top}: mean count per report of the radiologists' significant and total errors, of the single call and of RadMatch's sER, and their mean absolute error (MAE) against the radiologists' significant count. \emph{Bottom}: $\tau_b$ with the radiologists' significant (sig.) and insignificant (insig.) counts, under each benchmark's protocol; $^*$: their paired study-clustered bootstrap difference excludes zero. The single call ran at the judge's default temperature ($T{=}1$), RadMatch at $T{=}0$.}
  \label{tab:singlecall-significance}
  \scriptsize
  \setlength{\tabcolsep}{2.5pt}
  \begin{tabular*}{0.86\linewidth}{@{\extracolsep{\fill}}lcccccc@{}}
    \toprule
    & \multicolumn{2}{c}{Radiologists} & \multicolumn{2}{c}{Single call} & \multicolumn{2}{c}{{\color{rmgreen}RadMatch sER}} \\
    \cmidrule(lr){2-3}\cmidrule(lr){4-5}\cmidrule(lr){6-7}
    & sig. & total & count & MAE & count & MAE \\
    \midrule
    ReXVal & 1.17 & 1.64 & 1.69 & 0.83 & 1.17 & 0.69 \\
    RadEvalExpert & 3.94 & 4.67 & 3.43 & 1.51 & 2.18 & 2.08 \\
    RadPref & 1.36 & 2.64 & 2.29 & 1.26 & 1.40 & 0.83 \\
    MetricEval-BodyCT & 0.77 & 2.64 & 2.40 & 1.72 & 1.15 & 0.79 \\
    \bottomrule
  \end{tabular*}

  \vspace{3pt}
  \begin{tabular*}{0.86\linewidth}{@{\extracolsep{\fill}}lcccc@{}}
    \toprule
    & \multicolumn{2}{c}{Single call} & \multicolumn{2}{c}{{\color{rmgreen}RadMatch sER}} \\
    \cmidrule(lr){2-3}\cmidrule(lr){4-5}
    $\tau_b$ with & sig. & insig. & sig. & insig. \\
    \midrule
    ReXVal & 0.80\rlap{$^*$} & 0.58 & 0.75\rlap{$^*$} & 0.50 \\
    RadEvalExpert & 0.56\rlap{$^*$} & 0.10 & 0.52\rlap{$^*$} & $-$0.06 \\
    RadPref & 0.61 & 0.65 & 0.63 & 0.54 \\
    MetricEval-BodyCT & 0.42 & 0.46 & 0.45\rlap{$^*$} & 0.17 \\
    \bottomrule
  \end{tabular*}
\end{table}

\begin{table}[t]
  \centering
  \caption{\textbf{Agreement of RadMatch, a single-call count and other scores with the radiologists.} Kendall's $\tau_b$ under each benchmark's protocol (RadPref and MetricEval-BodyCT: total errors). CRIMSON and GREEN: their official normalized scores; single call: the judge asked for the number of clinically significant errors. All values are computed on our runs; on ReXVal and RadEvalExpert they can differ by up to $0.02$ from those of \cref{fig:agreement}, and on RadPref and MetricEval-BodyCT they are the total-error values of \cref{tab:agreement-values,tab:bodyct}.}
  \label{tab:agreement-sources}
  \scriptsize
  \setlength{\tabcolsep}{3pt}
  \begin{tabular}{lcccc}
    \toprule
    & ReXVal & \begin{tabular}[b]{@{}c@{}}RadEval\\Expert\end{tabular} & RadPref & \begin{tabular}[b]{@{}c@{}}MetricEval-\\BodyCT\end{tabular} \\
    \midrule
    {\color{rmgreen}RadMatch (gemma-4-31b)} & 0.79 & 0.52 & 0.78 & 0.53 \\
    Single-call count (gemma-4-31b) & 0.81 & 0.56 & 0.76 & 0.66 \\
    \midrule
    CRIMSON score (gpt-5.2) & 0.65 & 0.23 & 0.54 & 0.44 \\
    CRIMSON score (gemma-4-31b) & 0.61 & 0.18 & 0.54 & 0.43 \\
    GREEN score & 0.62 & 0.19 & 0.62 & 0.16 \\
    \midrule
    Candidate word count & 0.59 & 0.24 & 0.21 & 0.02 \\
    \bottomrule
  \end{tabular}
\end{table}

\subsection{Padding}
\label{app:padding}
Preference optimization with GREEN as reward produced reports up to $2.5\times$ longer, mostly through repeated statements~\cite{chexalign}. On ReXVal (\cref{tab:padding}), repeating each candidate raises its GREEN score by $0.08$ but RadMatch's error count by only $0.02$ [$-0.02$, $0.07$] (unchanged for $94.5\%$ of the reports), as a repeated finding matches the same reference finding. Appending generic normal statements barely moves RadMatch's count ($+0.08$), while hedged abnormal statements add $2.3$ errors per report, as unsupported claims should. Both metrics keep their agreement under duplication.

\begin{table}[t]
  \centering
  \caption{\textbf{Padding stress test on ReXVal.} On the 200 ReXVal candidates, each candidate is repeated (\emph{duplicated}), or followed by five generic normal statements (\emph{normal}) or three hedged abnormal statements (\emph{hedged}). Mean change of the score with a bootstrap 95\% CI (RadMatch: aER, lower is better; GREEN: higher is better), and $\tau_b$ with the radiologists' total error count (\emph{none}: the unpadded candidates in this run).}
  \label{tab:padding}
  \scriptsize
  \setlength{\tabcolsep}{2.5pt}
  \providecommand{\valwidths}[2]{}\valwidths{$+$0.00}{[$-$.00,$-$.00]}
  \begin{tabular}{lcccc}
    \toprule
    & \multicolumn{2}{c}{RadMatch (gemma-4-31b)} & \multicolumn{2}{c}{GREEN} \\
    \cmidrule(lr){2-3}\cmidrule(lr){4-5}
    Padding & change & $\tau_b$ & change & $\tau_b$ \\
    \midrule
    None & -- & 0.80 & -- & 0.62 \\
    Duplicated & \val{$+$0.02}{$-$.02}{.07} & 0.80 & \val{$+$0.08}{.05}{.11} & 0.63 \\
    Normal & \val{$+$0.08}{.00}{.16} & 0.79 & \val{$-$0.06}{$-$.10}{$-$.02} & 0.54 \\
    Hedged & \val{$+$2.33}{2.08}{2.58} & 0.52 & \val{$-$0.32}{$-$.40}{$-$.25} & 0.23 \\
    \bottomrule
  \end{tabular}
\end{table}

\section{Report Generation Benchmark: Setup and Full Results}
\label{app:rrg}
\begin{table*}[tp]
  \centering
  \caption{\textbf{CT report generation: full RadMatch results.} Gemma 4 31B judge. sER $=$ sFN $+$ sFP, significant errors per report (lower is better); sFN: significant reference findings missed or contradicted, pertinent negatives included; sFP: unsupported or misstated significant claims; sPrec / sRec: significant precision and recall; Saved: the normal template's sER minus the method's on the same studies ($>0$: fewer errors than the template), with a paired study-clustered bootstrap 95\% CI (2{,}000 draws); ROUGE-L for reference. Rows sorted by sER; bold: best model per column. Italic rows are references, not ranked: the image-blind normal template (the zero point) at its rank, and an empty report (every significant reference finding missed, nothing claimed). n/a: precision undefined (no significant claim); the template's precision only reflects how the extractor tiers its negative statements and is not comparable with the models'. $^\diamond$In-domain run: the model is trained on this dataset's training split (on AMOS-MM, HoloCT's split holds 315 of the 399 studies scored here). Every method scored every study.}
  \label{tab:rrg-ct-detail}
  \scriptsize
  \setlength{\tabcolsep}{3pt}
  \begin{tabular*}{\linewidth}{@{\extracolsep{\fill}}lccccccc@{}}
    \toprule
    Method & sER$\downarrow$ & sFN$\downarrow$ & sFP$\downarrow$ & sPrec$\uparrow$ & sRec$\uparrow$ & Saved vs.\ template [95\% CI]$\uparrow$ & ROUGE-L$\uparrow$ \\
    \midrule
    \multicolumn{8}{@{}l}{\textbf{Merlin-CT} (abdomen, test split, $n{=}5{,}125$)} \\
    \midrule
    \textit{Normal template} & \textit{5.34} & \textit{5.34} & \textit{0.00} & \textit{n/a} & \textit{0.010} & \textit{0} & \textit{0.268} \\
    nnFoundationViT$^\diamond$ & \textbf{5.38} & \textbf{4.07} & 1.31 & \textbf{0.392} & \textbf{0.245} & \textbf{$-$0.04}\,\cis{[$-$0.09, $+$0.01]} & \textbf{0.334} \\
    \jolia{}$^\diamond$ & 5.76 & 4.79 & 0.97 & 0.333 & 0.113 & $-$0.42\,\cis{[$-$0.45, $-$0.38]} & 0.322 \\
    NV-Reason-CT & 5.80 & 4.90 & \textbf{0.90} & 0.281 & 0.092 & $-$0.46\,\cis{[$-$0.49, $-$0.42]} & 0.193 \\
    HoloCT-4B$^\diamond$ & 5.82 & 4.60 & 1.22 & 0.328 & 0.148 & $-$0.48\,\cis{[$-$0.52, $-$0.43]} & 0.249 \\
    Merlin$^\diamond$ & 6.50 & 4.98 & 1.52 & 0.196 & 0.078 & $-$1.16\,\cis{[$-$1.20, $-$1.13]} & 0.284 \\
    HoloCT-8B$^\diamond$ & 6.72 & 4.74 & 1.98 & 0.211 & 0.122 & $-$1.37\,\cis{[$-$1.43, $-$1.32]} & 0.231 \\
    MedGemma 1.5 4B & 6.77 & 5.23 & 1.54 & 0.108 & 0.032 & $-$1.42\,\cis{[$-$1.47, $-$1.38]} & 0.106 \\
    RadFM & 6.90 & 5.38 & 1.53 & 0.056 & 0.004 & $-$1.56\,\cis{[$-$1.58, $-$1.53]} & 0.070 \\
    Med3DVLM & 7.49 & 5.35 & 2.14 & 0.039 & 0.009 & $-$2.15\,\cis{[$-$2.20, $-$2.09]} & 0.088 \\
    M3D-LaMed & 8.59 & 5.35 & 3.24 & 0.034 & 0.009 & $-$3.25\,\cis{[$-$3.34, $-$3.15]} & 0.087 \\
    \textit{Empty report} & \textit{5.40} & \textit{5.40} & \textit{0.00} & \textit{n/a} & \textit{0.000} & \textit{$-$0.06\,\cis{[$-$0.06, $-$0.05]}} & \textit{0.000} \\
    \midrule
    \multicolumn{8}{@{}l}{\textbf{AMOS-MM} (abdomen and pelvis, validation set, used as test, $n{=}399$)} \\
    \midrule
    \textit{Normal template} & \textit{2.41} & \textit{2.41} & \textit{0.00} & \textit{n/a} & \textit{0.000} & \textit{0} & \textit{0.333} \\
    NV-Reason-CT & \textbf{2.79} & 2.24 & \textbf{0.55} & \textbf{0.166} & 0.072 & \textbf{$-$0.38}\,\cis{[$-$0.48, $-$0.29]} & 0.158 \\
    \jolia{} & 3.21 & 2.30 & 0.92 & 0.099 & 0.047 & $-$0.80\,\cis{[$-$0.92, $-$0.69]} & 0.148 \\
    HoloCT-4B$^\diamond$ & 3.49 & 2.19 & 1.30 & 0.135 & 0.091 & $-$1.08\,\cis{[$-$1.22, $-$0.95]} & 0.320 \\
    HoloCT-8B$^\diamond$ & 3.65 & 2.16 & 1.49 & 0.138 & 0.103 & $-$1.24\,\cis{[$-$1.41, $-$1.07]} & \textbf{0.334} \\
    MedGemma 1.5 4B & 3.66 & 2.34 & 1.32 & 0.032 & 0.029 & $-$1.25\,\cis{[$-$1.40, $-$1.10]} & 0.120 \\
    nnFoundationViT & 3.70 & \textbf{1.99} & 1.70 & 0.165 & \textbf{0.172} & $-$1.29\,\cis{[$-$1.47, $-$1.10]} & 0.165 \\
    Merlin & 3.76 & 2.37 & 1.39 & 0.025 & 0.015 & $-$1.36\,\cis{[$-$1.44, $-$1.26]} & 0.151 \\
    RadFM & 3.92 & 2.40 & 1.52 & 0.003 & 0.002 & $-$1.51\,\cis{[$-$1.59, $-$1.43]} & 0.104 \\
    Med3DVLM & 4.55 & 2.39 & 2.17 & 0.017 & 0.009 & $-$2.14\,\cis{[$-$2.33, $-$1.95]} & 0.105 \\
    M3D-LaMed & 5.14 & 2.39 & 2.75 & 0.015 & 0.006 & $-$2.73\,\cis{[$-$3.05, $-$2.45]} & 0.104 \\
    \textit{Empty report} & \textit{2.41} & \textit{2.41} & \textit{0.00} & \textit{n/a} & \textit{0.000} & \textit{0.00\,\cis{[0.00, 0.00]}} & \textit{0.000} \\
    \midrule
    \multicolumn{8}{@{}l}{\textbf{CT-RATE} (chest, validation split, one volume per study, $n{=}1{,}549$)} \\
    \midrule
    NV-Reason-CT$^\diamond$ & \textbf{1.67} & \textbf{1.29} & \textbf{0.38} & \textbf{0.526} & \textbf{0.384} & \textbf{$+$0.44}\,\cis{[$+$0.37, $+$0.50]} & 0.250 \\
    nnFoundationViT$^\diamond$ & 1.91 & 1.34 & 0.57 & 0.457 & 0.358 & $+$0.19\,\cis{[$+$0.13, $+$0.26]} & 0.398 \\
    \textit{Normal template} & \textit{2.11} & \textit{1.85} & \textit{0.26} & \textit{0.660} & \textit{0.116} & \textit{0} & \textit{0.448} \\
    HoloCT-4B$^\diamond$ & 2.19 & 1.59 & 0.60 & 0.371 & 0.239 & $-$0.08\,\cis{[$-$0.15, $-$0.02]} & \textbf{0.421} \\
    CT-CHAT$^\diamond$ & 2.24 & 1.77 & 0.47 & 0.366 & 0.153 & $-$0.13\,\cis{[$-$0.19, $-$0.07]} & 0.347 \\
    HoloCT-8B$^\diamond$ & 2.26 & 1.60 & 0.66 & 0.359 & 0.233 & $-$0.16\,\cis{[$-$0.21, $-$0.10]} & 0.420 \\
    MedGemma 1.5 4B & 2.57 & 1.93 & 0.64 & 0.186 & 0.078 & $-$0.46\,\cis{[$-$0.52, $-$0.40]} & 0.109 \\
    Reg2RG$^\diamond$ & 2.76 & 1.80 & 0.96 & 0.234 & 0.138 & $-$0.66\,\cis{[$-$0.70, $-$0.61]} & 0.268 \\
    RadFM & 2.79 & 1.99 & 0.81 & 0.142 & 0.050 & $-$0.68\,\cis{[$-$0.74, $-$0.62]} & 0.065 \\
    M3D-LaMed & 3.75 & 2.06 & 1.69 & 0.022 & 0.016 & $-$1.64\,\cis{[$-$1.72, $-$1.57]} & 0.070 \\
    Med3DVLM & 3.76 & 2.05 & 1.71 & 0.024 & 0.019 & $-$1.65\,\cis{[$-$1.73, $-$1.57]} & 0.065 \\
    BTB3D$^\diamond$ & 3.86 & 1.81 & 2.05 & 0.111 & 0.135 & $-$1.75\,\cis{[$-$1.88, $-$1.63]} & 0.299 \\
    \textit{Empty report} & \textit{2.09} & \textit{2.09} & \textit{0.00} & \textit{n/a} & \textit{0.000} & \textit{$+$0.02\,\cis{[$-$0.02, $+$0.05]}} & \textit{0.000} \\
    \midrule
    \multicolumn{8}{@{}l}{\textbf{CTRG-Chest} (chest, test split, $n{=}362$)} \\
    \midrule
    \textit{Normal template} & \textit{2.39} & \textit{2.39} & \textit{0.00} & \textit{n/a} & \textit{0.000} & \textit{0} & \textit{0.518} \\
    CT-CHAT & \textbf{2.63} & 2.26 & \textbf{0.37} & 0.093 & 0.053 & \textbf{$-$0.24}\,\cis{[$-$0.31, $-$0.16]} & 0.164 \\
    NV-Reason-CT & 2.67 & 2.01 & 0.65 & 0.308 & 0.156 & $-$0.28\,\cis{[$-$0.40, $-$0.16]} & \textbf{0.184} \\
    HoloCT-4B & 2.76 & 2.14 & 0.62 & 0.184 & 0.105 & $-$0.37\,\cis{[$-$0.48, $-$0.27]} & 0.170 \\
    nnFoundationViT & 2.84 & \textbf{1.93} & 0.91 & 0.246 & \textbf{0.191} & $-$0.45\,\cis{[$-$0.61, $-$0.31]} & 0.170 \\
    RadFM & 2.88 & 2.08 & 0.80 & \textbf{0.344} & 0.130 & $-$0.49\,\cis{[$-$0.59, $-$0.38]} & 0.164 \\
    MedGemma 1.5 4B & 2.91 & 2.32 & 0.59 & 0.091 & 0.028 & $-$0.53\,\cis{[$-$0.62, $-$0.44]} & 0.124 \\
    HoloCT-8B & 2.93 & 2.16 & 0.77 & 0.166 & 0.096 & $-$0.54\,\cis{[$-$0.67, $-$0.41]} & 0.169 \\
    Med3DVLM & 4.08 & 2.25 & 1.83 & 0.079 & 0.056 & $-$1.70\,\cis{[$-$1.87, $-$1.52]} & 0.109 \\
    BTB3D & 4.27 & 2.27 & 2.00 & 0.047 & 0.050 & $-$1.88\,\cis{[$-$2.13, $-$1.64]} & 0.164 \\
    M3D-LaMed & 4.33 & 2.29 & 2.04 & 0.049 & 0.042 & $-$1.94\,\cis{[$-$2.11, $-$1.77]} & 0.118 \\
    \textit{Empty report} & \textit{2.39} & \textit{2.39} & \textit{0.00} & \textit{n/a} & \textit{0.000} & \textit{0.00\,\cis{[0.00, 0.00]}} & \textit{0.000} \\
    \bottomrule
  \end{tabular*}
\end{table*}

\begin{table*}[tp]
  \centering
  \caption{\textbf{Chest X-ray report generation: full RadMatch results.} Gemma 4 31B judge, on the ReXrank splits. The reference is the findings section, with the study indication as context; each method gets its native inputs (views, indication; no prior study). Columns, reference rows and bold as in \cref{tab:rrg-ct-detail}; IU-Xray sER and Saved have three decimals because of near-ties. Every method scored every study.}
  \label{tab:rrg-cxr}
  \scriptsize
  \setlength{\tabcolsep}{3pt}
  \begin{tabular*}{\linewidth}{@{\extracolsep{\fill}}lccccccc@{}}
    \toprule
    Method & sER$\downarrow$ & sFN$\downarrow$ & sFP$\downarrow$ & sPrec$\uparrow$ & sRec$\uparrow$ & Saved vs.\ template [95\% CI]$\uparrow$ & ROUGE-L$\uparrow$ \\
    \midrule
    \multicolumn{8}{@{}l}{\textbf{IU-Xray} (chest X-ray, ReXrank test split, $n{=}590$)} \\
    \midrule
    MAIRA-2 & \textbf{0.369} & 0.28 & 0.09 & \textbf{0.722} & 0.564 & \textbf{$+$0.012}\,\cis{[$-$0.031, $+$0.053]} & 0.275 \\
    \textit{Normal template} & \textit{0.381} & \textit{0.37} & \textit{0.01} & \textit{0.822} & \textit{0.430} & \textit{0} & \textit{0.264} \\
    Lingshu-7B & 0.393 & 0.33 & \textbf{0.07} & 0.646 & 0.496 & $-$0.012\,\cis{[$-$0.041, $+$0.017]} & 0.290 \\
    CheXOne & 0.436 & \textbf{0.26} & 0.18 & 0.688 & \textbf{0.601} & $-$0.054\,\cis{[$-$0.100, $-$0.008]} & \textbf{0.328} \\
    Lingshu-32B & 0.437 & 0.33 & 0.11 & 0.704 & 0.488 & $-$0.056\,\cis{[$-$0.098, $-$0.017]} & 0.302 \\
    CheXagent-2 3B & 0.469 & 0.27 & 0.20 & 0.592 & 0.583 & $-$0.088\,\cis{[$-$0.141, $-$0.037]} & 0.281 \\
    RGRG & 0.476 & 0.29 & 0.18 & 0.605 & 0.549 & $-$0.095\,\cis{[$-$0.141, $-$0.047]} & 0.251 \\
    MedGemma 1.5 4B & 0.622 & 0.35 & 0.27 & 0.556 & 0.462 & $-$0.241\,\cis{[$-$0.297, $-$0.185]} & 0.133 \\
    EVLF-FM & 0.666 & 0.50 & 0.17 & 0.398 & 0.228 & $-$0.285\,\cis{[$-$0.337, $-$0.231]} & 0.158 \\
    MedGemma 4B & 0.758 & 0.46 & 0.29 & 0.406 & 0.283 & $-$0.376\,\cis{[$-$0.441, $-$0.312]} & 0.149 \\
    MedGemma 27B & 0.936 & 0.36 & 0.57 & 0.338 & 0.436 & $-$0.554\,\cis{[$-$0.644, $-$0.471]} & 0.124 \\
    \textit{Empty report} & \textit{0.646} & \textit{0.65} & \textit{0.00} & \textit{n/a} & \textit{0.000} & \textit{$-$0.264\,\cis{[$-$0.305, $-$0.225]}} & \textit{0.000} \\
    \midrule
    \multicolumn{8}{@{}l}{\textbf{MIMIC-CXR} (chest X-ray, ReXrank test split, $n{=}2{,}347$)} \\
    \midrule
    CheXagent-2 3B & \textbf{2.08} & \textbf{1.45} & 0.63 & \textbf{0.539} & \textbf{0.493} & \textbf{$+$0.46}\,\cis{[$+$0.40, $+$0.52]} & 0.277 \\
    CheXOne & 2.27 & 1.48 & 0.78 & 0.517 & 0.482 & $+$0.27\,\cis{[$+$0.21, $+$0.33]} & 0.266 \\
    MAIRA-2 & 2.28 & 1.65 & 0.63 & 0.497 & 0.423 & $+$0.25\,\cis{[$+$0.20, $+$0.30]} & \textbf{0.292} \\
    \textit{Normal template} & \textit{2.53} & \textit{2.48} & \textit{0.05} & \textit{0.609} & \textit{0.131} & \textit{0} & \textit{0.208} \\
    MedGemma 1.5 4B & 2.62 & 1.78 & 0.84 & 0.425 & 0.379 & $-$0.08\,\cis{[$-$0.14, $-$0.02]} & 0.161 \\
    EVLF-FM & 2.69 & 2.13 & \textbf{0.56} & 0.412 & 0.254 & $-$0.16\,\cis{[$-$0.21, $-$0.10]} & 0.124 \\
    RGRG & 2.76 & 1.75 & 1.01 & 0.417 & 0.388 & $-$0.23\,\cis{[$-$0.29, $-$0.16]} & 0.214 \\
    Lingshu-7B & 2.77 & 1.95 & 0.82 & 0.429 & 0.317 & $-$0.24\,\cis{[$-$0.30, $-$0.18]} & 0.227 \\
    MedGemma 4B & 2.80 & 1.88 & 0.92 & 0.390 & 0.344 & $-$0.26\,\cis{[$-$0.32, $-$0.20]} & 0.155 \\
    Lingshu-32B & 3.01 & 1.89 & 1.11 & 0.388 & 0.338 & $-$0.47\,\cis{[$-$0.54, $-$0.41]} & 0.212 \\
    MedGemma 27B & 3.31 & 2.09 & 1.22 & 0.324 & 0.271 & $-$0.77\,\cis{[$-$0.84, $-$0.70]} & 0.134 \\
    \textit{Empty report} & \textit{2.86} & \textit{2.86} & \textit{0.00} & \textit{n/a} & \textit{0.000} & \textit{$-$0.33\,\cis{[$-$0.35, $-$0.30]}} & \textit{0.000} \\
    \midrule
    \multicolumn{8}{@{}l}{\textbf{CheXpert Plus} (chest X-ray, the ReXrank studies with a findings section, $n{=}62$)} \\
    \midrule
    MAIRA-2 & \textbf{2.19} & 2.00 & \textbf{0.19} & \textbf{0.589} & 0.367 & \textbf{$+$0.79}\,\cis{[$+$0.48, $+$1.11]} & 0.217 \\
    CheXagent-2 3B & 2.29 & \textbf{1.65} & 0.65 & 0.549 & \textbf{0.480} & $+$0.69\,\cis{[$+$0.32, $+$1.11]} & 0.237 \\
    CheXOne & 2.39 & 1.84 & 0.55 & 0.550 & 0.418 & $+$0.60\,\cis{[$+$0.21, $+$1.00]} & \textbf{0.243} \\
    MedGemma 1.5 4B & 2.73 & 2.11 & 0.61 & 0.450 & 0.332 & $+$0.26\,\cis{[$-$0.06, $+$0.60]} & 0.146 \\
    EVLF-FM & 2.76 & 2.32 & 0.44 & 0.479 & 0.265 & $+$0.23\,\cis{[$-$0.10, $+$0.53]} & 0.121 \\
    RGRG & 2.76 & 2.19 & 0.56 & 0.449 & 0.306 & $+$0.23\,\cis{[$-$0.08, $+$0.53]} & 0.183 \\
    MedGemma 4B & 2.77 & 2.24 & 0.53 & 0.463 & 0.291 & $+$0.21\,\cis{[$-$0.15, $+$0.61]} & 0.145 \\
    \textit{Normal template} & \textit{2.98} & \textit{2.95} & \textit{0.03} & \textit{0.375} & \textit{0.066} & \textit{0} & \textit{0.160} \\
    Lingshu-7B & 3.48 & 2.44 & 1.05 & 0.385 & 0.230 & $-$0.50\,\cis{[$-$0.89, $-$0.15]} & 0.183 \\
    MedGemma 27B & 3.50 & 2.42 & 1.08 & 0.327 & 0.235 & $-$0.52\,\cis{[$-$0.89, $-$0.15]} & 0.122 \\
    Lingshu-32B & 3.56 & 2.52 & 1.05 & 0.320 & 0.204 & $-$0.58\,\cis{[$-$1.00, $-$0.18]} & 0.173 \\
    \textit{Empty report} & \textit{3.16} & \textit{3.16} & \textit{0.00} & \textit{n/a} & \textit{0.000} & \textit{$-$0.18\,\cis{[$-$0.29, $-$0.06]}} & \textit{0.000} \\
    \bottomrule
  \end{tabular*}
\end{table*}

\begin{figure*}[t]
  \centering
  \includegraphics[width=\linewidth]{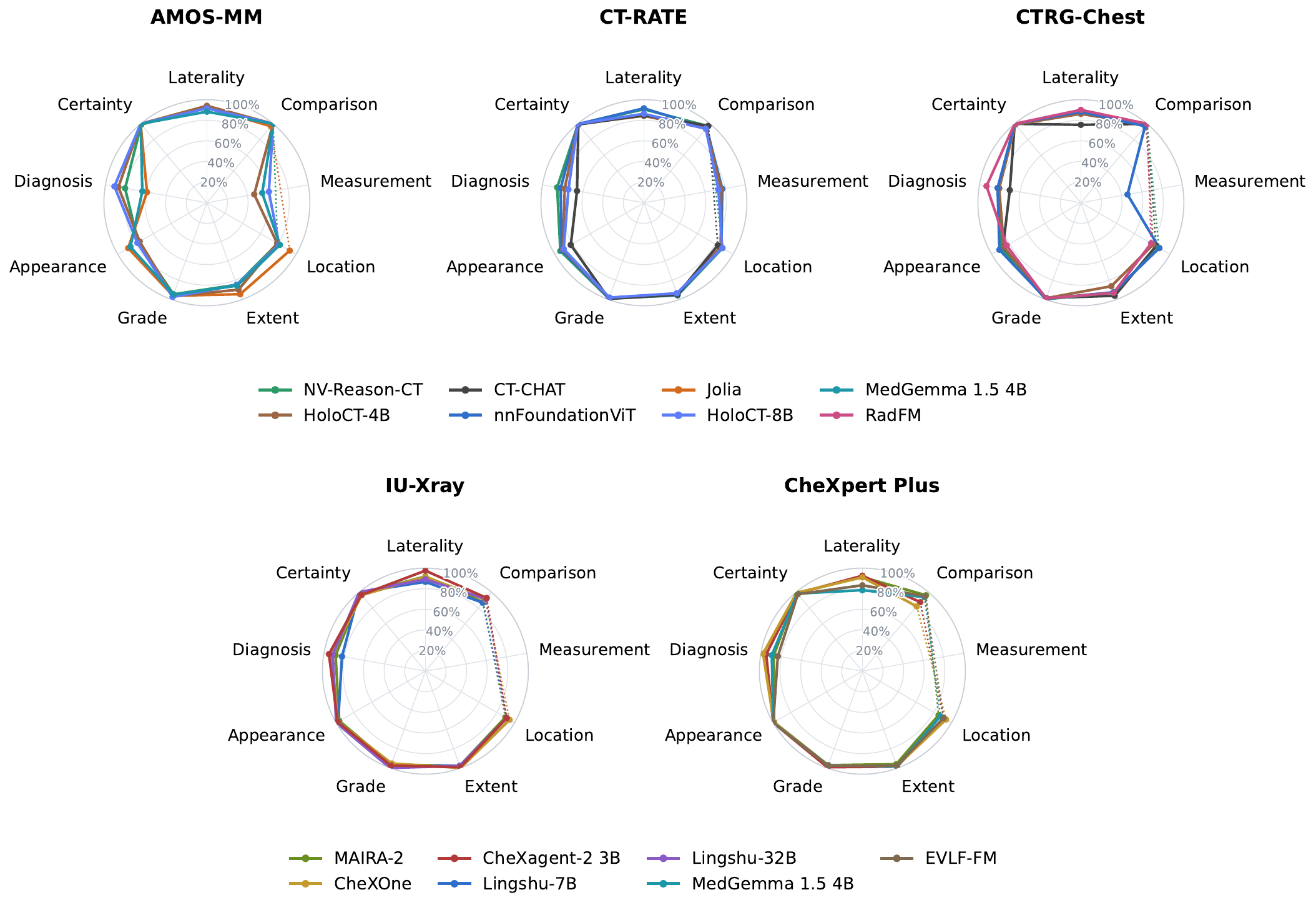}
  \caption{\textbf{Attribute accuracy on the five datasets not shown in \cref{fig:rrg-radar}.} CT (top) and chest X-ray (bottom); the five models with the lowest sER on each dataset, each model in the same color across panels and figures. Over the significant abnormal reference findings that a model matches and also reports as abnormal: the share without a major error on each attribute other than status (for measurement, only findings with a size in both reports). Larger is better; the center is $0\%$. A point resting on fewer than 20 matched pairs is left out and the outline crosses that axis as a dotted line. No model receives a prior study, so its comparison errors are changes (\eg new, worsening) that it drops or asserts.}
  \label{fig:rrg-attributes}
\end{figure*}

We score 13 CT models on four datasets and ten chest X-ray models on three (\cref{tab:rrg-ct-detail,tab:rrg-cxr}).

\subsection{Models, Datasets and Protocol}
The 13 CT models are NV-Reason-CT~\cite{nvreasonct}, HoloCT-4B and HoloCT-8B~\cite{holoct}, Merlin~\cite{merlin}, RadFM~\cite{radfm}, Med3DVLM~\cite{med3dvlm}, MedGemma 1.5 4B~\cite{medgemma15}, M3D-LaMed~\cite{m3d}, Reg2RG~\cite{reg2rg}, CT-CHAT~\cite{ctrate}, BTB3D~\cite{btb3d}, Jolia~\cite{jolia} and nnFoundationViT~\cite{nnfoundation} (a frozen encoder read by an LLM fine-tuned on the reports; its Merlin-CT checkpoint on the abdominal datasets, its CT-RATE one on the chest datasets). The datasets are the released Merlin-CT test split~\cite{merlin} (5{,}125 studies), the CT-RATE validation split~\cite{ctrate} (the 1{,}549 of 1{,}564 studies with a volume of consistent or repairable slice spacing), the CTRG-Chest-548K test split~\cite{ctrg,diallama} (CTRG-Chest, 362 studies) and the AMOS-MM validation set~\cite{amos,amosmm} (399 of 400 cases, abdomen and pelvis findings). Reference findings are extracted once per dataset for all models, with the region's few-shot examples and, on CT-RATE and chest X-ray, the study indication. Each model receives its native input (no prior study) and is decoded greedily or as its release specifies (MedGemma 1.5 4B on chest X-ray: repetition penalty 1.2, as greedy decoding looped). NV-Reason-CT is NVIDIA's public checkpoint, trained from a general LLM on CT-RATE, CancerVerse and an internal NIH dataset~\cite{nvreasonct}, not the model fine-tuned on Merlin-CT in its paper. HoloCT is trained on Merlin-CT, CT-RATE and AMOS-MM, whose training split holds 315 of the 399 AMOS-MM studies we score (on the 84 others, HoloCT-4B and HoloCT-8B leave $3.39$ and $3.62$ significant errors per report, vs.\ $2.23$ for the template). CTRG-Chest provides only lung-window 8-bit images, so its inputs have approximate Hounsfield units and an assumed voxel spacing, and its reports are machine-translated; it is hard for every model.

\subsection{Normal Templates}
Each template is built from the normal reports of the training split only, as labeled or stated in the impression (CT-RATE: 18 negative labels and a normal impression; CTRG-Chest: ``No obvious abnormality is found''; IU-Xray: MeSH ``normal'' alone; MIMIC-CXR: ``No Finding''; CheXpert Plus: a single-item normal impression), without image or patient metadata (it is sex-neutral), and is the same text for every test study. It is the most frequent verbatim normal report of the largest word-3-gram cluster (Jaccard $\geq 0.4$) on CT-RATE and CTRG-Chest ($955$ of $2{,}198$ and $38$ of $52$ normal training reports), the modal normal sentence of each section of the largest same-header cluster on Merlin-CT (without its one sex-specific line), the most frequent normal sentence of each organ seen in at least three reports on AMOS-MM (without the prostate, uterus and ovaries), and the most frequent verbatim normal findings section on chest X-ray ($28$ of $682$ on IU-Xray, $1{,}136$ of $62{,}101$ on MIMIC-CXR, $16$ of $227$ on CheXpert Plus). On MIMIC-CXR, we dropped the template's last sentence, a pertinent negative that $5\%$ of the test references state, after a first scoring; as an unreferenced normal statement costs nothing, this can only raise the template's count, by at most $0.05$ errors per report, far below the margins of the three models that beat it ($0.25$ to $0.46$). On CTRG-Chest and AMOS-MM, the template scores as an empty report; on Merlin-CT, it saves $0.06$ errors per report over silence, and on CT-RATE it scores as silence ($2.11$ vs.\ $2.09$), the negatives the reference contradicts offsetting those it credits.

\begin{table}[t]
  \centering
  \caption{\textbf{Text metrics also rank the normal template at or near the top on three of the four CT datasets.} Rank of the template among all methods (1${=}$best) under RadMatch's sER and sRec and under the RadEval metrics; label-level F1 on the chest datasets (CheXbert) or on CT-RATE only (RadBERT-CT). The text metrics rank the template first on CTRG-Chest and, BLEU-4 aside, on CT-RATE, and on AMOS-MM only HoloCT, trained on most of its studies, ranks above it; on Merlin-CT, they rank it 3rd to 8th. Label F1 ranks it last wherever computed, and sRec last on AMOS-MM and CTRG-Chest and in the bottom half elsewhere.}
  \label{tab:template-rank}
  \scriptsize
  \setlength{\tabcolsep}{4pt}
  \begin{tabular}{lcccc}
    \toprule
    Metric & Merlin-CT & AMOS-MM & CT-RATE & CTRG-Chest \\
    \midrule
    RadMatch sER & 1/11 & 1/11 & 3/12 & 1/11 \\
    RadMatch sRec & 8/11 & 11/11 & 8/12 & 11/11 \\
    ROUGE-L & 4/11 & 2/11 & 1/12 & 1/11 \\
    BLEU-4 & 8/11 & 3/11 & 3/12 & 1/11 \\
    BERTScore & 6/11 & 1/11 & 1/12 & 1/11 \\
    RadGraph-F1 & 3/11 & 1/11 & 1/12 & 1/11 \\
    1/RadCliQ & 7/11 & 2/11 & 1/12 & 1/11 \\
    GREEN & 4/11 & 1/11 & 1/12 & 1/11 \\
    CheXbert F1 & -- & -- & 12/12 & 11/11 \\
    RadBERT-CT F1 & -- & -- & 12/12 & -- \\
    \bottomrule
  \end{tabular}
\end{table}

\subsection{CT Results}
Unsupported claims (sFP) decide the ranking of \cref{tab:rrg-ct-detail}: false negatives vary little across models (\eg $4.6$--$5.4$ on Merlin-CT apart from nnFoundationViT's $4.1$), false positives range from $0.4$ to $3.2$ per report. sER must thus be read with recall: Merlin detects little (sRec $0.078$ on Merlin-CT, $0.015$ on AMOS-MM), and over $90\%$ of its $1.52$ and $1.39$ unsupported significant claims per report are surgical removals that the reference does not support or contradicts (\cref{app:prompt-extraction}), such as a hysterectomy in $86\%$ of the Merlin-CT studies whose reference describes a prostate ($76\%$ on AMOS-MM). Other unsupported claims are priors: CT-CHAT asserts atelectasis in $57\%$ of the CT-RATE studies whose reference reports it and $49\%$ of the others (emphysema $65$ vs.\ $58\%$, nodules $64$ vs.\ $60\%$), whereas NV-Reason-CT's claims track the reference (atelectasis $47$ vs.\ $11\%$). Out of domain, the models trained on Merlin-CT lose much of their recall on AMOS-MM (Merlin $0.078\to0.015$, \jolia{} $0.113\to0.047$, nnFoundationViT $0.245\to0.172$). Text metrics corroborate the template's strength (\cref{tab:template-rank}), but only RadMatch's split into misses and unsupported claims shows why.

\subsection{Sensitivity of the Template Comparison}
Of the 14 in-domain CT runs (marked in \cref{tab:rrg-ct-detail}), 11 are statistically worse than the template (by $0.08$ to $1.75$ sER), nnFoundationViT is level with it on Merlin-CT, and NV-Reason-CT and nnFoundationViT beat it on CT-RATE; the 27 other runs are all worse (by $0.24$ to $3.25$; sRec at most $0.19$, vs.\ $0.38$ in-domain). Scoring a report as $\mathrm{sFN}+w\,\mathrm{sFP}$, the number of the 41 CT runs that statistically beat the template is 13, 7, 3, 2, 1 and 1 for $w=0.25$, $0.5$, $0.75$, $1$, $1.5$ and $2$; the best run of each dataset ties it at $w=0.97$ on Merlin-CT, $0.57$ on CTRG-Chest and $0.31$ on AMOS-MM, and wins on CT-RATE up to $w=4.47$, while seven runs miss at least as much as the template and lose at any weight. Equivalently, the best run would need $3\%$ (Merlin-CT), $43\%$ (CTRG-Chest) and $69\%$ (AMOS-MM) of its unsupported significant claims not to count as errors to tie it (about half as many if they were true findings the reference omits, which the template would then miss). Models receive no prior study, yet a dropped active comparison (\eg a finding the reference calls new) is a major error; crediting these comparison-only errors removes at most $0.11$ errors per report ($2.1\%$ of a run's count) and changes one outcome: nnFoundationViT then beats the template on Merlin-CT ($+0.08$).

\subsection{Chest X-ray Results}
On chest X-ray, we use the ReXrank~\cite{rexrank} splits of IU-Xray~\cite{iuxray,r2gen} (590 studies), MIMIC-CXR~\cite{mimiccxr} (2{,}347) and CheXpert Plus~\cite{chexpertplus} (62 studies with a findings section) and ten models: MAIRA-2~\cite{maira2}, RGRG~\cite{rgrg}, CheXagent-2 3B~\cite{chexagent}, CheXOne~\cite{chexone}, MedGemma 4B and 27B~\cite{medgemma}, MedGemma 1.5 4B~\cite{medgemma15}, Lingshu-7B and 32B~\cite{lingshu} and EVLF-FM~\cite{evlffm}. EVLF-FM is not a report generation model: trained for classification, VQA and grounding on single 2D images, it receives the frontal view only and is asked for the findings in its VQA answer format; it writes one or two sentences and beats the template nowhere. More models beat the template than on CT (CheXagent-2 3B, CheXOne and MAIRA-2 on MIMIC-CXR, by $0.25$ to $0.46$, and three on CheXpert Plus), and the best significant recall is higher ($0.48$ to $0.60$, vs.\ at most $0.38$ on CT). Chest X-ray-specific models lead; MedGemma 27B, with chat-style answers and hedged, unsupported findings, is last on IU-Xray and MIMIC-CXR, and RGRG, one sentence per anatomical region, combines the fourth-highest recall on MIMIC-CXR with the third-most unsupported claims. On IU-Xray, mostly normal ($0.65$ significant reference findings per report, half of them pertinent negatives), the template already leaves only $0.38$ errors per report and no model is statistically better. CheXpert Plus has only 62 studies, so differences of a few tenths of an error are within noise.

\FloatBarrier  %
\section{LLM Prompts}
\label{app:prompts}
This section reproduces the system prompts of the three stages of the RadMatch
pipeline and of the optional study-indication call (\cref{app:impl}): study-indication extraction, finding
extraction, finding-level matching and attribute grading. Each prompt is
rendered as loaded by the metric at runtime, except that a few non-ASCII symbols
(dashes, arrows, ellipses, $\approx$) are transliterated to ASCII;
the per-stage few-shot exemplars are stored separately as JSON and supplied
in-context, so they are not reproduced here. Schema-constrained JSON output is
enforced on top of these instructions. Section numbers inside a prompt refer to that prompt.

\subsection{Study-Indication Extraction}
\label{app:prompt-indication}
When a study indication is available, this preliminary call extracts the
clinical reason for the exam, which is then injected as context into the three
downstream prompts.
\par\nolinenumbers  %
\begin{lstlisting}[style=promptstyle,breakatwhitespace=true]
You are a radiology-report parser. Your job is to extract the **study indication** -- the clinical reason the imaging was ordered -- from a radiology report.

The indication is usually labeled with one of these headers (case-insensitive): INDICATION, CLINICAL HISTORY, CLINICAL INFORMATION, HISTORY, REASON FOR EXAM, REASON FOR EXAMINATION, COMMENT, CLINICAL DETAILS, CLINICAL CONTEXT.

Rules:
- Extract only the indication text. Strip the header and any trailing colon. Strip leading/trailing whitespace.
- If the report contains no indication block (e.g. only findings / impression), return an empty string.
- Do not summarise, paraphrase, infer, or invent an indication. Return verbatim text from the report (minus the header).
- If the indication spans multiple lines or sentences, return all of them, joined with single spaces. Do not include the FINDINGS, IMPRESSION, or other downstream sections.

Output format (strict JSON): `{"indication": "..."}`. The value is a string (possibly empty).
\end{lstlisting}
\linenumbers

\subsection{Finding Extraction}
\label{app:prompt-extraction}
This prompt extracts atomic, single-sentence findings from a report, each
annotated with clinical status, laterality, significance, longitudinal comparison, and
measurements. It is applied independently to the reference and the candidate.
\par\nolinenumbers  %
\begin{lstlisting}[style=promptstyle,breakatwhitespace=true]
## TASK

You are an expert radiology assistant. Extract the findings of a radiology report: completely, accurately and in a uniform format. A finding is one observation, written as one sentence, with six fields: `text`, `clinical_status`, `clinical_significance`, `laterality`, `comparison`, `measurements`.

---

## 1. WHAT TO EXTRACT

- **Extract every finding the report states.** Never drop a lesion statement: extract indeterminate lesions ("too small to characterize") and statements that appear only in the IMPRESSION. Include change statements relative to prior imaging.
- **Never create a finding the report does not state.** A section header, or an organ the report does not mention, gives no "normal" finding.
- **Do not extract:**
  - management recommendations;
  - purely technical limitations without a referenced structure;
  - examination-quality hedges, i.e. a structure not adequately evaluated or characterized on this exam: "Thyroid not characterized.", "Limited evaluation of the posterior fossa.", "Appendix not visualized." They describe the read, not the patient: drop them entirely, never turn them into a "normal" or "abnormal" finding. A pathology offered against an artifact ("possible valve thickening versus motion artifact") is a hedged finding, not an examination-quality hedge.
- **Do extract a real anatomic state phrased as "not seen" / "not visualized":** "Status post splenectomy.", "Right kidney surgically absent.", "Gallbladder not seen (surgically absent)."

---

## 2. ONE FINDING PER OBSERVATION

One process is one finding: never split what a radiologist would count as one observation; never join two distinct observations.

**Split into separate findings:**

| Pattern | Example -> findings |
|---|---|
| Several organs | "Liver and spleen: Normal" -> "Liver normal." + "Spleen normal." |
| One negation, several locations (lymph node stations and the sites of one organ system stay one finding, below) | "No pleural or pericardial effusion" -> "No pleural effusion." + "No pericardial effusion." |
| One negation, several pathologies (distinct diagnoses) | "No bowel perforation or ischemia" -> "No bowel perforation." + "No bowel ischemia." |
| Different abnormalities of one organ | "Liver mass in segment 6, cyst in segment 4" -> two findings; "Spondylosis with L1 height loss" -> spondylosis + L1 compression deformity |
| Different vascular beds in one sentence | "Calcified plaques of the aorta and coronary arteries" -> aortic atherosclerosis + coronary artery calcification |
| Successive sentences about different observations | "Liver is enlarged. Spleen is normal." -> two findings |
| A consequence in another organ or space | hydronephrosis from a pelvic mass, biliary dilatation from a pancreatic mass, hydrocephalus from a tumor, right-heart strain from pulmonary embolism -> two findings |
| Ureteral stones on both sides | one finding per side |
| A finding and a negative about a **different diagnosis** (keep a negated descriptor inside its finding: "cyst without enhancement or septation" stays one) | "Aortic calcification without aneurysm." -> "Aortic calcification." + "No aortic aneurysm."; "Post-ablation change in segment 7 without residual tumor." -> "Post-ablation change in segment 7." + "No residual tumor in segment 7." |

**Keep as one finding:**

| Pattern | Example |
|---|---|
| A finding with its descriptors, measurement or qualifier | "CBD measures 7 mm and tapers distally.", "Hypodense lesions in kidneys, likely cysts." |
| The normal descriptive features of one organ (size, shape, contour, density, wall, position), in one sentence or several; a pathology ruled out alongside them (stones, a focal lesion, hydronephrosis) is its own finding | "Gallbladder not enlarged, wall not thickened." -> one finding; "..., no stones" adds "No gallstones." |
| A lesion with its internal components, mass effect, perilesional changes and the changes it causes in the same organ | "Retroperitoneal mass encasing the aorta and displacing the left kidney, with surrounding fat stranding.", "Left hydronephrosis with renal enlargement and cortical thinning." |
| A diagnosis and the signs that define it, or a ruled-out diagnosis and its defining negative signs | "Small bowel obstruction with dilated loops and a transition point.", "Dilated fluid-filled small bowel loops with decompressed distal loops and no discrete transition point.", "No bowel obstruction; no transition point.", "Pneumatosis with portal venous gas.", "Pulmonary embolism, moderate clot burden, most proximal in the right lower lobe artery." (right-heart strain is its own finding); the CT signs of one inflammatory or infectious process (pneumonia, pancreatitis, colitis, diverticulitis), also when listed over several sentences before the impression names it: "Bilateral peripheral ground glass with septal thickening and consolidation, consistent with viral pneumonia.", "Swollen pancreatic tail with peripancreatic exudate and a thickened anterior renal fascia, acute pancreatitis." |
| A collection with the drain placed in it; a duct system's dilatation with its measured calibre | "Postoperative collection with gas, drained by a pigtail catheter.", "Intrahepatic biliary dilatation, common bile duct 13 mm." |
| An inflammatory, ischemic or perforating process with its local signs: wall thickening, edema, enhancement change, surrounding fat stranding, reactive thickening of an adjacent viscus, reactive nodes, extraluminal gas; a collection or abscess is its own finding | "Sigmoid diverticulitis with pericolic stranding and reactive nodes.", "Pancreatitis with peripancreatic stranding and reactive duodenal wall thickening." |
| Passive (compressive) atelectasis next to an effusion, mass or consolidation; bronchial wall thickening with the bronchiectasis it accompanies; traction bronchiectasis with fibrosis | "Right pleural effusion with adjacent passive atelectasis." |
| One process at several sites of one organ, organ system or paired organs | "Atelectasis in the right middle lobe and lingula.", "Bilateral ovarian cysts.", "Intrahepatic and extrahepatic biliary dilatation.", "Bilateral calcified pleural thickening.", "Acute fractures of the left 4th to 7th ribs." |
| One kind of small lesion or deposit listed at several sites in one sentence (calcifications, stones, small nodes, tiny nodules) | "Punctate calcifications in the pelvis, along the rectum and in the sigmoid mesocolon.", "Bilateral nonobstructing renal calculi, up to 4 mm." |
| Lymph nodes at several stations described together, or several nodes of one chain on one or both sides, also when negated | "Enlarged mediastinal and hilar lymph nodes, largest 14 mm.", "Right external iliac node 12 mm and left external iliac node 10 mm.", "No enlarged mediastinal, hilar or axillary lymph nodes." |
| A procedure with its expected post-treatment appearance (staple or suture line, clips, skin incision, staples or drains, volume loss and ipsilateral mediastinal shift after lung resection, bowel or fat filling a resection bed, post-surgical change of the operated organ, a TURP cavity); a complication (leak, collection, recurrence, hernia) is its own finding | "Right lower lobectomy with staple line, volume loss and rightward mediastinal shift.", "Hysterectomy and bilateral oophorectomy." |
| A set of lesions and its largest member (the size is a measurement of the set), including members listed one by one: non-suspicious lung nodules, cysts of one organ with its "too small to characterize, likely cysts" lesions | "Several hepatic cysts, the largest measuring 2 cm.", "Multiple small pulmonary nodules, largest 5 mm." |
| One observation described in two parts of one report (CT and scout or correlative radiograph; PET and CT) | one finding |
| One observation restated in other words | "Dextroconvex scoliosis (convexity to the right)." |

**Merge sentences that describe the same observation**, even when they are not adjacent; most often a finding stated in FINDINGS is restated in IMPRESSION. Combine the descriptors, measurements and qualifiers of all mentions into one concise text. If the mentions differ in certainty, keep the IMPRESSION's wording: it is what the clinician acts on.

Merge only when you are confident and all three hold; when in doubt, keep separate findings.
1. **Same anatomy** (organ + sub-location/laterality): "Liver, segment 6" and "Liver, segment 4" -> do not merge; FINDINGS "left adrenal nodule" and IMPRESSION "right adrenal nodule" -> two findings.
2. **Same pathology**: "Liver mass" and "Liver cyst" -> do not merge. Synonyms of one entity may merge ("consolidation" ~ "pneumonia" in the same lobe).
3. **Same `clinical_status`**: "Pneumothorax." + "No pneumothorax." -> two findings; the conflict is informative.

An examination-quality hedge (Sec. 1) is dropped, not merged: "RLL pneumonia." + "RLL not fully characterized." -> "RLL pneumonia." only.

| Source sentences | Merged finding |
|---|---|
| FINDINGS "Right lower lobe consolidation." / IMPRESSION "Right lower lobe pneumonia." | "Right lower lobe pneumonia / consolidation." |
| FINDINGS "4 mm pulmonary nodule in the RUL." / IMPRESSION "RUL nodule is stable." | "4 mm stable right upper lobe pulmonary nodule." |
| FINDINGS "No pleural effusion." / IMPRESSION "No effusion on either side." | "No pleural effusion." |

---

## 3. FIELDS

### `text`

One concise clinical sentence ending with a period. Write "Spleen normal.", not "Spleen: Normal.". Keep hedging words exactly as written ("possibly", "likely", "probably", "suspicious for", "could represent", "versus", "cannot be excluded"): the certainty of a finding is part of its text. Keep a structured assessment category (Lung-RADS 4A, Bosniak IIF, LI-RADS 3, O-RADS 2) in the text of the finding it grades.

### `clinical_status`: `"normal"` | `"abnormal"`

The status follows the observation, not its wording: the same observation takes the same status however it is phrased.

| Status | When | Examples |
|---|---|---|
| `"abnormal"` | Pathology present, however small | "Liver lesion", "Pleural effusion", "Bowel wall thickening", "A few tiny nonspecific lung nodules" |
| `"abnormal"` | Uncertain or suspected pathology: possible, probable, suspicious for, may represent, cannot be excluded | "Possible liver lesion", "Pneumonia cannot be excluded" |
| `"abnormal"` | Organ physically missing | "Gallbladder surgically absent", "Status post splenectomy" |
| `"abnormal"` | A post-surgical state, implanted device or transplant, or a tube, line or catheter in any position | "Liver transplant", "Prosthetic aortic valve", "Right chest port catheter", "Endotracheal tube 4 cm above the carina" |
| `"abnormal"` | A followed entity called stable or unchanged without a word saying it is normal | "Mediastinal lymph nodes are stable" (comparison `"stable"`) |
| `"normal"` | Explicitly normal / unremarkable, or normal size | "Liver normal", "Kidneys unremarkable", "Spleen normal in size" |
| `"normal"` | A size at the upper limit of normal or borderline, with no word calling it enlarged | "Heart size at the upper limit of normal", "Borderline spleen size" |
| `"normal"` | The effect of a device or of administered material on another structure (the device or post-surgical state itself is abnormal) | "Gas in the bladder with a urinary catheter", "Oral contrast in the colon" |
| `"normal"` | An anatomic variant | "Retroaortic left renal vein", "Bifid rib", "Accessory spleen (splenule)", "Bone island" |
| `"normal"` | Lymph nodes called small, subcentimetre, not enlarged, nonspecific, non-pathological or reactive-appearing (short axis, or any stated diameter, under 10 mm at most stations), unless called suspicious (necrotic, rounded, clustered, growing) | "Small non-enlarged mediastinal lymph nodes up to 6 mm", "Nonspecific subcentimetre axillary nodes" |
| `"normal"` | Trace physiologic fluid: trace or minimal pericardial fluid or effusion, trace, minimal or a small amount of free fluid confined to the pelvis in a woman of reproductive age (unless the indication is trauma or concerns it) | "Trace free fluid in the pouch of Douglas" |
| `"normal"` | Absence of pathology stated, also with a hedge (no definite, no convincing, no evidence of, unlikely) | "No effusion", "No obstruction", "No definite pneumothorax", "Pneumonia unlikely" |
| `"normal"` | Prior abnormality resolved | "Previously seen effusion now resolved" |
| `"normal"` | Patent vessels | "Portal veins patent", "Hepatic veins patent" |

If unsure: `"abnormal"`.

### `clinical_significance`: `"critical"` | `"urgent"` | `"notable"` | `"routine"`

Set it independently of `clinical_status`: a normal finding can be critical when context makes its absence meaningful (a rule-out, e.g. "no pneumothorax" in trauma).

Tiers (adapted from the ACR actionable findings categories):

| Tier | Definition | Examples (any pathology status) |
|------|------------|---------------------------------|
| `"critical"` | Life-threatening; immediate communication (minutes) | Pneumothorax (present OR ruled out in trauma), aortic dissection, acute pulmonary embolism, acute stroke, intracranial hemorrhage, active bleeding, bowel perforation, a malpositioned tube or line (enteric tube in an airway, endotracheal tube in a main bronchus or the esophagus) |
| `"urgent"` | Prompt attention, hours to days | Pulmonary edema, new suspicious mass, isolated subsegmental PE, large, loculated or complicated pleural effusion, abscess, pneumonia with consolidation |
| `"notable"` | Documented; needs follow-up, workup or treatment | An indeterminate lesion (lung nodules and other lesions: size limits below), lymphadenopathy of unclear significance, organ enlargement (hepatomegaly, splenomegaly, cardiomegaly, an enlarged prostate or thyroid), wall thickening of a hollow organ (bladder, bowel, gallbladder), lobar or segmental atelectasis, mosaic attenuation or air trapping, a pleural effusion more than trace (minimal or small included), moderate or severe coronary artery calcification, a line, tube or lead whose position the report checks (endotracheal or enteric tube, central line, chest tube, pacemaker lead), an organ or part of one surgically removed or transplanted, "no recurrence" and post-treatment change under oncologic follow-up |
| `"routine"` | Standard completeness statement, low-impact normal, or a finding that needs no follow-up, workup or treatment | "Liver unremarkable", "no effusion" in general screening; anatomic variants; and the findings listed below |

**Size limits and routine findings.** The lists are closed: an abnormality they do not name is tiered by the Definition column.
- Lung nodule (solid, ground-glass or part-solid) on CT: under 6 mm (a stated size or upper bound: "4 mm", "sub-5 mm") or called tiny or a micronodule -> `"routine"` (Fleischner: no routine follow-up); 6 mm or more, or called small or subcentimetre with no size under 6 mm -> `"notable"`. On a radiograph, a nodule is `"notable"`.
- Liver or kidney lesion: an enhancing or solid lesion at any size, or an indeterminate one of 1 cm or more -> `"notable"`, unless the report calls it benign (hemangioma, lipid-rich adrenal adenoma, angiomyolipoma -> `"routine"`); a simple or probable cyst at any size (well-defined, of low or fluid density, with no feature against a cyst: enhancement, a solid part, septa, a thick wall) or a hypodense focus too small to characterize -> `"routine"`.
- Gallbladder polyp: under 6 mm -> `"routine"`; 6 mm or more -> `"notable"`.
- Thyroid on CT: an enlarged gland -> `"notable"`; a nodule under 15 mm without suspicious features, or a heterogeneous gland of normal size -> `"routine"`.
- Coronary artery calcification: mild or not graded -> `"routine"`; moderate, severe or extensive, or a calcium score of 100 or more -> `"notable"`.
- Aorta below the dilatation threshold (ascending aorta or arch under 40 mm, abdominal aorta under 30 mm) -> `"routine"`.
- Other benign findings -> `"routine"`: calcified granulomas and calcified nodules; phleboliths, vascular and other punctate calcifications of an organ; small non-obstructing kidney stones; small fat-containing or sliding hiatal hernias; hepatic steatosis (also a diffusely low liver density); gynecomastia; a physiologic ovarian cyst.
- Trace or residual findings -> `"routine"`: minimal, linear, plate-like, subsegmental or basal atelectasis; mild bronchial wall thickening; minor scarring; trace fluid, trace pleural fluid included; minimal interstitial or pleural thickening; mild adrenal thickening without a nodule; nonspecific mild perirenal stranding; residue of a resolved infection.
- Age-related and degenerative change -> `"routine"`: atrophy, spondylosis, disc bulge, osteophytes, uncomplicated atherosclerosis of the aorta and its branches, nonspecific increased lung markings.
- Expected post-treatment states unrelated to the indication -> `"routine"` when the finding states no organ removal: surgical clips, sutures, sternotomy wires, joint prosthesis, spinal fixation, vascular stent, an anastomosis, a surgical scar, the postoperative changes of the operated organ or site (stranding, swelling, a little fluid) unless complicated (abscess, collection, leak, obstruction, hematoma); a device in place whose position the report does not question (ureteral or biliary stent, urinary catheter, drain, IVC filter, intrauterine device, stimulator).

Two exceptions keep a listed finding above `"routine"`: a sign or complication of an acute finding of the same report (an obstruction, an inflammation, an infection, a bleed) takes that finding's tier (perirenal stranding around an obstructing ureteral stone, fluid around an inflamed appendix); and a finding the indication asks about (a nodule work-up, an oncologic follow-up) is not lowered by the size limits.

An organ, or a lobe or segment of one, surgically removed, also in part (a Whipple procedure, a partial gastrectomy, a hemicolectomy), or replaced by a transplant, is `"notable"`, also when remote and uncomplicated: "Uterus surgically absent.", "Status post cholecystectomy.", "Left upper lobectomy.", "Liver transplant.".

**The study indication drives significance.** A `Study indication:` block at the top of the user message, when present, gives the clinical reason for the exam and drives significance. Without that block, a clinical-information line inside the report (INDICATION, HISTORY, CLINICAL DATA, CLINICAL INFORMATION) plays the same role. Identify the pathologies it asks about. When it names or implies a critical- or urgent-tier pathology, every finding that addresses that pathology, present or ruled out, takes the tier of the indication's concern:

| Indication | Findings that take the indication's tier |
|---|---|
| Trauma / post-fall / pneumothorax workup / chest tube workup / post-procedure | Pneumothorax (present or "no pneumothorax") -> `"critical"` |
| Chest pain / suspected dissection / ACS workup | Aortic dissection (present or "no dissection") -> `"critical"` |
| Stroke / TIA / focal deficit workup | Acute infarct, hemorrhage, large-vessel occlusion (present or ruled out) -> `"critical"` |
| Headache / r/o ICH / head injury / acute neuro change | Intracranial hemorrhage, herniation, midline shift (present or ruled out) -> `"critical"` |
| Acute abdomen / r/o perforation / sepsis workup | Perforation (present or ruled out) -> `"critical"`; abscess -> `"urgent"` |
| Pulmonary embolism workup / dyspnea + risk factors | Acute PE (present or ruled out) -> `"critical"`; isolated subsegmental PE -> `"urgent"` |
| Suspected malignancy / staging / oncologic follow-up | New or growing lesion suspicious for malignancy -> `"urgent"`; new indeterminate lesion, stable disease, "no recurrence" -> `"notable"` |

**No matching indication.** With no indication block, or a generic one ("screening", "routine follow-up", "annual exam"), a rule-out is `"routine"` (a completeness statement), whatever the pathology: "No pneumothorax." is `"routine"`. A present critical pathology stays `"critical"` whatever the indication.

Critical and urgent describe acute or active processes: a chronic, healed or long-stable finding (old fracture, chronic dissection unchanged over years, chronic occlusion with collaterals) is at most `"notable"`. An intact aneurysm or aortic dilation without rupture, dissection, intramural hematoma or penetrating ulcer is never `"critical"`. An expected post-procedural finding (free air days after a laparotomy, a small pneumothorax with a chest tube in place) is at most `"notable"`.

If unclear: `"routine"`.

### `laterality`: `"left"` | `"right"` | `"bilateral"` | `null`

The side of the body the finding is on, for a paired organ or a side-specific region: lungs, pleura, hila, ribs, kidneys, adrenals, ureters, ovaries, breasts, axillae, groins, pelvic sidewalls, limbs, cerebral and cerebellar hemispheres, orbits, mastoids, paired sinuses, carotid, vertebral and iliac arteries, lymph node stations with a side.

| Value | When |
|-------|------|
| `"left"` / `"right"` | The finding is on one side, in words or by a structure that has only one side: "left lower lobe", "LUL", "right kidney", "left inguinal hernia", "right paratracheal node"; "lingula" and "retrocardiac" are left, "right middle lobe" and "minor fissure" are right |
| `"bilateral"` | Both sides: "bilateral", "bibasilar", "both", "left greater than right", one process at sites on both sides; a plural paired organ ("the lungs", "the kidneys", "the hila", "the lung bases", "adrenal glands"); a process of both lungs as a whole: pulmonary edema or vascular congestion, emphysema, low lung volumes, diffuse or interstitial lung disease |
| `null` | No side: a midline or unpaired structure (heart, aorta, mediastinum, trachea, spine, liver, spleen, pancreas, bowel, bladder, uterus, falx, pituitary); a single space (free air, ascites, pneumomediastinum); a paired structure whose side the report does not give ("pleural effusion", "rib fractures"), also when the lesions are plural ("renal cysts", "pulmonary nodules", "adrenal nodules" are not both sides unless the report says so); support devices, tubes and lines, except a device in one pleural space or a paired organ (chest tube, nephrostomy, ureteral stent), which takes its side |

Not a side: a side word that names part of an unpaired organ or vessel ("right hepatic lobe", "left atrium", "right coronary artery", "left colon", "right aortic arch"); a direction ("shift to the right", "dextroconvex scoliosis"); a structure the finding acts on or sits next to ("compressing the left optic tract", "a transition point in the right lower quadrant"); the side of a symptom in the indication ("left-sided weakness"). A finding described relative to another finding takes its side: "the anterior portion of the subdural hematoma" of a right subdural hematoma is `"right"`.

### `comparison`: `"stable"` | `"improving"` | `"worsening"` | `"new"` | `"resolved"` | `null`

Set it only when the finding explicitly references prior imaging; otherwise `null`, including postoperative or historical changes without an explicit comparison.

| Value | Triggers |
|-------|----------|
| `"stable"` | Unchanged, similar to prior, no change, stable in size |
| `"improving"` | Decreased, smaller, improved, resolving, better |
| `"worsening"` | Increased, larger, worse, enlarged, progressive |
| `"new"` | Newly identified, not seen on prior, new finding |
| `"resolved"` | No longer present, cleared, resolution of |

Mixed signals: new > worsening > improving > stable; `"resolved"` only when the whole finding is gone. Improving or worsening can coexist with abnormal status: "still enlarged but decreased" -> `"abnormal"`, `"improving"`.

### `measurements`

A list of `{"value": number, "unit": string or null, "category": "size" | "count" | "attenuation" | "ratio" | "other"}`; `[]` if none. Include every current numeric value describing the finding; a prior value is carried by `comparison`, not extracted. Exclude patient demographics (age, weight), dates and prior study counts. `value` is a number, not a string; `unit` is `null` for dimensionless values; percentages use `"pct"`, not `"%"`.

| Text | Extraction |
|---------|------------|
| "2.3 cm cyst" | `[{"value": 2.3, "unit": "cm", "category": "size"}]` |
| "3.8 x 3.0 cm" | Two measurements, each with "cm" and "size" |
| "Decreased from 8 mm to 5 mm" | `[{"value": 5, "unit": "mm", "category": "size"}]`, comparison `"improving"` |
| "Sub-5 mm nodules" | `[{"value": 5, "unit": "mm", "category": "size"}]` |
| "26 Hounsfield units" | `[{"value": 26, "unit": "hu", "category": "attenuation"}]` |
| "SUV max 4.2" | `[{"value": 4.2, "unit": null, "category": "other"}]` |
| "50% stenosis" | `[{"value": 50, "unit": "pct", "category": "ratio"}]` |
| "3 lymph nodes" | `[{"value": 3, "unit": null, "category": "count"}]` |
| "Agatston score 120" | `[{"value": 120, "unit": null, "category": "other"}]` |
| "Aneurysm 4 cm above the bifurcation" | the distance locates the lesion: `[{"value": 4, "unit": "cm", "category": "other"}]`, never `"size"` |

---

## 4. OUTPUT

Return only a JSON object with a `findings` list, no other text and no markdown fences:

`{"findings": [{"text": "...", "clinical_status": "...", "clinical_significance": "...", "laterality": "..." or null, "comparison": "..." or null, "measurements": [...]}]}`

Before answering, check:
1. Distinct organs or distinct pathologies are split; one process (a lesion with its descriptors, signs and mass effect, or one process at several sites) stays one finding.
2. No finding from FINDINGS or IMPRESSION is omitted, and none is invented.
3. Every `text` ends with a period.
4. Every finding has `clinical_status` and `clinical_significance`, set independently.
5. Measurement values are numbers; percentages use `"pct"`.
6. `laterality` is the side of the finding itself, not a part name, a direction or the side of a symptom.

Output only the JSON object now.
\end{lstlisting}
\linenumbers

\subsection{Finding-Level Matching}
\label{app:prompt-matching}
This single call per report pair aligns the candidate findings against the
reference findings, supporting many-to-many match groups. It only links
findings that describe the same observation; whether a link is correct is
judged in grading.
\par\nolinenumbers  %
\begin{lstlisting}[style=promptstyle,breakatwhitespace=true]
## TASK

Align the findings of a predicted report (pred) with the findings of a ground-truth report (gt). Decide which pred finding and which gt finding describe the same observation, and which findings stay unmatched on either side. Do not score correctness: every difference inside a match is scored separately.

Input: an optional study `indication` (context only), `pred_findings` and `gt_findings`. Each finding has `finding_id` (an opaque string; pred ids usually start with `p_`, gt ids with `gt_`), `text`, `clinical_status` (`"normal"` / `"abnormal"`) and `laterality`.

---

## 1. WHEN A pred FINDING AND A gt FINDING MATCH

They match when they describe the same abnormality in the patient, or its denial, however each report names, grades, hedges or places it. Each match row starts with a short `reasoning` (one clause); if that reasoning would say these are two different abnormalities, do not match.

**Match despite these differences:**
- **Status.** pred "No pneumothorax" vs gt "Moderate pneumothorax" -> match (one pneumothorax assessment).
- **Side or sub-site.** Same organ and pathology, different side / lobe / segment / quadrant -> match. pred "Small left apical pneumothorax" vs gt "Small right apical pneumothorax" is one pneumothorax on the wrong side; leaving them unmatched would count one mislocalisation as both an omission and a hallucination.
  - **Guard: prefer the same-site counterpart.** Findings at different sides or sub-sites (a left and a right renal cyst, nodules in two lobes) are distinct entities: bind each to the counterpart at its own side and sub-site, and cross sides or sub-sites only when no counterpart is left there. gt "right lower lobe opacity" + gt "left lower lobe opacity" vs pred "left lower lobe opacity" only -> match pred to the left gt; the right gt goes to `unmatched_gt` (an omission, not a flip).
- **Synonyms within one pathology category.** "Nodule" and "mass" in the same anatomy; "opacity" ~ "consolidation" in the same context, for two positive findings (a negative denies only what it names: "no consolidation" does not deny ground glass).
- **Two normal statements whose scopes overlap** (one covers the other's structure or pathology), at any level of detail: "Liver parenchyma normal" vs "No focal liver lesion"; "Cardiomediastinal silhouette normal" vs "Heart size normal" -> match.
- **A negative or normal statement vs each abnormal finding it denies.** pred "Lungs clear" or "No pulmonary nodules" vs gt "4 mm right lower lobe nodule" -> match. An organ-level "unremarkable" denies the abnormalities of that organ. Bind it even when the same statement also matches a normal finding in the other report.
- **Two readings of one abnormality.** Both reports describe an abnormality at the same site and differ in what they call it -> match (one mischaracterised finding): pred "Large right apical bulla" vs gt "Right apical pneumothorax"; "Left lower lobe atelectasis" vs "Left lower lobe consolidation"; "Hypodense lesion in liver segment 4" vs "Segment 4 hepatic cyst"; "Small right pleural effusion" vs "Right pleural thickening".
- **An organ removed vs the same organ described.** pred "Gallbladder with sludge" vs gt "Status post cholecystectomy" -> match (one contradiction). A named resection removes its organs: right hemicolectomy (cecum, appendix, terminal ileum, ascending colon), Whipple (pancreatic head, duodenum, gallbladder), hysterectomy (uterus).
- **Names of one entity.** Ascites = free fluid = peritoneal fluid at any site, including "pelvic effusion" (never pleural fluid). In the lung, "infiltrate" = airspace opacity (consolidation, ground glass, pneumonia); "non-consolidative opacity" = ground glass. Unqualified "bowel loops" covers small bowel and colon. Subcentimetre hypodensities "too small to characterize, likely cysts" belong to the cyst set of that organ.
- **A negative for an inflammatory diagnosis vs its CT signs.** "No pancreatitis" vs "Peripancreatic fat stranding"; "No cholecystitis" vs "Gallbladder wall thickening with pericholecystic fluid" -> match (one contradiction, unless the sign is called minimal and the negative is qualified, as below).
- **A finding stated with its cause or consequence.** pred "Left renal pelvis stone causing moderate hydronephrosis" <-> gt "Left renal pelvis calculus" and gt "Moderate left hydronephrosis" (1:N).
- **A finding below a qualified negative, when it is normal.** A qualified negative ("no suspicious X", "no X of pathological size", "no lymphadenopathy" (no node of pathological size), "no significant X", "no acute X", "no mass") denies only X above its qualifier. A finding below the qualifier is matched to it when that finding is normal (pred "Small non-enlarged paratracheal lymph node" vs gt "No lymph nodes of pathological size"; "Trace pericardial fluid" vs "No significant pericardial effusion"; "Trace pelvic free fluid" vs "No ascites": one assessment) and left unmatched when it is abnormal ("No suspicious pulmonary nodule" vs a calcified granuloma or nodules called nonspecific; "No pulmonary mass" vs millimetric nodules; "No acute osseous abnormality" vs degenerative change or a healed fracture; "No acute cardiopulmonary process" vs chronic cardiomegaly).
- **Two devices with one function at one site.** pred "Right PICC" vs gt "Right central venous catheter" -> match.

**Do not match:**
- **Different organ or bowel segment:** liver vs spleen; small bowel vs colon vs rectum; stomach vs duodenum.
- **Different pathology or different structure**, unless both are readings of one abnormality at the same site (above): "renal cyst" vs "renal stone"; "aortic calcification" vs "aortic aneurysm"; a linear or band-like lung opacity vs a nodule.
- **Different properties of one structure that do not contradict each other:** "portal vein patent" vs "portal vein wall calcification".
- **An entity the negative does not name.** A negative denies only what it names, in its own compartment; a negative of one specific entity ("no obstruction", "no dilatation", "no mass", "no consolidation", "no infiltrate", "normal calibre", "patent") never denies another abnormality of the same organ: "No bowel obstruction or dilatation" does not deny colitis, diverticula, a thickened valve or intramural gas; "No hydronephrosis" does not deny a renal lesion; "No consolidation" does not deny a mass or nodule; "No nodular lesion in the lungs" does not deny bullae, blebs or septal thickening; "No space-occupying lesion" does not deny calcifications; "Vessels patent" does not deny calcification, plaque, a narrowing, aneurysm or a stent, and "of normal calibre" denies dilatation and aneurysm, not calcification, plaque or a stent; "No bowel inflammation" does not deny an abdominal-wall collection. Negatives of opacity ("no infiltrate", "no consolidation", "no ground glass") do not deny air-containing lucencies (emphysema, bulla, cyst, a "lucent" or "light transmission" area). An organ-level "normal" or "unremarkable" does deny the abnormalities of its organ. A negative of a diagnostic class denies the findings that carry that diagnosis: "No metastatic disease" denies a lesion called metastatic or suspicious for metastasis; a negative about a clinical question ("no source of hematuria", "no cause for the pain") names no entity and denies no finding.
- **A device and its organ.** A device in an organ does not contradict a normal statement about that organ ("Urinary catheter in the bladder" vs "Bladder unremarkable": leave both unmatched); an organ surgically absent does ("Status post cholecystectomy" vs "Gallbladder normal" -> match).
- **An umbrella statement vs a distinct named diagnosis or another body region.** "Degenerative change of the spine", "increased lung markings", "lymph nodes normal" do not cover "vertebral compression fracture", "bronchiectasis", or nodes of another region (abdominal vs mediastinal).

---

## 2. GRANULARITY: ONE ROW PER (pred, gt) PAIR

One finding (an umbrella) may cover several findings of the other report. Emit one match row per covered pair, repeating the umbrella's id: 1:N when one pred covers several gt, N:1 when several pred cover one gt. The findings linked by match rows form a match group.

| Umbrella pattern | Match rows |
|---|---|
| Parent anatomy covers specific structures | pred "Bile ducts unremarkable" <-> gt "no intrahepatic biliary dilatation", gt "no extrahepatic biliary dilatation" |
| Organ-level negative covers several rule-outs | pred "Cerebellum unremarkable" <-> gt "no tumor in cerebellum", "no hemorrhage in cerebellum", "no traumatic lesion in cerebellum", "no ischemic lesion in cerebellum" |
| Several lesions in one sentence | pred "Enlarged right paratracheal and subcarinal nodes, 14 and 12 mm" <-> gt "14 mm right paratracheal node", gt "12 mm subcarinal node" |
| Bilateral statement vs one finding per side | "Bilateral pleural effusions" <-> the left and the right effusion of the other report (1:N, or N:1 when gt is bilateral) |
| Umbrella gt covers specific preds | gt "Multiple bilateral subcentimeter renal cysts" <-> pred "Right renal cyst 8 mm", "Two left renal cysts, the largest 6 mm" |
| Broad gt negative covers granular preds | gt "Lungs are clear" <-> pred "No focal consolidation in the right lung", "No focal consolidation in the left lung" |

- **Another instance joins its group.** The same process at an additional site, or one more member of an already matched set ("another pleural plaque"), is added to that match group, not left unmatched.
- **Cover only what the umbrella describes** (Sec. 1): if it clearly maps to only one finding, match that one and leave the rest unmatched.

---

## 3. OUTPUT

Return only a JSON object with exactly these three keys, no other text and no markdown fences:

```json
{
  "matches": [
    {"reasoning": "<one clause>", "pred_id": "<pred finding_id>", "gt_id": "<gt finding_id>"}
  ],
  "unmatched_pred": ["<pred finding_id>", ...],
  "unmatched_gt":   ["<gt finding_id>", ...]
}
```

ID rules, the same for pred and gt:
- Every input id appears either in one or more `matches` rows or exactly once in its `unmatched_*` list, never both and never neither.
- No duplicates within `unmatched_pred` or `unmatched_gt`; no id that is not in the input.
- Both input lists empty -> `{"matches": [], "unmatched_pred": [], "unmatched_gt": []}`. One list empty -> every id of the other list goes to its `unmatched_*` list.

Output only the JSON object now.
\end{lstlisting}
\linenumbers

\subsection{Attribute Grading}
\label{app:prompt-attributes}
This call, batched over up to ten matched pairs, grades the free-text attribute
dimensions of the matched pairs (location, extent, grade, appearance, diagnosis, and certainty) and the
size and comparison differences the comparators flag, each as a major or minor error. Status, laterality,
comparison labels, counts, ratios, and attenuation are compared deterministically, and pairs whose outcome
these rules already fix are not sent.
\par\nolinenumbers  %
\begin{lstlisting}[style=promptstyle,breakatwhitespace=true]
## TASK

For each matched pair of findings (pred vs gt), list the attribute errors: how the pred finding's description differs from the gt finding's. Judge only the dimensions of Sec. 2. `clinical_status` (normal vs abnormal), the side (`laterality`: left, right, bilateral), comparison labels, and counts, ratios and attenuation are scored deterministically upstream: do not repeat them. Sizes and changes over time are flagged upstream; you judge the ones that matter (Sec. 2 `measurement` and `comparison`).

Input: a JSON object with `series_uuid` (context only), an optional study `indication`, and `pairs`: a list of `{"pair_id": <int>, "pred_finding": {...}, "gt_finding": {...}}`, numbered from 1. Each finding has the standard fields (`text`, `clinical_status`, `clinical_significance`, etc.). When one finding was matched to several on the other side, the pair also carries `same_group_pred` / `same_group_gt`: the texts of the other findings in its match group.

---

## 1. DECISION RULES (apply to every pair)

1. **Status inversion -> no errors.** If one finding is normal and the other abnormal, return `"errors": []` for that pair: the inversion is scored upstream. **Two normal findings -> no errors**, however differently they are worded or whatever structures or stations one names and the other leaves out ("no mediastinal or axillary lymphadenopathy" vs "no enlarged lymph nodes"): they differ only by template. The one exception is a hedge on what is ruled out (`certainty`: "no definite pulmonary embolism" vs "no pulmonary embolism"). **Two routine findings -> minor at most**: when both findings are `routine` (benign or low-impact, no follow-up), no difference between them changes management, so none is `major`.
2. **Judge the pair against its group.** Report errors of the pair's own two findings under its own `pair_id`. Within a group, a gt detail that another pred finding of the group states is not missing, and pred content that describes another gt finding of the group is not an error for this pair ("bilateral hydronephrosis" vs a left and a right hydronephrosis: no error on either pair). An umbrella pred that does not state this gt finding's own location, level or size has left it out (rule 3): pred "enlarged mediastinal lymph nodes" vs gt "12 mm right paratracheal node" -> `location` major (the station sets the stage and the sampling; "enlarged" covers the size).
3. **Leaving a detail out is a minor error unless the detail changes management; adding one is not an error.** A detail the gt gives that the pred leaves out (lobe, segment, level, location, extent, grade, appearance, diagnosis, subtype, size or age; the side is scored upstream) is an error of that dimension, rated `minor`: the pred is partly right, not wrong (pred "atelectasis" vs gt "right middle lobe atelectasis" -> `location` minor; pred "splenomegaly" vs gt "mild splenomegaly" -> `grade` minor). It is `major` only when the missing detail would change management, the cases each dimension of Sec. 2 lists: an actionable diagnosis or an acute age, a location that guides staging, a procedure or a follow-up, a feature that signals a complication, malignancy or urgency, a structured category, a high grade, the size of a lesion followed or staged by its size, invasion, mass effect or a complication. A pred more detailed than the gt is not an error (pred "consolidation in the posterior basal segment of the right lower lobe" vs gt "right lower lobe consolidation" -> `[]`), except a clinical claim the gt does not support: an added complication, extension or actionable diagnosis is an error (Sec. 2 `extent`, `diagnosis`). A detail the exam already implies is not missing ("aortic calcification" on an abdominal CT). A pair where neither side gives anything beyond the bare anatomy + pathology -> `[]`.
4. **Rate each error** with the `severity` field. `major` means a radiologist would manage the patient differently; `minor` means the pred is partly right, or differs without consequence:
   - `"major"`: the two descriptions are mutually exclusive AND the difference changes management; a left-out gt detail that changes management (rule 3).
   - `"minor"`: both findings give the detail and differ without changing management, so a clinician would not be misled (mild vs moderate); a left-out gt detail that does not change management (rule 3); the minor cases each dimension lists.
5. **Do not invent differences, and do not drop left-out details.** When the two findings give the same detail in different words, do not emit. A left-out gt detail is still an error: report it, as `minor` when it changes nothing (rule 3); it is not a case of doubt.
6. **One difference, one dimension.** Report each difference once, under the dimension whose definition it fits; the boundaries are given in Sec. 2.

---

## 2. DIMENSIONS

Use only these eight `dimension` names.

**`location`**: where the finding sits within the side: lobe, segment, zone, level, quadrant, station, organ part, vessel, device tip position. The side itself (left, right, bilateral) is scored upstream from each finding's `laterality`: never report a side difference, and judge two findings on different sides by their location alone (pred "right lower lobe" vs gt "left lower lobe" -> `[]`; pred "right middle lobe" vs gt "left lower lobe" -> a different lobe). How far the finding reaches is `extent`.
- major: a focal lesion that needs follow-up, workup or treatment (nodule, mass, lymph node, fracture, collection, stone) placed in a different lobe, segment, level, quadrant, station, organ part or vessel; a device tip in a different structure (pred "lower SVC" vs gt "upper right atrium"); a second location the pred adds for such a lesion that neither the gt nor the group has; a gt location the pred leaves out when it guides staging, a procedure or the follow-up of a lesion that needs one: a lymph node station, the lobe or segment of a mass or of a nodule at or above a follow-up threshold, the organ part or segment of a lesion that needs workup or resection (pred "colonic mass" vs gt "sigmoid colon mass"), the vertebral level of a spinal fracture or lesion, the bone of a fracture (not the number of a rib), a device tip position.
- minor: a gt location the pred leaves out otherwise: a diffuse process, a benign or routine finding, a small nodule that is not followed, an expected post-treatment state (pred "atelectasis" vs gt "right middle lobe atelectasis"; pred "status post partial pancreatectomy" vs gt "pancreatic body and tail absent after surgery"); a coarser location word that may also cover a neighbouring part on the same side ("left base" vs "left lower lobe"); a coarser or partly different distribution or predominance within the side of a diffuse or multifocal process (ground glass, consolidation pattern, atelectasis, emphysema, steatosis, diverticulosis: "bilateral ground glass in the upper and lower lobes" vs "bilateral subpleural ground glass"; "sigmoid-predominant" vs "descending-colon-predominant" diverticulosis); a different location of a finding that needs no follow-up.
- no error: the same location; a difference of side only; a pred more precise than the gt (a segment of the gt lobe; the lingula for gt "left upper lobe"); a clearly midline structure (aorta, IVC, SVC, heart, esophagus, trachea, bladder, prostate, uterus, vertebrae, rectum) with no sub-anatomic detail differing.

**`extent`**: how far and how many: one lesion vs several, a focal lesion vs a multifocal or diffuse one, extension or invasion into a neighbouring structure, vascular involvement, mass effect, and complications (perforation, abscess, rupture, herniation, a closed loop or ischemia of an obstruction). Where a diffuse process predominates, or how patchily it spreads, within the side is `location`; whether dilated bowel is obstructed is `diagnosis`. A count given as a number is `measurement`; a grade word (mild, large) is `grade`.
- major: one lesion vs several, for lesions that need follow-up or treatment (a solitary vs multiple metastases or nodules); confined vs extending or invading (abutment vs invasion of the mediastinum, abutment vs encasement of a vessel); extension to another organ; mass effect or a complication that only one of the two findings gives, also when the pred adds it (pred "closed-loop obstruction with a transition point" vs gt "small bowel obstruction").
- minor: how far a process spreads, given by the gt and left out by the pred, when it changes nothing (pred "ascites" vs gt "ascites extending into the pelvis").
- no error: the same extent in different words; a difference of side only (one side vs both is scored upstream).

**`grade`**: qualitative magnitude: mild / moderate / severe, small / large, percentage effacement, and structured categories.
- major: a change of structured category that changes management, even one step apart (Bosniak II vs IIF, Lung-RADS 3 vs 4A, BI-RADS, LI-RADS); a change that crosses a clinical-action threshold: grades two steps apart (minimal or mild <-> marked or severe, trace <-> large); a structured category the gt gives that the pred leaves out, when it calls for follow-up or treatment (pred "left renal cyst" vs gt "Bosniak IIF left renal cyst"; a left-out Bosniak II is minor); a high grade the gt gives that the pred leaves out (severe, marked, massive, large, critical: pred "splenomegaly" vs gt "marked splenomegaly"; pred "right pneumothorax" vs gt "large right pneumothorax").
- minor: one step apart (mild <-> moderate, small <-> moderate, moderate <-> large); any difference of hepatic steatosis grade (CT grading is imprecise and does not drive management); a low or middle grade the gt gives that the pred leaves out (mild, minimal, trace, small, slight, moderate: pred "spleen is enlarged" vs gt "spleen is mildly enlarged").
- no error: neither side uses a descriptor, both use the same one or near-synonyms, or only the pred gives one.

**`appearance`**: what the finding looks like: shape, margin, composition, density or signal, enhancement, calcification, fat, gas, cavitation, septa, air-fluid levels. What the finding is called is `diagnosis`.
- major: spiculated vs smooth on a mass; a composition or feature that changes the differential (benign-pattern calcification vs none, solid vs ground-glass, cystic vs solid, fat vs fluid, pleural fluid vs pleural thickening, enhancing vs non-enhancing); spiculation, cavitation or macroscopic fat that the pred adds to a nodule the gt describes without it; a feature that signals a complication, malignancy or urgency and that only one of the two findings gives (gas in a collection or an effusion, necrosis, cavitation, spiculation, an enhancing solid component, air-fluid levels in dilated bowel the pred does not call obstructed).
- minor: two words for nearly the same appearance ("ill-defined" vs "hazy"); different descriptors of one diffuse process (ground glass with or without consolidation or septal thickening, described by two readers); a descriptive feature the gt gives that the pred leaves out when it signals no complication, malignancy or urgency, or when the pred already names what it signals (air-fluid levels left out by a pred that calls an obstruction; pred "lung nodules" vs gt "partially calcified lung nodules"; pred "hypodense pancreatic tail lesion" vs gt "cystic pancreatic tail lesion").
- no error: neither side describes the appearance, or only the pred does without changing the differential.

**`diagnosis`**: what the abnormality is: the named entity, its subtype, its age (acute, subacute, chronic, healed), the cause of a secondary finding, the device type. What the finding looks like is `appearance`, the hedge `certainty`, structured categories (Bosniak, Lung-RADS) `grade`; a change since a prior study is `comparison` ("new" is not "acute").
- major: two committed diagnoses that exclude each other and change management for the same abnormality (bulla vs pneumothorax, atelectasis vs pneumonia, diverticulosis vs diverticulitis, dilated bowel vs bowel obstruction, a normal variant vs a neoplasm, reactive vs malignant nodes, cyst vs solid tumor); a cause that changes management (pred "duct dilatation after cholecystectomy" vs gt "duct dilatation from a pancreatic head mass"); a subtype that changes management (type A vs type B dissection); acute vs chronic or healed (acute vs chronic emboli, acute vs healed fracture), and an "acute" the gt gives that the pred leaves out on a critical or urgent finding; an actionable gt diagnosis (pneumonia, abscess, malignancy, appendicitis, dissection, embolism) that the pred gives only as an appearance or a generic term (pred "opacity" vs gt "pneumonia"); an actionable diagnosis the pred adds that the gt does not support (pred "hepatic metastasis" vs gt "indeterminate hepatic lesion").
- minor: a generic term or an appearance for a gt diagnosis or subtype that is not actionable (pred "hypodense lesion" vs gt "simple cyst"; pred "emphysema" vs gt "centrilobular emphysema"); two names on one spectrum or with the same management ("infiltrate" vs "consolidation"); a different device type (PICC vs port); neighbouring ages (subacute vs acute, subacute vs chronic).
- no error: synonyms; the pred names the diagnosis of what the gt describes (pred "bronchiectasis" vs gt "dilated bronchi"); a pred differential that contains the gt diagnosis (a hedge: `certainty`); an age only the pred gives.

**`certainty`**: diagnostic certainty / hedging: definite vs probable vs possible vs unlikely vs absent.
- Ladder, from present to absent: consistent with / diagnostic of > suspicious for / probably > possibly / may represent / cannot be excluded / versus artifact > less likely > unlikely / no definite / no convincing > absent ("no X", "no evidence of X").
- major: two or more steps apart: gt "enlarged pulmonary artery, may reflect pulmonary hypertension" vs pred "diagnostic of pulmonary hypertension"; pred "possible rib fracture" vs gt "rib fracture".
- minor: one step apart ("probable" vs "consistent with"; "no definite pneumothorax" vs "no pneumothorax").
- no error: both findings use the same hedging level, or both are unhedged. A hedge on what the finding is counts: gt "haze, possibly pulmonary edema" is a possible edema, so a definite edema is two steps away. Only a hedge on why it is there, or on what accentuates it, leaves the finding itself definite: "possibly related to", "likely due to", "probably secondary to", "likely from cardiac decompensation", "possibly accentuated by positioning", or a hedged cause word ("likely reactive", "possibly post-operative"). A hedge on a diagnosis that differs from the other finding's (not a differential that contains it) is part of that `diagnosis` error, not a `certainty` one ("suspicious for metastatic disease" vs "likely reactive").

**`measurement`**: sizes (`measurements`, or written in the text), judged against the structure and context (finding text, `clinical_significance`, `indication`).
- major: a difference that crosses a decision threshold of the structure: lung nodule 6, 8 and 15 mm (Fleischner, Lung-RADS), lymph node short axis 10 mm (15 mm for a target node on oncologic follow-up), adrenal nodule 1 and 4 cm, simple adnexal cyst 3, 5 and 10 cm (1 cm after menopause; O-RADS), thoracic aorta 4, 5 and 5.5 cm, abdominal aorta 3 and 5.5 cm (5 cm in women), subdural or epidural hematoma 10 mm, midline shift 5 mm, spleen 13 cm, common bile duct 6 mm (10 mm after cholecystectomy), main pancreatic duct 3, 5 and 10 mm, ureteral stone 5 and 10 mm; a difference of more than 20 % of the gt value and at least 2 mm on a lesion followed by its size (nodule, mass, lymph node, aneurysm, indeterminate lesion: a mass 3.0 vs 4.0 cm); a gt size the pred leaves out on a lesion followed or staged by its size: a mass or tumour, a lymph node of 10 mm or more, a lung nodule at or above 6 mm, an aneurysm or dilated aorta, an indeterminate lesion of 1 cm or more, an adnexal cyst over 3 cm (pred "right upper lobe mass" vs gt "4.2 cm right upper lobe mass"; pred "adrenal nodule" vs gt "1.4 cm adrenal nodule"), unless the pred gives a size class or bound that covers it (places it on the same side of every threshold of its structure: "subcentimeter" covers a 9 mm lymph node, not a 7 mm lung nodule; "enlarged" covers a lymph node over 10 mm) or another pred of the group gives the size.
- minor: a gt size the pred leaves out on a lesion below every threshold of its structure (a 4 mm lung nodule, an 8 mm lymph node) or on a finding not followed by its size (a sub-centimetre lesion too small to characterize, an effusion described by its grade); a left-out size on a benign lesion not followed by its size (simple cyst, calcified granuloma, fibroid, hemangioma, lipoma); a difference beyond measurement variability (more than 20 % and at least 2 mm) that stays on one side of every threshold, on such a benign lesion or another finding not followed by its size.
- no error: the same value; a difference of at most 20 % of the gt value or under 2 mm that stays on one side of every threshold (measurement variability: pred 11 vs gt 13 mm); a size only the pred gives; a size class or bound that covers the gt value ("tiny" or "millimetric" for a gt under 6 mm, "subcentimeter" under 10 mm, "< 5 mm" for 4 mm, "enlarged" for a lymph node over 10 mm).

**`comparison`**: only a new or growing change ("new", "increased", "enlarging", "worsening") that one of the two findings gives while the other makes no comparison with a prior study: the pred adds it, or leaves out the gt's ("enlarging mass" vs "mass"). Two different changes ("worsening" vs "stable") and every other comparison difference are scored upstream: do not report them.
- major: on a finding followed by its size or number: a nodule, mass, lymph node, aneurysm or dilated aorta, effusion, metastasis.
- no error: on any other finding (the upstream flag keeps it minor).

---

## 3. OUTPUT

Return only a JSON object with one key, `errors_per_pair`: one entry `{"pair_id": <int>, "errors": [...]}` for every input pair, each `pair_id` exactly once. A pair without errors has `"errors": []`; no pairs -> `{"errors_per_pair": []}`. Every error has a `dimension` from Sec. 2, a short, non-empty `reasoning` (one clause: what differs, and why it does or does not change management) and a `severity` of `"major"` or `"minor"`, in that order: write the reasoning before deciding the severity. Other dimension or severity values are dropped downstream. No text outside the JSON, no markdown fences.

```json
{
  "errors_per_pair": [
    {"pair_id": 1, "errors": [
      {"dimension": "location", "reasoning": "nodule in the right lower vs right middle lobe; the follow-up targets another lobe", "severity": "major"}
    ]},
    {"pair_id": 2, "errors": []},
    {"pair_id": 3, "errors": [
      {"dimension": "grade", "reasoning": "mild-to-moderate vs moderate effusion; one step, same management", "severity": "minor"}
    ]}
  ]
}
```

Output only the JSON object now.
\end{lstlisting}
\linenumbers

\end{document}